 \documentclass[final,5p,times,twocolumn,authoryear]{elsarticle}

\usepackage{amssymb}
\usepackage{lipsum}

\usepackage{array} 
\usepackage{longtable, booktabs}

\usepackage{tikz}
\usetikzlibrary{positioning,calc}

\usepackage{amsmath}

\usepackage{cuted}
\usepackage{capt-of}   

\usepackage{dblfloatfix}

\usepackage{pdflscape} 

\usepackage{natbib}
\usepackage{booktabs}

\usetikzlibrary{arrows.meta,positioning,shapes.geometric}
\usepackage{float}

\usepackage{booktabs,multirow,tabularx}
\newcolumntype{Y}{>{\raggedright\arraybackslash}X}

\usepackage{graphicx}
\usepackage[
    colorlinks=true,
    linkcolor=black,   
    urlcolor=blue,     
    citecolor=black    
]{hyperref}


\journal{Nuclear Physics A}

\begin{document}

\begin{frontmatter}

\title{A Survey on Bridging EEG Signals and Generative AI: From Image and Text to Beyond}

\author[addr1]{Shreya Shukla}
\author[addr1]{Jose Torres}
\author[addr1]{Akshaj Murhekar}
\author[addr1]{Christina Liu}
\author[addr1]{Abhijit Mishra \corref{cor1}}
\author[addr1,addr2]{Jacek Gwizdka}
\author[addr1]{Shounak Roychowdhury}

\address[addr1]{
School of Information, The University of Texas at Austin,
Austin, TX, USA
}

\address[addr2]{
Institute of Applied Computer Science,
Łódź University of Technology,
Łódź, Poland
}

\cortext[cor1]{%
    Correspondence to: Abhijit Mishra, 1616 Guadalupe St Suite 5.202, Austin, TX, USA, 78701.\\
    E-mail addresses: abhijitmishra@utexas.edu, shreya.shukla@utexas.edu
}

\begin{abstract}
Decoding neural activity into human-interpretable representations is a key research direction in brain-computer interfaces (BCIs) and computational neuroscience. Recent progress in machine learning and generative AI has driven growing interest in transforming non-invasive Electroencephalography (EEG) signals into images, text, and audio. This survey consolidates and analyzes developments across EEG-to-image synthesis, EEG-to-text generation, and EEG-to-audio reconstruction. We conducted a structured literature search across major databases (2017-2025), extracting key information on datasets, generative architectures (GANs, VAEs, transformers, diffusion models), EEG feature-encoding techniques, evaluation metrics, and the major challenges shaping current work in this area. Our review finds that EEG-to-image models predominantly employ encoder-decoder architectures built on GANs, VAEs, or diffusion models; EEG-to-text approaches increasingly leverage transformer-based language models for open-vocabulary decoding; and EEG-to-audio methods commonly map EEG signals to mel-spectrograms that are subsequently rendered into audio using neural vocoders. Despite promising advances, the field remains constrained by small and heterogeneous datasets, limited cross-subject generalization, and the absence of standardized benchmarks. By consolidating methodological trends and available datasets, this survey provides a foundational reference for advancing EEG-based generative AI and supporting reproducible research. We further highlight open-source datasets and baseline implementations to facilitate systematic benchmarking and accelerate progress in EEG-driven neural decoding.
\end{abstract}




\end{frontmatter}




\section{Introduction \& Motivation}
\label{sec:intro}

\begin{figure*}
    \centering
    {\includegraphics[width=0.90\textwidth]{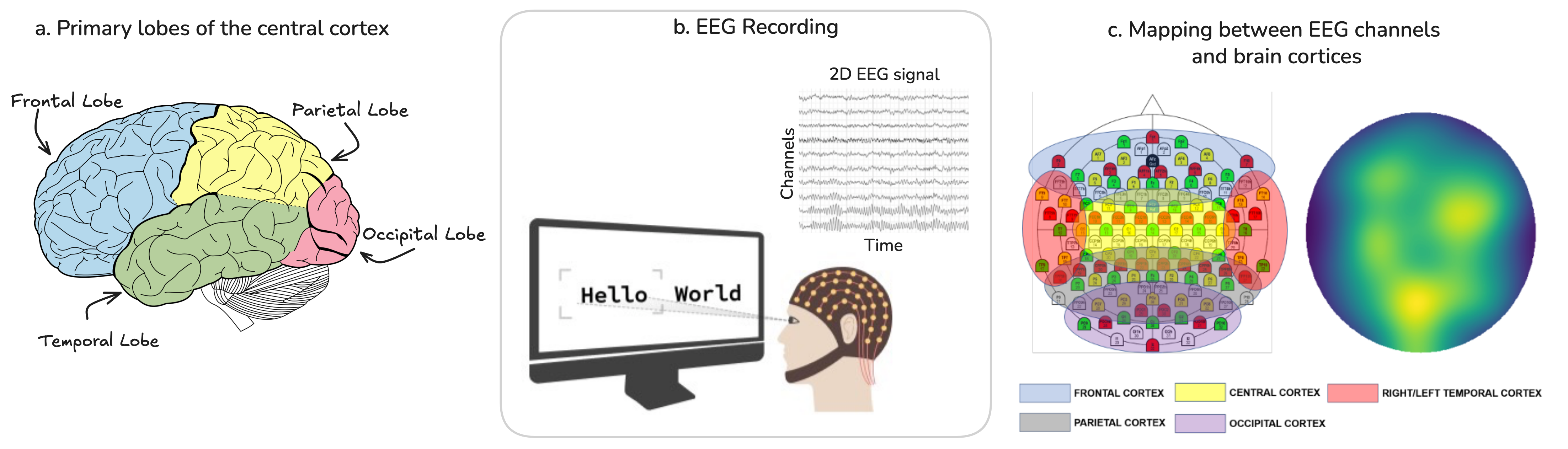}}
   \captionof{figure}{(a) Diagram of the primary lobes of cerebral cortex, including frontal, parietal, temporal, and occipital, highlighting their anatomical boundaries.(b) EEG Recording: Illustration of EEG activity recorded while participants view text stimuli, showing eye-gaze position and a 2-dimensional representation of the corresponding EEG signals. (c) Mapping between EEG channels and brain cortices: On the left is the visualization of each electrode in the 128-Channel EEG electrode placement mapped to a specific cortical region, with mappings shown across frontal, central, temporal, parietal, and occipital cortices. On the right is the neural activation visualization taken from the top of the scalp \citep{palazzo2020decoding}.}
	\label{fig:eeg_recording}
\end{figure*}

\begin{figure*}[!t]
     \centering
    {\includegraphics[width=0.89\textwidth]{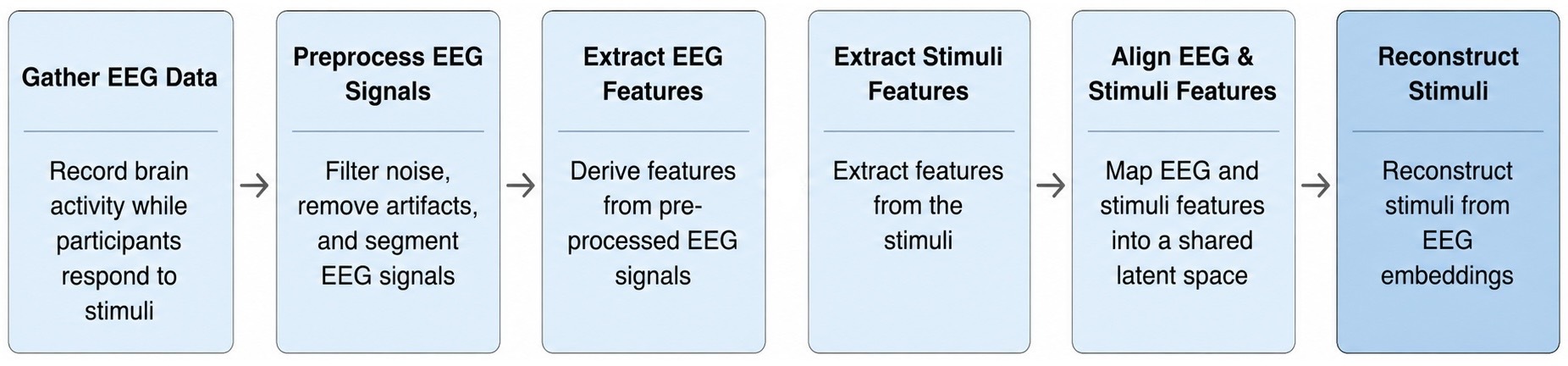}}
   \captionof{figure}{\textbf{General pipeline for EEG-based generative modeling}. EEG signals are collected while participants respond to external stimuli, preprocessed to remove artifacts and noise, and transformed into feature representations. These features are aligned with stimulus representations in a shared latent space, enabling the reconstruction of images, text, or audio from EEG embeddings.}
	\label{fig:overall}
\end{figure*}

\begin{figure*}[t]
     \centering
{\includegraphics[width=0.90\textwidth]{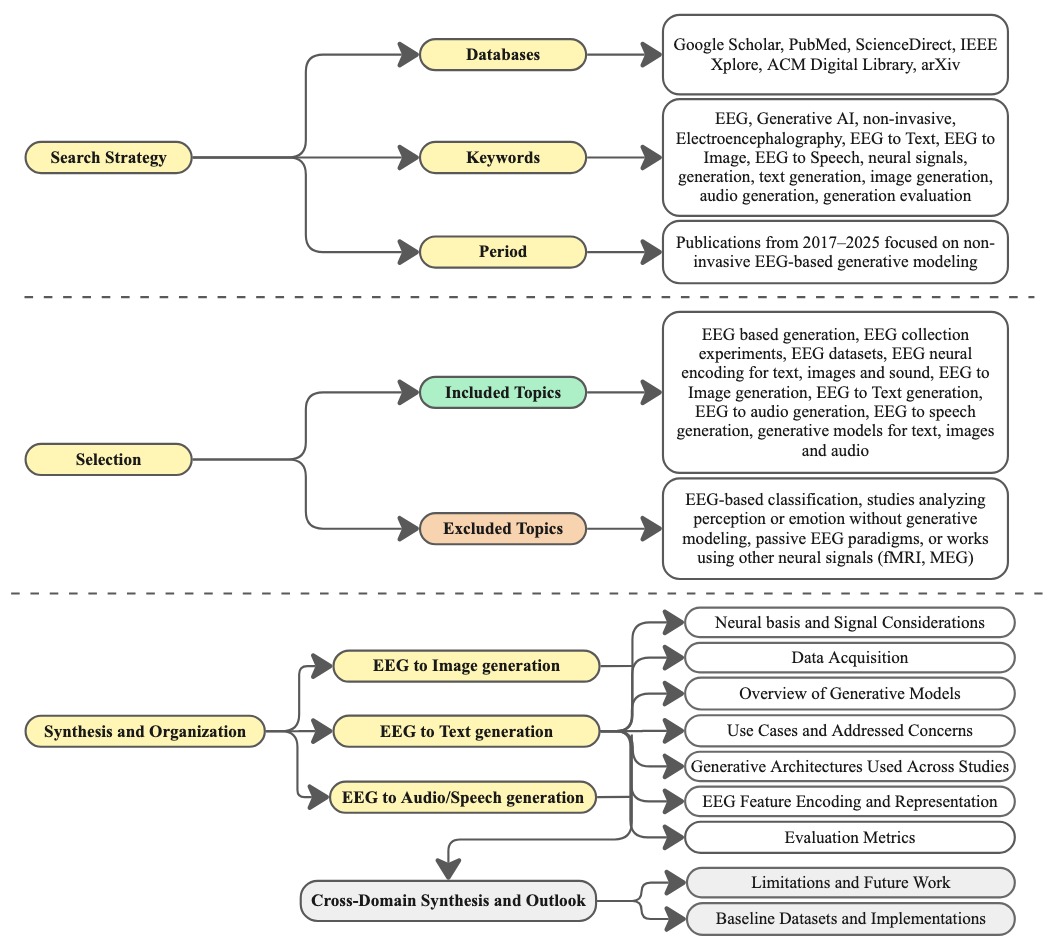}}
   \captionof{figure}{Overview of the literature search, selection, and synthesis framework for this topical review.
The upper tiers summarize the search strategy and inclusion criteria for EEG-based generative modeling studies. The synthesis and organization tier maps to modality-specific sections (EEG-to-image, EEG-to-text, and EEG-to-audio/speech generation), each analyzed through dimensions such as neural basis, generative architectures, challenges tackled by studies, and evaluation methods. The cross-domain synthesis and outlook section integrates future research directions and suggested baseline implementations for practitioners.}
	\label{fig:method_flowchart}
\end{figure*}

The convergence of Brain-Computer Interfaces (BCIs) and Generative Artificial Intelligence (GenAI) is opening a transformative path toward direct brain-to-device communication. Advances in human-computer interaction have already supported applications in assistive communication for individuals with disabilities, cognitive neuroscience, mental health assessment, augmented and virtual reality (AR/VR), and neural art generation. Although invasive and minimally invasive BCIs are beginning to show real viability for industrial deployment, non-invasive systems remain comparatively underdeveloped and are still mostly confined to academic research and prototypes. Their broader adoption is constrained by the lower spatial and temporal resolution of different non-invasive signals, which presents significant challenges. These constraints highlight the need to examine recent progress in generative methods for non-invasive neural data in order to clarify emerging methodologies, identify current limitations, and opportunities for advancement in this domain.

Among non-invasive neural recording techniques such as functional Magnetic Resonance Imaging (fMRI), Electroencephalography (EEG), and Magnetoencephalography (MEG), this survey primarily focuses on EEG. EEG measures the brain's electrical activity from the scalp, providing high temporal resolution that enables detection of rapid neural changes at the millisecond scale. Figure~\ref{fig:eeg_recording} illustrates the EEG acquisition process: electrical signals generated by neural activity are captured using headsets equipped with 12-256 electrodes (or channels) while participants are exposed to external stimuli such as images, text, or audio. These electrodes are spatially mapped to specific cortical regions, enabling localized observation of brain responses. While EEG has lower spatial resolution compared to fMRI or MEG, numerous studies demonstrate that it can still capture distinct activity patterns across different cortical regions in response to various stimuli \citep{lutzenberger1995visual, pfurtscheller1994event, khadir2023brain, bastiaansen2008see, marinkovic2004spatiotemporal}. EEG serves as a key neuro-physiological tool for assessing brain activity, particularly within the cerebral cortex, which is organized into four major lobes: \textbf{frontal, temporal, parietal, and occipital}, shown in Figure~\ref{fig:eeg_recording}(a). The frontal lobe is associated with higher cognitive functions and decision-making; the temporal lobe plays a primary role in auditory processing and multimodal sensory integration; the parietal lobe is involved in attention and language-related processes; and the occipital lobe processes visual information.

EEG signals are inherently \textbf{spatiotemporal}, consisting of a temporal component that reflects dynamic neural oscillations and a spatial component defined by the configuration of electrodes across the scalp. Typical EEG amplitudes range from 2--500~\textmu V and frequencies from 1--100~Hz. These signals are conventionally grouped into five frequency bands: $\delta$ (delta, 0.5--4~Hz), $\theta$ (theta, 4--7~Hz), $\alpha$ (alpha, 8--12~Hz), $\beta$ (beta, 13--30~Hz), and $\gamma$ (gamma, 30--100~Hz). Throughout this survey, we use these symbols and band names interchangeably.
    
Traditionally, researchers have investigated the synchronization and desynchronization of EEG rhythms across these bands to understand how neural activity changes in response to cognitive engagement or external stimuli \citep{pfurtscheller1994event, krause1996event, krause1997relative}. Synchronization occurs when oscillations across cortical regions become more coordinated and rhythmic, often reflecting an idle or resting state with reduced information processing. Desynchronization, in contrast, reflects a reduction in rhythmic coherence and is typically associated with active cortical engagement during sensory or cognitive tasks. These changes in oscillatory activity across frequency bands provide a window into large-scale neural dynamics in response to stimuli \citep{marinkovic2004spatiotemporal, bastiaansen2008see, weiss2005increased, lutzenberger1995visual, pfurtscheller1994event}, and may contain decodable traces of perceptual, cognitive, and semantic representations.

\textbf{We hypothesize that frequency-specific and region-specific oscillatory patterns enrich the information content available in EEG signals}, thereby enabling EEG-to-media generation models to decode and reconstruct underlying perceptual, cognitive, and semantic representations. Building on this hypothesis, and given that EEG supports both passive and active Brain-Computer Interface (BCI) paradigms, it holds significant potential for \textbf{adaptive human-computer interaction} \citep{zander2010enhancing, wolpaw2010brain}.

In light of recent breakthroughs in Generative AI, this survey presents a comprehensive review of advances in \textbf{EEG-based cross-modal generation}, focusing on two primary directions. The first is \textbf{EEG-to-image generation}, which involves the generation and reconstruction of visual stimuli from brain signals by leveraging models such as \textit{Generative Adversarial Networks (GANs) and Diffusion Models} \citep{goodfellow2020generative, ho2020denoising} to decode visual perception. The second direction explores \textbf{EEG-to-text translation}, where \textit{Recurrent Neural Networks (RNNs) and Transformer-based language models} \citep{vaswani2017attention} are employed to learn and generate linguistic representations from neural activity. Beyond these two main lines of research, this survey also discusses emerging efforts in \textbf{audio/speech decoding from EEG signals} and multimodal integration, where the stimuli presented to subjects and the generated media belong to different modalities. 

We highlight in Figure~\ref{fig:overall} a high-level overview of the EEG-to-stimuli generation pipeline, illustrating the \textbf{key stages from neural data acquisition to the generation of text, images, or audio}, serving as a conceptual starting point for the survey. The survey is organized into three main sections: EEG-to-Image, EEG-to-Text, and EEG-to-Audio/Speech generation, with each section progressing in a linear and structured manner. Because survey papers can be conceptually dense, \textbf{we encourage readers to follow the sections in order for a coherent understanding of the field}. We first outline the neural basis of EEG responses to different modalities, describing the relevant cortical regions, how EEG captures their activity, and the dominant frequency bands involved. Next, we summarize data collection procedure and highlight the characteristics of datasets used across the literature. We then provide an overview of foundational Generative AI methodologies to establish the technological context for interpreting the studies that follow. The modality-specific sections then review the use cases examined in prior work, the challenges addressed, the generative architectures, EEG feature-encoding techniques employed, and the reported evaluation metrics.

Finally, we discuss the limitations identified across studies, propose future research directions, and highlight baseline datasets and implementations in \ref{appendix:baseline} that can support reproducibility and enable systematic tracking of progress in the field. By consolidating these developments, this survey aims to provide researchers and practitioners with a cohesive understanding of EEG-based generative AI and its potential to expand the frontiers of \textbf{brain-computer interaction}. 
\section{Related Work}
\label{sec:related_work}

The growing application of generative AI to EEG-based media generation and brain–computer interfaces (BCIs) has led to several surveys examining different aspects of this rapidly evolving field. These reviews can be broadly divided into two groups: those focusing on \textbf{EEG-based media generation} and those examining the wider \textbf{role of generative AI in EEG based BCI systems}, including data augmentation, sensor design and optimization, and performance enhancement across a range of neural decoding tasks.

Early work by \citet{cao2020review} provided a broad overview of artificial intelligence in EEG-based BCIs, covering machine learning and deep learning methods for monitoring and providing feedback on cognitive states, as well as BCI-related applications in computer vision, natural language processing, and robotic control. More directly related to media generation, \citet{sabharwal2024comprehensive} conducted a systematic review of EEG-to-output research. Their study summarized generative architectures such as GANs, VAEs, and transformers, along with commonly used evaluation metrics and challenges, including the lack of standardized datasets for EEG-based image, video, and audio generation.

Other surveys have examined have examined the broader application of generative AI in EEG-based BCIs. \citet{eldawlatly2024role} reviewed its use for EEG data augmentation, improving the spatiotemporal resolution of EEG recordings, and enhancing cross-subject generalization. Focusing specifically on generative adversarial networks, \citet{habashi2023generative} surveyed the use of GANs to improve BCI performance across paradigms such as motor imagery, P300-based systems, emotion recognition, and epileptic seizure detection and prediction. From a causal perspective, \citet{barbera2026using} discussed AI-based EEG–BCI systems in relation to distribution shifts, data acquisition, data augmentation, and subject-invariant modeling. Their review also highlighted the emergence of EEG foundation models, limitations caused by data scarcity and the lack of interventional datasets, and directions for addressing current methodological and technological gaps. Extending this broader perspective, \citet{wang2026generative} surveyed generative AI for BCIs across both invasive and non-invasive neural signals. Their review covered neural decoding, sensor design and optimization, data augmentation, fairness, cross-subject generalization, and broader security, privacy, and ethical concerns.

Although these surveys provide valuable perspectives on AI-enabled BCIs and selected forms of EEG-based generation, \textbf{a comprehensive synthesis centered specifically on EEG-based generative media} remains limited. Our work addresses this gap by reviewing EEG-to-image, EEG-to-text, and EEG-to-audio generation within a unified framework. In addition to summarizing the datasets, generative architectures, EEG feature extraction methods, and evaluation methodologies used across these modalities, we compare shared and modality-specific methodological trends, identify persistent limitations, discuss ethical and security considerations, and highlight future research directions for advancing EEG-based generative AI.
\section{Method}
\label{sec:method}

To capture the breadth of work at the intersection of EEG and generative AI, we conducted a literature search across multiple indexing services, including Google Scholar, PubMed, ScienceDirect, IEEE Xplore, ACM Digital Library, and arXiv. Search strings combined terms such as \textit{“EEG,” “Electroencephalography,” “text generation,” “image reconstruction,” “speech synthesis,” and “generative AI.”} These terms were iteratively refined to capture studies across different modalities and architectures. In Figure~\ref{fig:method_flowchart}, we present a flowchart outlining the search strategy, selection criteria, and organizational synthesis of the reviewed studies and associated technologies. Our primary focus was on work published between 2017 and 2025, reflecting the period during which deep learning and generative modeling have driven most advances in this field. Studies were included if they used non-invasive EEG signals and applied machine learning or generative AI approaches to produce cross-modal outputs such as text, images, or audio. We considered both peer-reviewed publications and influential preprints. Excluded works included those that focused exclusively on classification tasks (e.g., sentiment analysis, motor imagery recognition, emotion recognition), used other non-invasive techniques like functional magnetic resonance imaging (fMRI), magnetoencephalography (MEG) or invasive neural recordings as these fall outside the scope of EEG-based generative modeling.

For each selected study, we compiled structured notes in tabular form, which had the investigated modality (image, text, audio or speech), study use case, addressed challenges, model architectures employed by studies (e.g., GANs, VAEs, transformers, diffusion models), EEG feature encoding strategies, datasets used and reported evaluation metrics. These tables provided a consistent framework to compare studies and enabled us to identify patterns across the literature. We then grouped studies into three major categories of EEG-driven generation-image, text, and audio-and within each modality, analyzed use cases, architectures, encoding techniques, and evaluation practices. This organization allowed us to highlight both common challenges and emerging trends.

Additionally, for the \textbf{foundational knowledge} required to interpret the reviewed studies, we also include background on the neural basis of EEG responses to different stimuli and an overview of generative methods used across modalities. Our goal overall is to provide a balanced, comprehensive and structured overview, ensuring that readers can follow how different approaches have been applied, what challenges they help overcome, existing limitations, and the future direction for further progress in this field.

\begin{table*}[!t]
    \centering
    \scriptsize
    \renewcommand{\arraystretch}{1.5} 
    \resizebox{\textwidth}{!}{%
     \begin{tabular}{>{\centering\arraybackslash}p{0.1\linewidth}> {\centering\arraybackslash}p{0.15\linewidth}> {\centering\arraybackslash}p{0.1\linewidth}>{\centering\arraybackslash}p{0.1\linewidth}> {\centering\arraybackslash}p{0.25\linewidth}> {\centering\arraybackslash}p{0.1\linewidth}> {\centering\arraybackslash}p{0.05\linewidth}> {\centering\arraybackslash}p{0.15\linewidth}}
        \toprule
        \textbf{Dataset} & \textbf{Reference} & \textbf{Stimuli Type} & 
        \textbf{Channels /Electrodes} & \textbf{Stimuli Details} & \textbf{Sampling rate} & \textbf{Subjects} & \textbf{Availability and License}\\
        \midrule
        Zuco 1.0 & \cite{hollenstein2018zuco} & Text & 128 & Sentences from Stanford Sentiment Treebank, Wikipedia corpus & 500 Hz & 12 & Public (OSF; CC BY 4.0 International)\\
        \cmidrule(lr){1-8}
        Zuco 2.0 & \cite{hollenstein2019zuco} & Text & 128 & Expanded subjects with similar content as Zuco 1.0 & 500 hz & 18 & Public (OSF; CC BY 4.0 International)\\
        \cmidrule(lr){1-8}
        Envisioned Speech & \cite{kumar2018envisioned} & Imagined Speech & 14 & 20 text stimuli (digits, characters), 10 objects &  2048 Hz (downsampled to 128 Hz) & 23 & On request, public mirror accessible on Kaggle \\
        \cmidrule(lr){1-8}
        Chisco & \cite{zhang2024chisco} & Text, Imagined Speech & 128 & 6,681 sentences of daily Chinese expressions & 1,000Hz & 3 & Public (OpenNeuro; CC0 1.0) \\
        \cmidrule(lr){1-8}
        ChineseEEG & \cite{lu2025eeg2text} & Text & 128 & 2 Chinese novels, each character shown 350 ms without punctuation & 250 Hz (Reading Aloud) 1,000 Hz (Passive Listening) & 10 & Public (OpenNeuro; CC0 1.0) \\
        \cmidrule(lr){1-8}
        Neural Spelling & \cite{jiang2025neural} & Text/Handwriting & 64 & 26 letters, tablet stylus; 25 reps/letter (650 trials) &  1,000 Hz & 28 & On request \\
        \cmidrule(lr){1-8}
        OCED & \cite{kaneshiro2015representational} & Image & 128 & 12 images per 6 object categories & 1,000 Hz (downsampled to 62.5 Hz) & 10 & Pubic (Stanford Digital Repository; CC BY 3.0)\\
        \cmidrule(lr){1-8}
        ImageNet EEG & \cite{spampinato2017deep} & Image & 128 & 40 ImageNet classes, 50 images/class, 2000 total & 1,000 Hz & 6 & Public  (GitHub)\\
        \cmidrule(lr){1-8}
        Texture Perception & \cite{orima2021analysis} & Image & 19 & 166 grayscale natural texture images & 1,000 Hz & 15 & On request\\
        \cmidrule(lr){1-8}
        THINGS-EEG & \cite{grootswagers2022human} & Image & 64 & 22,248 images across 1854 object concepts & 1,000 Hz & 50 & Public (OSF; CC BY 4.0)\\
        \cmidrule(lr){1-8}
        DCAE dataset & \cite{zeng2023dcae} & Image & 32 & 200 ImageNet images (cats, dogs, flowers, pandas) & 128 Hz & 26 & Unavailable (confidential) \\
        \cmidrule(lr){1-8}
        Alljoined & \cite{xu2024alljoined} & Image & 64 & 10,000 images per participant from 80 MS-COCO categories & 512 Hz & 8 & Previously public; current access unavailable\\
        \cmidrule(lr){1-8}
        Low-Density EEG Reconstruction & \cite{guenther2024image} & Image & 8 & 600 images/session (20 classes, 19 ImageNet + Faces), 2s on-screen, random shuffling & 250 Hz & 9 & On request\\
        \cmidrule(lr){1-8}
        Alljoined-1.6M & \cite{xu2025alljoined} & Image & 32 &  26,000 high-resolution photographs spanning
1,854 everyday object categories & 256 Hz & 20 & Public (HuggingFace; CC BY-NC-SA 4.0)\\
        \cmidrule(lr){1-8}
        ThoughtViz & \cite{tirupattur2018thoughtviz} & Imagined Objects & 14 & EEG recorded while participants imagined digits, characters, and objects & 128 & 23 & Public (Github)\\
        \cmidrule(lr){1-8}
        KARA ONE & \cite{zhao2015classifying} & Text, Audio, Speech & 64 & Rest state, stimulus, imagined speech, speaking task & 1,000 Hz & 12 & Public (University of Toronto) \\
        \cmidrule(lr){1-8}
        NMED-H & \cite{kaneshiro2016naturalistic} & Music & 125 & 4 versions of 4 songs, total 16 stimuli & 1,000 Hz (downsampled to 125 Hz)& 48 & Public (Stanford Digital Repository; CC BY 3.0)\\
        \cmidrule(lr){1-8}
        NMED-T & \cite{losorelli2017nmed} & Music & 128 & 10 songs (4:30-5:00 mins) with tempos 56-150 BPM & 1,000 Hz (downsampled to 125 Hz) & 20 & Public (Stanford Digital Repository; CC BY 3.0)\\
        \cmidrule(lr){1-8}
        Narrative Speech & \cite{broderick2019semantic} & Audio & 128 & Audio book of \textit{The old man and the sea by Hemingway} & 512 Hz (downsampled to 128 Hz)& 19 & Public (Dryad; CC0 1.0)\\
        \cmidrule(lr){1-8}
        Alice & \cite{bhattasali2020alice} & Audio & 61 + 1 ground & 2,129 words, 84 sentences from \textit{Alice in Wonderland} & 500 Hz & 52 & Public (OpenNeuro; CC0 1.0) \\
        \cmidrule(lr){1-8}
        N400 & \cite{toffolo2022evoking} & Audio & 128 & 402 sentences, each containing 5-8 words contained in \textit{Medical Research Council Psycholinguistic (MRCP)
database} & 512 Hz & 20 & Public (Dryad; CC0 1.0)\\
        \cmidrule(lr){1-8}
        Japanese Speech EEG & \cite{mizuno2024investigation} & Audio & 64 & 503 spoken sentences (male/female speaker) & 8,192 Hz (downsampled to 1,024 Hz)& 1 & Not publicly available/specified\\
        \cmidrule(lr){1-8}
        Phrase/Word Speech EEG & \cite{park2024towards} & Audio & 64 & Audio of 13 words/phrases, followed by speech replication & 2,500 Hz & 10 & Not publicly available/specified \\
        \bottomrule
    \end{tabular}%
    }
    \caption{EEG-based datasets from surveyed studies using text, image, and audio/speech/music stimuli. Public availability and licensing are reported where specified. CC BY allows reuse with credit; CC BY-NC-SA allows noncommercial reuse with credit under the same license; CC0 allows reuse without restrictions.}
    \label{tab:datasets}
\end{table*}

\section{EEG-to-Image Generation}
\label{sec:eeg_image}

\subsection{Neural Basis and Signal Considerations}
\label{subsec:neural_basis_image}

In EEG-based generation studies, the underlying neural basis is often overlooked, which we believe is essential to motivate how perceptual, semantic, and auditory information can be reconstructed from the recorded signals. 

EEG has long been used as a tool to study visual perception. An early study by \cite{lutzenberger1995visual} investigates the effect of visual field presentation by changing the stimulus position in the upper versus lower visual field. The authors found that coherent stimuli in the upper visual field, which is processed by the lower half of the retina and thus the more ventral (lower) sites of the primary visual cortex, elicited a \textbf{40 Hz spectral power enhancement} over \textbf{lower occipital electrodes}. Conversely, stimuli presented in the lower visual field produced a corresponding enhancement over upper occipital electrodes. These findings demonstrate that EEG can reliably  capture location-specific cortical responses to visual stimuli.

Beyond this, EEG has been extensively used to understand the broader dynamics of visual perception. Cognitive activity in response to visual stimuli unfolds through successive stages, including primary visual processing, feature extraction, higher-level cognitive analysis, and attentional modulation. Although early visual processing is primarily associated with occipital regions, EEG studies consistently show that visual processing engages a broader network spanning occipito-parietal, and frontal areas as information progresses from low-level sensory analysis to higher-order cognitive interpretation. To understand the \textbf{timing} (relative to stimulus onset) and \textbf{spatial distribution} (across cortical regions) of these processes, researchers commonly examine \textbf{changes in oscillatory activity across EEG frequency bands}, such as event-related desynchronization (ERD), along with visual evoked potentials (VEPs), which reflect electrical responses originating in the visual cortex. 

ERD in the \textbf{alpha band} serves as a well-established marker of cortical activation: while high alpha amplitude is characteristic of a resting or “idle” state, a reduction in alpha power (desynchronization) indicates a change from a resting to an activated state. A foundational study by \cite{pfurtscheller1994event} analyzed ERD in alpha sub-bands during visual processing and identified two distinct components: (i) \textbf{a short-lasting upper-alpha(10-12 Hz) desynchronization} localized over \textbf{occipital regions}, and (ii) a \textbf{longer-lasting lower-alpha (6-8 Hz) desynchronization} that is more broadly distributed, with maximal effects over \textbf{parietal areas}. The occipital ERD component peaks approximately \textbf{200-300 ms after stimulus onset}, whereas the parietal component reaches its maximum around 300-500 ms, reflecting the progression from early visual analysis to broader cognitive engagement. These temporal and spatial patterns, observed through EEG, provide a useful way to \textbf{trace how visual information propagates through cortical networks.}

Cross-frequency interactions also play a role in visual perception. Another study by \cite{demiralp2007gamma} examined the interaction between theta and gamma oscillations during visual perception using EEG. They found that event-related gamma activity was strongly dependent on the phase of concurrent theta oscillations, indicating that \textbf{theta rhythms modulate gamma-band responses}. This cross-frequency coupling is believed to play an important role in perceptual and cognitive processes, including \textbf{visual perception}. In their findings, gamma activity was primarily localized over the occipital cortex, whereas theta activity was more broadly distributed, extending into frontal regions, highlighting the \textbf{integration of localized sensory processing in response to visual stimuli, with more widespread cognitive networks}.

EEG studies have also examined how oscillatory activity varies with \textbf{color perception}. \cite{khadir2023brain} analyzed theta, alpha, beta, and gamma oscillations across occipital, occipito-parietal, and prefrontal regions and identified distinct spectral patterns associated with different colors. Green stimuli elicited \textbf{slower beta-band oscillations} over occipital areas relative to red and blue stimuli, while blue stimuli showed a \textbf{decrease in theta-band power} during the late post-stimulus period in \textbf{occipital} electrodes. The authors also reported \textbf{increased phase consistency} across trials for green stimuli, in contrast to a decrease for blue stimuli. These findings demonstrate that \textbf{color information is represented through frequency-specific and region-specific oscillatory patterns}, highlighting the potential of EEG signals to support the decoding of fine-grained visual attributes.

Collectively, these studies show that visual perception is represented in distributed, frequency-specific oscillatory patterns that evolve across cortical regions over time. This spatiotemporal structure forms the neural basis upon which EEG-to-image decoding methods can operate.

\subsection{Data Acquisition}
\label{subsec:eeg2imagedata}

We highlight key EEG datasets captured for image stimuli in Table~\ref{tab:datasets}. Here, we include a key dataset to highlight relevant factors like the number and demographics of subjects, the nature of visual tasks, the image corpus, EEG recording setup, and the total size of the collected data.  

\textbf{Dataset experimental details:} EEG-Based Visual Classification Dataset by \citet{spampinato2017deep},  has widely been used for EEG-to-image generation. The dataset was collected from \textbf{six healthy subjects} who were presented with visual stimuli drawn from a subset of ImageNet, comprising \textbf{2,000 images} (50 per class across 40 object categories). Each image was displayed for 0.5 seconds, with bursts of 25 seconds followed by a 10-second pause on a black screen. The entire experiment lasted 1,400 seconds (approximately 23 minutes). EEG was recorded using a \textbf{128-channel} actiCAP system with active, low-impedance electrodes, yielding \textbf{12,000 EEG sequences} (2,000 per subject across 6 participants). The dataset includes data already filtered in\textbf{ three frequency ranges: 14-70Hz, 5-95Hz and 55-95Hz}.

This dataset has since been used in multiple studies on EEG-based image classification and visual stimulus reconstruction, making it one of the most widely used benchmarks for EEG-to-image generation studies. 

\subsection{Overview of Generative Models and Techniques for Image Generation}
\label{subsec:overview_image_gen}

In this section, we review some key generative models and machine learning techniques for image generation as a preliminary foundation before discussing their application in the EEG-image generation domain.

\begin{figure}[htbp]
    \centering
    \includegraphics[width=\linewidth]{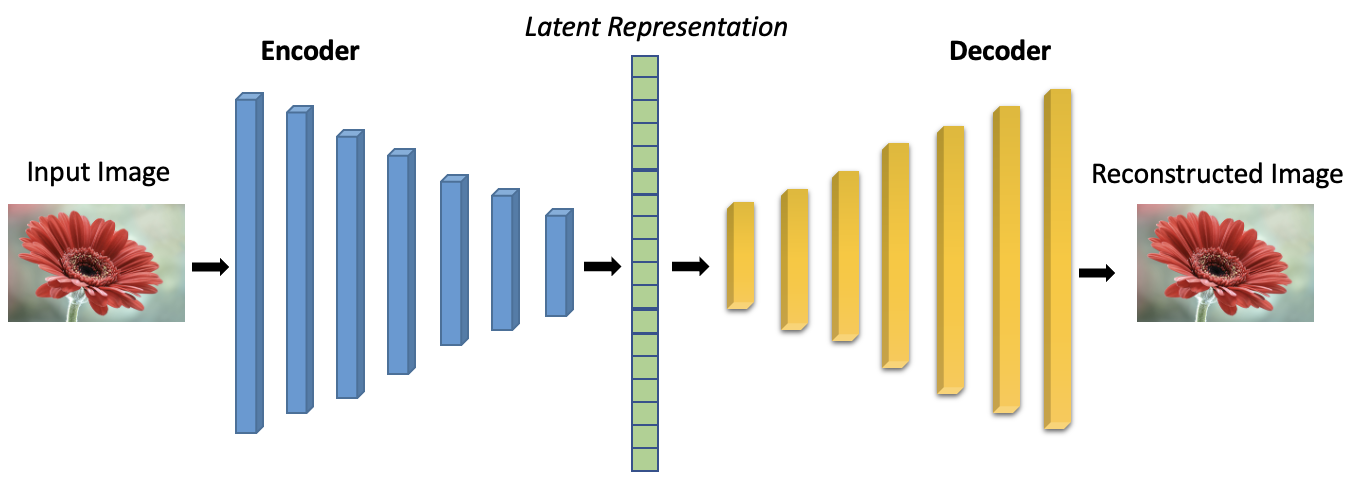}
    \caption{Traditional CNN-based encoder-decoder architecture. The encoder processes a high-dimensional input image and generates a lower-dimensional latent representation capturing the most important features of the data. The decoder then reconstructs the image from this latent representation. Model parameters are updated by minimizing the reconstruction loss between the original and reconstructed images.}
    \label{fig:image-encoder-decoder}
\end{figure}

\begin{figure}[!htbp]
    \centering
    \includegraphics[width=\linewidth]{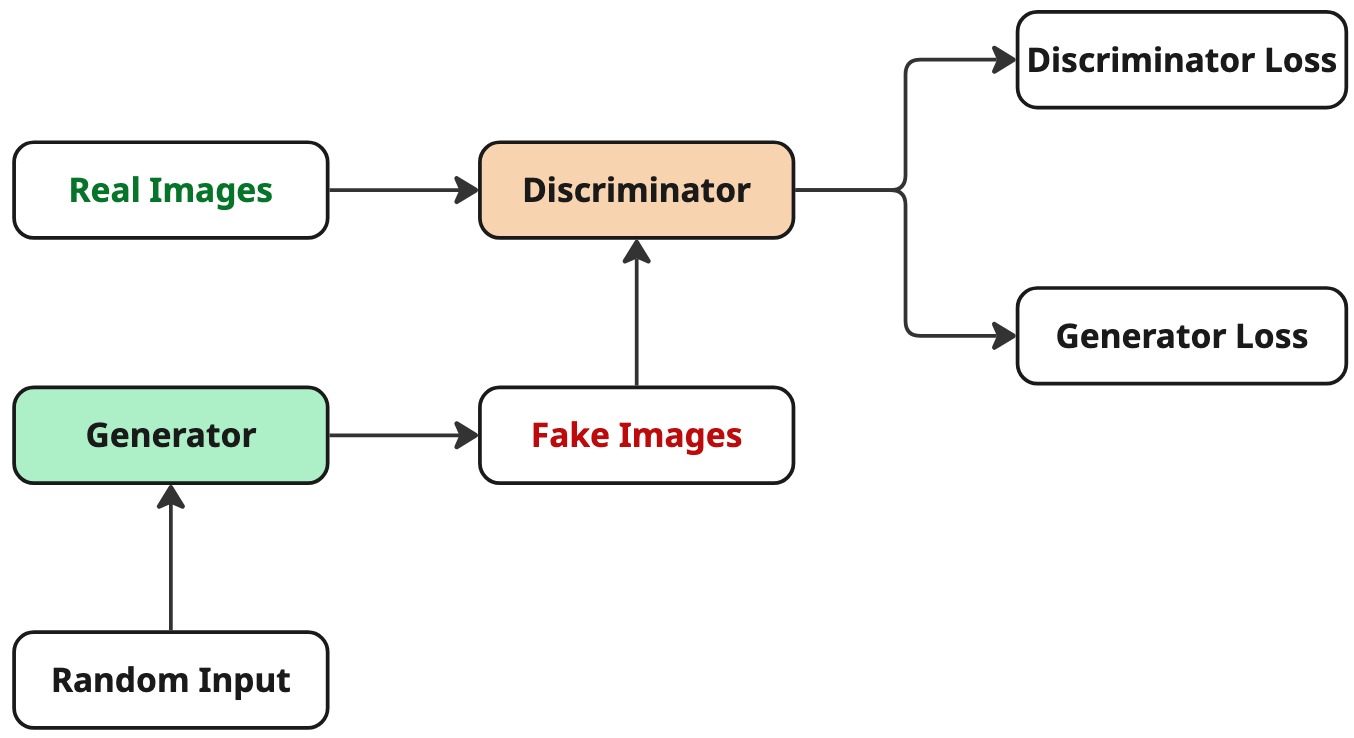}
    \caption{Structure of a Generative Adversarial Network (GAN). The model is trained through an adversarial process, where the generator creates fake images to fool the discriminator, while the discriminator learns to distinguish between real and generated images.}
    \label{fig:image-gan}
\end{figure}

A \textbf{traditional encoder-decoder architecture}, as shown in Figure~\ref{fig:image-encoder-decoder}, is a neural network framework in which the encoder takes an input and compresses it into a lower-dimensional latent representation (latent space). This concept of \textbf{dimensionality reduction} is what makes these architectures extremely useful in different domains. This process forces the model to learn the most salient and informative features (latent variables) of the input data rather than modeling every fine-grained, high-dimensional detail. The decoder then takes this latent representation and attempts to reconstruct the original input. In the context of image generation from EEG signals, this becomes a \textbf{cross-modal encoder-decoder setup}. The training data consists of EEG-image pairs, where the encoder learns to map EEG signals into a shared latent space, and the decoder learns to reconstruct the corresponding image from that representation. During training, the model \textbf{minimizes a reconstruction loss}, a measure of how different the generated image is from the original one. Minimizing this loss enables the model to \textbf{align EEG features with visual representations} in the latent space, allowing it to generate images that reflect brain signal patterns.

While encoder-decoder models are deterministic, \textbf{Variational Autoencoders (VAEs)} \citep{kingma2019introduction} are probabilistic, encoding a continuous representation of the latent space. This allows them not only to reconstruct the input but also to generate new samples similar to the original data using variational inference. Instead of fixed latent variables, VAEs learn a distribution (mean and variance) over the latent variables and generate outputs by sampling from it. The latent space should be continuous (nearby points yield similar decoded content) and complete (every point produces meaningful output). To ensure this, the latent space is regularized to follow a standard normal (Gaussian) distribution using the Kullback-Leibler (KL divergence), which minimizes the difference between the learned and target distributions. The model is trained with KL divergence loss in addition to reconstruction loss. \textbf{The probabilistic nature of VAEs might be particularly useful for EEG data, which often has a low signal-to-noise ratio.}

Another model architecture used extensively for image generation is \textbf{Generative Adversarial Network (GAN)} \citep{goodfellow2020generative}, which consist of two components - a generator and a discriminator. They are trained through an \textbf{adversarial process}, where the generator produces images from random noise and attempts to fool the discriminator into classifying them as real. The discriminator, in turn, evaluates both the generated images and real samples from the training data to determine whether each is real or fake, thereby improving its ability to distinguish between them. The generator is trained using a generator loss, which measures how successfully it can deceive the discriminator; a lower generator loss indicates more realistic outputs. The discriminator loss measures how accurately the discriminator can distinguish real from fake data; a lower discriminator loss means it is effectively identifying generated samples. The outline of the adversarial process is illustrated in Figure~\ref{fig:image-gan}. Through this adversarial training, both networks improve simultaneously, enabling the generator to produce increasingly realistic images over time.

\begin{figure}[!htbp]
    \centering
    \includegraphics[width=\linewidth]{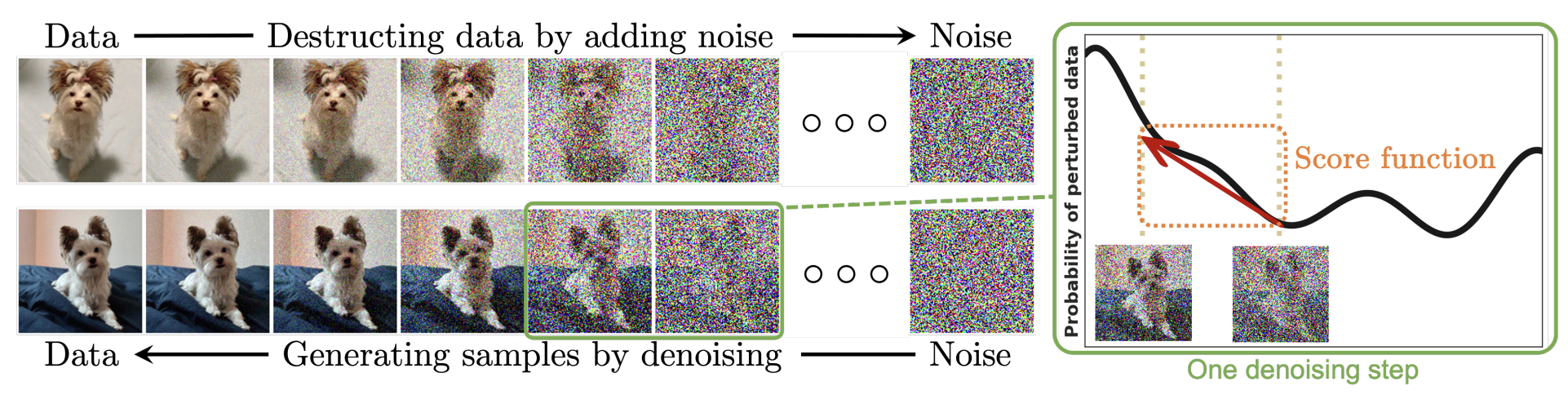}
    \caption{Probabilistic diffusion process, where data is gradually diffused into noise during the forward process, and the model learns to reconstruct data from noise by reversing these steps in the reverse diffusion process. \citep{yang2023diffusion}}
    \label{fig:image-diffusion}
\end{figure}

\textbf{Diffusion models} \citep{ho2020denoising} are another class of generative models used for image generation. Their core idea can be understood in three stages. In the forward diffusion process, an image from the training dataset is gradually transformed into pure noise, typically following a Gaussian distribution. In the reverse diffusion process, the model learns to reverse these steps, reconstructing the original image from noisy inputs, as shown in Figure~\ref{fig:image-diffusion}. During image generation, the trained model starts from random noise and progressively refines it into a high-quality image by applying the learned reverse diffusion steps. The number of steps during generation need not exactly match those used in training, since the model is trained to predict the noise added at each step rather than directly predicting the denoised image. This sampling process introduces randomness, enabling diffusion models to generate new and diverse images that closely resemble the training data. \textbf{Latent Diffusion Models (LDMs)} extend diffusion models by operating in a latent representation of the data instead of directly using high-dimensional input. This architecture employs an encoder, similar to that in a Variational Autoencoder (VAE), to first project the input image into a lower-dimensional latent space where the diffusion process is applied. The diffusion steps are thus performed more efficiently while retaining semantic quality. Stable Diffusion is a widely used implementation of this latent diffusion framework.

\begin{figure}[htbp]
    \centering
    \includegraphics[width=\linewidth]{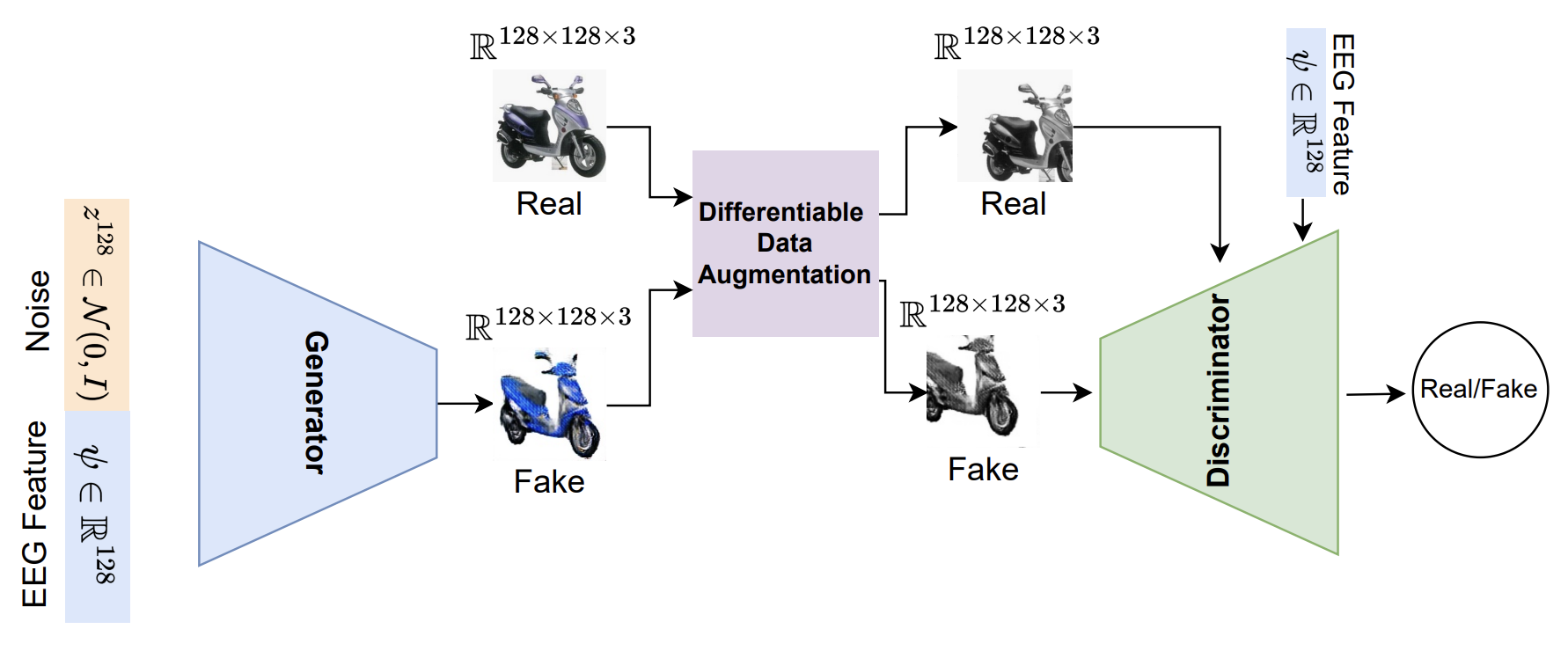}
    \caption{The Conditional GAN (cGAN) implementation from the EEG2Image framework by \citet{singh2023eeg2image} uses EEG features as conditioning inputs to guide the image generation process, enabling the model to generate images that correspond to the brain activity captured by EEG.}
    \label{fig:image-cgan}
\end{figure}

For \textbf{cross-modal generation}, several mechanisms such as contrastive learning, attention mechanisms, adversarial training, and conditional modeling have been employed. These approaches are often integrated with the primary generative architectures discussed above to enhance image generation from EEG signals. In the context of EEG, contrastive learning can help a model learn more robust representations by \textbf{distinguishing between EEG signals corresponding to the same image versus different images, or between signals from different subjects.} Vanilla VAEs and GANs have the limitation that users have no control over the outputs produced by the models. To generate specific or guided outputs, especially in cross-modal settings, \textbf{conditioning mechanisms} are introduced. By incorporating elements of supervised or semi-supervised learning, such as class labels or conditions, conditioning enables the model to generate outputs that are guided by specific inputs. In the context of EEG to Image generation, EEG features are used as conditional input, as shown in Figure~\ref{fig:image-cgan}. \textbf{Well-conditioned generators} have been shown to perform better in practice \citep{odena2018generator}. Conditioning can also be combined with adversarial objectives and \textbf{self-attention mechanisms} to improve image generation quality by modeling long-range dependencies \citep{zhang2019self} and enhancing feature consistency across modalities.

In the upcoming sections on EEG-to-image generation studies, knowledge of these fundamental generative architectures will be useful to grasp the underlying workings of proposed frameworks in this domain.

\subsection{Use Cases and Addressed Concerns}
\label{subsec:eeg_image_use_cases}

\begin{figure*}
     \centering
    {\includegraphics[width=\textwidth]{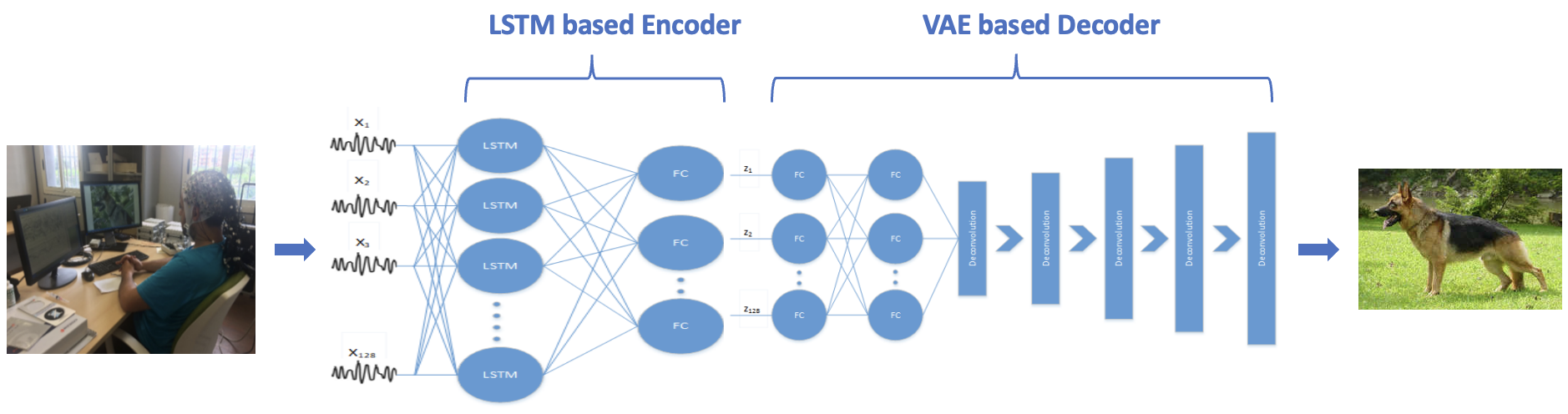}}
   \caption{Encoder-decoder framework for EEG-to-image generation. In this Brain2Image implementation, an LSTM-based encoder extracts compact EEG representations, which are then mapped to the image space by a VAE-based decoder \citep{kavasidis2017brain2image}.}
	\label{fig:brain2image}
\end{figure*}

Before discussing the technological aspects of generating images from EEG signals, we highlight the \textbf{key challenges addressed in existing studies}. A primary limitation lies in the \textbf{inherent noise of EEG recordings}, which results in a low signal-to-noise ratio \citep{bai2306dreamdiffusion, lan2023seeing, zeng2023dm} and restricts the amount of discriminative information that can be extracted from EEG signal. Mapping EEG embeddings and image embeddings into a shared latent space further compounds the difficulty, especially given \textbf{inter-subject variability in neural responses} \citep{bai2306dreamdiffusion}. Cross-modal encoder-decoder networks give \textbf{limited performance on images of natural objects} compared to simpler stimuli such as digits or characters \citep{mishra2023neurogan} and experimental constraints often lead to \textbf{small datasets} \citep{singh2023eeg2image}, limiting the robustness and generalizability of generative models. Moreover, applying convolution layers separately along temporal and spatial dimensions has been shown to disrupt inter-channel correlations, thereby \textbf{hindering the preservation of spatial properties in brain activity} \citep{song2023decoding}.

\textbf{To overcome the challenges} of low signal quality and limited discriminative power in EEG data, \citet{kavasidis2017brain2image}, \citet{song2023decoding}, and \citet{mishra2023neurogan} extract \textbf{image class-specific EEG encodings} to capture features that better distinguish between categories, capture class discriminative information and therefore enhance generation quality. To overcome the loss of temporal and spatial information, \citet{nemrodovneural} uses \textbf{spatiotemporal EEG features} to understand the neural correlates of facial identity, while \citet{khaleghi2022visual} map EEG activity to \textbf{visual saliency maps} to highlight the image regions most relevant to human perception. Other approaches focus on improving alignment between brain signals and visual data: for example, projecting EEG into a shared subspace with image embeddings \citep{shimizu2022improving} or generating \textbf{image class-specific latent representations} \citep{mishra2023neurogan}. Similarly, \citet{lan2023seeing} decodes multi-level perceptual information to produce more detailed, multi-grained outputs. To overcome the large data demands of supervised learning, \cite{li2020semi, song2023decoding} explore \textbf{self-supervised and semi-supervised training paradigms} to reduce reliance on extensive labeled datasets.   

Furthermore, recent studies have aimed to enhance the robustness and interpretability of EEG-based image generation. \cite{singh2024learning} emphasizes on improving the \textbf{generalizability of feature-extraction pipelines} across diverse datasets , \cite{sugimoto2024image} examine the \textbf{relative contribution of different EEG electrode settings} and \cite{li2024visual} explore attention mechanisms to highlight the \textbf{importance of specific channels or frequency bands}, thereby improving model performance and interpretability. 

More recent studies have shifted focus toward achieving \textbf{practical BCI implementation and enhanced model interpretability}. Addressing the need for real-world usability, \citet{guenther2024image} demonstrated that image classification and reconstruction are possible using only an \textbf{8-channel portable EEG setup}, greatly improving affordability and mobility over traditional high-density systems. For a streamlined, real-time BCI, \citet{lopez2025guess} proposed an \textbf{efficient pipeline} using minimal preprocessing and a lightweight adapter, successfully surpassing prior methods in both generation quality and semantic correctness.

To improve interpretability and semantic fidelity, \textbf{intermediate, structured semantic representations} are used as mediators between the EEG signal and the generative model. Studies like \citet{mehmood2025catvis}, \citet{rezvani2025interpretable}, and \citet{cheng2025fine} bypass direct, noisy EEG-to-image generation by aligning EEG signals with interpretable, structured semantic prompts, such as \textbf{multilevel captions} or \textbf{fine-grained visual attributes} (e.g., color and texture). This \textbf{text- or feature-mediated approach} enables cognitively aligned decoding and precise control over the image generation process.

Beyond interpretability and semantic alignment, practical deployment requires addressing challenges related to adaptation to new subjects, achieving \textbf{robust performance with affordable EEG hardware}, and enabling the large-scale deployment of computationally intensive architectures. To address scalability and subject adaptation, \citet{kneeland2026enigma} proposed ENIGMA, a multi-subject EEG-to-image decoding model with a simplified architecture requiring less than 1\% of the trainable parameters used by previous subject-specific approaches. For a 30-subject deployment, ENIGMA reduces the total parameter count by a factor of 165 compared with the subject-specific model of \citet{li2024visual}. Despite its substantially lower computational requirements, the model achieves state-of-the-art performance on the THINGS-EEG2 \citep{hebart2019things} and Alljoined-1.6M \citep{xu2025alljoined} datasets and can adapt to new subjects using as little as 15 minutes of calibration data. These findings demonstrate the potential of parameter-efficient, multi-subject architectures to support more scalable EEG-based generation. However, further research on computational efficiency, cross-subject generalization, low-cost EEG systems, and minimal-calibration adaptation remains necessary for translating EEG-based generative AI into practical real-world applications.

\subsection{Generative Architectures Used Across Studies}
\label{subsec:eeg_image_architectures}

An encoder-decoder (overview in Section~\ref{subsec:overview_image_gen}) based framework, introduced in Brain2Image \citep{kavasidis2017brain2image}, is illustrated in Figure~\ref{fig:brain2image}  to generate images from EEG signals. Since EEG is a time-series signal with both temporal dynamics and spatial structure across electrodes, an effective \textbf{encoder must capture both dimensions}. Encoders are often designed using recurrent units (e.g., LSTMs or GRUs) to model temporal dependencies, convolutional layers to capture spatial patterns across channels, or a hybrid combination of the two. Formally, given EEG signals $\mathbf{E}$ recorded during the presentation of a visual stimulus, the encoder $f_{\theta}$ maps $\mathbf{E}$ into a latent representation $\mathbf{z} = f_{\theta}(\mathbf{E})$, where $\mathbf{z}$ is compact and class-discriminative. The \textbf{decoder} then learns to generate images conditioned on the EEG representation $\mathbf{z}$ and is commonly implemented with CNN-based architectures such as GANs and VAEs, or more recently with Transformer and diffusion models. For further performance improvement, the \textbf{decoder can be conditioned on auxiliary information}, such as image class labels or additional EEG-derived features, which strengthens the alignment between neural representations and image space. 

We provide a detailed overview of these methodologies in Section~\ref{subsec:overview_image_gen} and here we outline the major approaches found in surveyed studies, describing how they work and highlighting their relevance to the task of translating EEG signals into images specifically.

\begin{itemize}

\item \textbf{Variational Autoencoders (VAEs)} are well-suited for handling noisy signals and extracting structured latent features. Early work, such as \cite{kavasidis2017brain2image, wakita2021photorealistic} employed VAEs to map EEG representations into compact embeddings that support image reconstruction tasks.

\item In the context of EEG decoding, \textbf{Generative Adversarial Networks (GANs)}, have been widely used to reconstruct visual stimuli \citep{kavasidis2017brain2image, khaleghi2022visual, mishra2023neurogan, singh2024learning, li2024visual}. Conditional GANs extend this by conditioning on image stimuli labels, thereby improving alignment between EEG inputs and generated outputs \citep{singh2023eeg2image, ahmadieh2024visual}. 
    
\item \textbf{Diffusion Models} are employed either to refine EEG embeddings into visual priors \citep{shimizu2022improving} or as pre-trained backbones, such as Stable Diffusion \citep{bai2306dreamdiffusion}. Further extensions leveraging U-Net architectures have been shown to improve reconstruction quality \citep{zeng2023dm, lan2023seeing}. The current state-of-the-art leverages pre-trained Latent Diffusion Models (LDMs), with recent advancements focusing on highly effective conditioning mechanisms: \citet{guenther2024image} introduced \textbf{double conditioning} to guide the LDM's cross-attention and time-embedding simultaneously with EEG features, while \citet{lopez2025guess} proposed using a \textbf{ControlNet adapter} trained exclusively on EEG features to condition a frozen LDM, streamlining the image generation process.

\item \textbf{Contrastive Learning} is used to \textbf{enhance discriminative feature extraction} \citep{singh2023eeg2image, lan2023seeing, song2023decoding, sugimoto2024image}, often using cosine similarity to enforce cross-modal alignment \citep{song2023decoding}.

\item \textbf{Attention Mechanisms} is used to weigh the relative importance of EEG channels or frequency bands, enhancing interpretability and spatial correlation modeling. They have been integrated into generative pipelines to improve both image quality and channel-level interpretability \citep{mishra2023neurogan, song2023decoding, li2024visual}.

\item \textbf{Hybrid Strategies} adopt combined approaches to exploit complementary strengths. For example, attention modules have been integrated into GANs \cite{mishra2023neurogan}, and diffusion models have been coupled with contrastive learning frameworks \cite{lan2023seeing}, resulting in more robust EEG-to-image generation systems. More recently, \citet{kneeland2026enigma} proposed a simplified hybrid architecture comprising a spatio-temporal CNN for EEG feature extraction, subject-wise latent alignment layers to account for inter-subject variability, an MLP projector that maps EEG representations into the CLIP embedding space, and a pretrained Stable Diffusion model for image generation.

Another major contemporary trend is \textbf{Semantic-Mediated Generation}, which aligns noisy EEG with an interpretable semantic space (like CLIP text embeddings or multilevel captions) that then condition a frozen Latent Diffusion Model (LDM). This semantic mediation approach is exemplified by:
    \begin{itemize}
        \item Systems that align EEG with \textbf{CLIP text embeddings} and use \textbf{class-guided caption re-ranking} to inject context-aware information for thought visualization \citep{mehmood2025catvis}.
        \item Models that map EEG to \textbf{multilevel captions} generated by large language models (LLMs) to enhance both the semantic alignment and the interpretability of the decoding process \citep{rezvani2025interpretable}.
        \item Frameworks that integrate both \textbf{high- and low-level semantic features} and use a conditional mechanism like FiLM (Feature-wise Linear Modulation) to achieve fine-grained control over generation \citep{cheng2025fine}.
    \end{itemize}

\begin{figure*}[t]
     \centering
    {\includegraphics[width=0.50\textwidth]{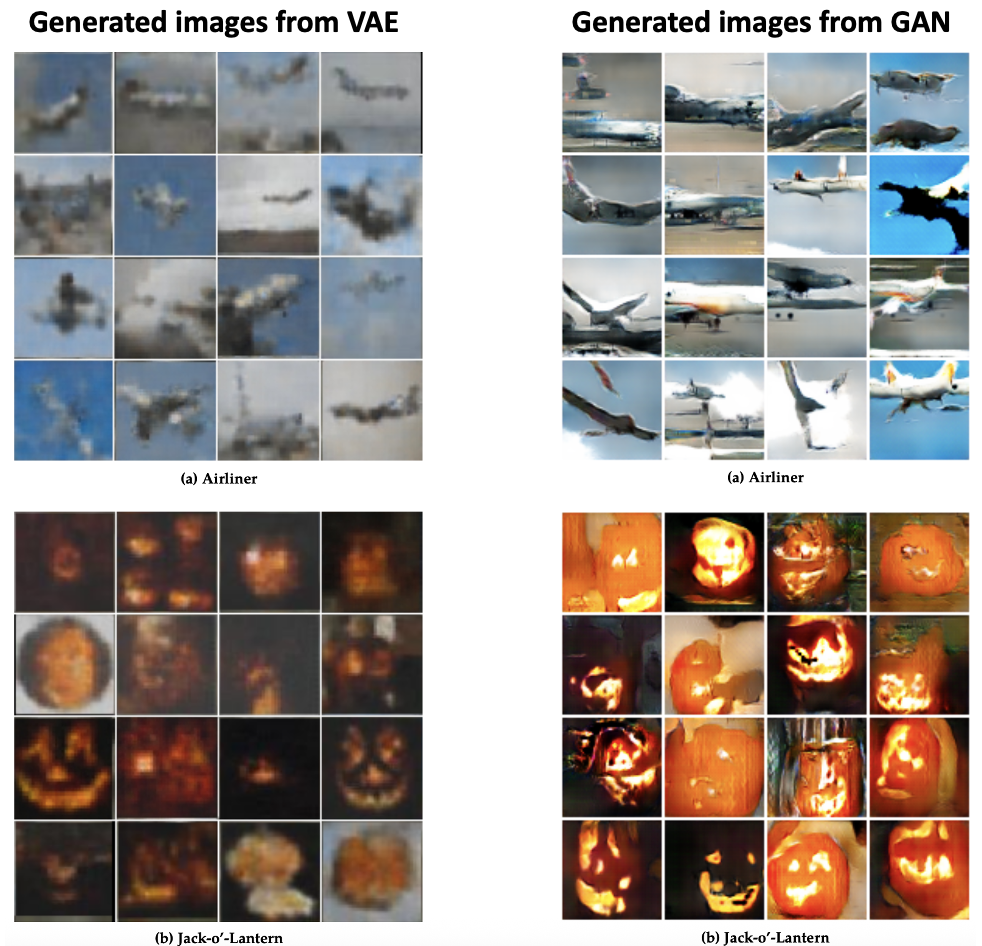}}
   \caption{Examples of images generated from EEG signals for two object categories (Airplane and Jack-o'-Lantern) using VAE- and GAN-based decoders \cite{kavasidis2017brain2image}.}
	\label{fig:eegimgresults}
\end{figure*}

\begin{table*}[t]
    \centering
    \footnotesize
    \renewcommand{\arraystretch}{1.5}
    \resizebox{\textwidth}{!}{%
    \begin{tabular}{>{\raggedright\arraybackslash}p{0.15\linewidth}>{\raggedright\arraybackslash}p{0.15\linewidth}p{0.4\textwidth}>{\raggedright\arraybackslash}p{0.3\linewidth}}
        \toprule
        \multicolumn{4}{c}{\textbf{EEG-to-Image Generation Evaluation Metrics}} \\
        \toprule
        \textbf{Category} & \textbf{Metric} & \textbf{Description / Usage in Studies} & \textbf{References} \\
        \midrule
        \multirow{1}{*}{\parbox{\linewidth}{Classification}} 
            & Top-$k$ Accuracy & Evaluates whether the correct class appears among the top-$k$ predictions, confirming EEG features are class-discriminative. & \cite{shimizu2022improving, lan2023seeing, song2023decoding} \\
        \midrule
        \multirow{5}{*}{\parbox{\linewidth}{Image quality / realism}} 
            & Inception Score (IS) & Measures image quality and diversity using a pretrained classifier; adapted from computer vision to EEG-to-image pipelines. & \cite{salimans2016improved, kavasidis2017brain2image, li2020semi, bai2306dreamdiffusion, singh2023eeg2image} \\
            \cmidrule(lr){2-4}
            & Frechet Inception Distance (FID) & Quantifies realism by comparing feature distributions of generated and real images. & \cite{bai2306dreamdiffusion, singh2024learning, ahmadieh2024visual} \\
            \cmidrule(lr){2-4}
            & Kernel Inception Distance (KID) & Computes Maximum Mean Discrepancy (MMD) between real and generated distributions; complements FID. & \cite{singh2024learning} \\
            \cmidrule(lr){2-4}
            & LPIPS & Learned perceptual similarity metric that assesses patch-level similarity between generated and reference images. & \cite{bai2306dreamdiffusion} \\
            \cmidrule(lr){2-4}
            & Diversity Score & Diversity Score evaluates the variety of generated images, providing an indication of how well the model avoids mode collapse (i.e., producing limited or repetitive outputs). & \cite{mishra2023neurogan} \\
        \midrule
        \multirow{2}{*}{\parbox{\linewidth}{Perceptual / Semantic}} 
            & SSIM & Structural Similarity Index, measuring perceptual similarity and structural fidelity. & \cite{khaleghi2022visual, shimizu2022improving, bai2306dreamdiffusion, ahmadieh2024visual, sugimoto2024image} \\
            \cmidrule(lr){2-4}
            & PixCorr & Pixel-wise correlation coefficient between generated and target images. & \cite{shimizu2022improving} \\
            \cmidrule(lr){2-4}
            & CLIP Score & Measures semantic alignment using CLIP features against ground-truth caption or class. & \cite{rezvani2025interpretable, cheng2025fine} \\
            \cmidrule(lr){2-4} 
            & Feature Distance & Measures image similarity using features from modern backbones (e.g., AlexNet Top-$k$, SwAV). & \cite{rezvani2025interpretable, cheng2025fine} \\

        \midrule
        \multirow{1}{*}{\parbox{\linewidth}{Qualitative}} 
            & Visual Inspection / User Study & Human-perceived quality assessment, often reported via expert visual inspection or user studies. & -- \\
        \bottomrule
    \end{tabular}%
    }
    \caption{Evaluation metrics used in EEG-to-image generation studies.}
    \label{tab:eeg2image_metrics}
\end{table*}

\end{itemize}

\subsection{EEG Feature Encoding and Representation}
\label{subsec:eeg_image_feature_encoding}

Feature encoding is a critical step in EEG-to-image reconstruction, as it defines \textbf{how spatial patterns across electrodes and temporal dynamics of neural activity are extracted from raw signals} and transformed into latent representations for generation. Existing approaches can be broadly categorized into temporal modeling, spatial modeling, graph-based approaches, hybrid temporal-spatial models, and self/contrastive learning strategies.

\textbf{Temporal modeling} aims to analyze and learn the time-varying patterns in electroencephalogram (EEG) signals to understand brain activity over time, commonly implemented using Long Short-Term Memory (LSTM) networks, to capture temporal dependencies in EEG signals. \citet{kavasidis2017brain2image} employ an LSTM to generate compact, class-discriminative feature vectors that can also be used for object recognition. Building on this idea, \citet{singh2023eeg2image} integrate an LSTM encoder with a triplet-loss-based contrastive learning framework to improve feature discrimination, while \citet{singh2024learning} extend the design by combining CNN and LSTM encoders under triplet-loss supervision to further enhance discriminative learning. Similarly, \citet{ahmadieh2024visual} apply LSTMs across EEG channels and temporal windows and augment the extracted features through regression methods, including polynomial regression, neural network regression, and fuzzy regression.

\textbf{Spatial modeling} uses spatial arrangement and relationships between electrode locations on the scalp to create a richer understanding of brain activity. This has been addressed primarily through \textbf{convolutional neural networks (CNNs)}, which naturally capture dependencies across EEG channels. For instance, \citet{li2020semi} apply a feedforward neural network to project EEG signals into semantic features and \citet{wakita2021photorealistic} adopt a one-dimensional convolutional encoder-decoder within a multimodal VAE to get mean and variance vectors for EEG signal representation. More recent work has \textbf{integrated attention with CNN architectures}. \citet{mishra2023neurogan} used a convolutional encoder-decoder framework enhanced with an attention module to emphasize the most informative EEG channels. Other studies leverage specialized EEG network architectures like EEGNet \cite{lawhern2018eegnet} and Sinc-EEGNet \citep{bria2021sinc}, integrating attention mechanisms to highlight relevant frequency bands and channels \citep{sugimoto2024image, li2024visual}.

\textbf{Graph-based methods} have also been explored, where EEG signals are represented as connectivity graphs. \citet{khaleghi2022visual} construct graph embeddings derived from EEG functional connectivity and process them with a Geometric Deep Network (GDN) to obtain EEG feature vectors. Likewise, \citet{song2023decoding} integrate temporal-spatial convolution with self-attention and graph attention modules to improve feature extraction from EEG signals.

\textbf{Hybrid temporal-spatial models} aim to jointly exploit both domains. \citet{zeng2023dm} design a framework inspired by EEGChannelNet \citep{palazzo2020decoding} and ResNet-18, combining spatial, temporal, and multi-kernel residual blocks. Similarly, \citet{shimizu2022improving} introduce a \textbf{time-series-inspired architecture} with channel-wise transformations and temporal-spatial convolution to capture richer EEG representations.

Finally, \textbf{self-supervised and contrastive learning strategies} have been explored to \textbf{improve EEG feature extraction} while reducing dependence on labeled datasets. These approaches broadly fall into \textbf{two categories}: masking/reconstruction and cross-modal alignment (contrastive learning).

    \begin{itemize}
        \item \citet{bai2306dreamdiffusion} apply \textbf{masked signal modeling} with a masked autoencoder that reconstructs partially masked EEG tokens, thereby refining latent representations.

        \item The \textbf{cross-modal alignment} methods primarily rely on contrastive learning to align EEG with semantic or visual information. \citet{lan2023seeing} use contrastive learning to align EEG and image embeddings, enabling \textbf{extraction of pixel-level semantics and generation of saliency maps}; their framework further integrates \textbf{CLIP-based image caption embeddings} with an EEG sample-level encoder to strengthen cross-modal alignment. More recently, \citet{mehmood2025catvis} and \citet{rezvani2025interpretable} use a symmetric InfoNCE loss (similar to CLIP) to align EEG embeddings directly with corresponding text caption embeddings.
    \end{itemize}

An interesting alternative approach for semantic feature extraction is \textbf{knowledge distillation}, used by \citet{cheng2025fine} to train a high-level EEG encoder. In this method, a frozen ResNet50 classifier provides soft-target supervision, distilling its learned representations into the EEG encoder.

\subsection{Evaluation Metrics}
\label{subsec:eeg_image_eval}

Evaluation of EEG-to-image generation involves both \textbf{classification-based measures} (to verify that EEG features capture class discriminative information) and \textbf{image quality metrics} (to assess the fidelity and realism of generated outputs). Example of generated images from \cite{kavasidis2017brain2image} are shown in Figure~\ref{fig:eegimgresults}. Table \ref{tab:eeg2image_metrics} lists the key evaluation metrics with brief descriptions and exemplary studies that have used them.
\section{EEG-to-Text Generation}
\label{sec:eeg_text}

\subsection{Neural Basis and Signal Considerations}
\label{subsec:neural_basis_text}

In this section, we highlight studies that examine EEG activity in response to text stimuli and the underlying neural basis of language processing, which is essential for motivating how linguistic information can be reconstructed from the recorded signals. 

Neural activity during reading unfolds through a sequence of cognitive operations, beginning with the \textbf{visual perception} of written text, followed by \textbf{lexical retrieval, syntactic processing}, and ultimately \textbf{semantic unification} into a coherent mental representation. Each of these stages engages distributed neural networks rather than a single brain region, and their dynamics are reflected in the \textbf{spatiotemporal oscillatory patterns} captured by EEG. Importantly, EEG frequency bands do not map one-to-one onto specific language operations; instead, language comprehension emerges from a \textbf{dynamic interplay across $\alpha$, $\beta$, $\theta$, and $\gamma$ rhythms}, modulated by task demands and cortical region.

Studies show that neurocognitive aspects of language processing are associated with brain oscillations at various frequencies. A foundational review by \cite{bastiaansen2006oscillatory} synthesizes oscillatory signatures associated with language processing. Their findings show that $\theta$-band (4-7 Hz) synchronization increases during \textbf{memory retrieval} operations, whereas $\beta$ (12-30 Hz) and $\gamma$ ($>$30 Hz) synchronization are associated with \textbf{unification processes}, where lexical, syntactic, and semantic information are integrated.

Further evidence comes from \cite{bastiaansen2008see}, study that investigated neural dynamics elicited by open-class (semantic-bearing) and closed-class (function) words. Both word types produced $\theta$ power increases and $\alpha$/$\beta$ power decreases over left occipital (visual processing) and midfrontal regions. Notably, open-class words elicited an \textbf{additional $\theta$ power increase in the left temporal cortex}, consistent with its role in lexical-semantic retrieval. Moreover, words with auditory semantic properties produced greater $\theta$ power over electrodes near the \textbf{left auditory cortex}, whereas visually descriptive words produced stronger $\theta$ responses over the left visual cortex. These findings reinforce the view that $\theta$ \textbf{synchronization supports lexical-semantic retrieval}. Studies have also linked semantic memory operations to $\alpha$-band power changes.

Working memory (WM) is crucial for maintaining linguistic input during comprehension. \cite{bastiaansen2002event} associated $\theta$ synchronization with the formation of a \textbf{working-memory trace}, while \cite{weiss2005increased} reported stronger $\theta$ coherence in syntactically demanding \textbf{object-relative clauses} compared to subject-relative clauses.

\textbf{Semantic unification} processes also demonstrate frequency-specific markers. For example, \cite{hagoort2004integration} observed increased $\gamma$-band activity when sentences violated world knowledge as compared to violations of semantic features. Similarly, \cite{weiss2003contribution} found \textbf{greater $\gamma$ coherence} for semantically congruent versus incongruent sentence endings.

To examine EEG coherence patterns associated with \textbf{syntactic unification}, \cite{haarmann2002neural} investigated the neural processes involved in gap filling during online sentence comprehension and found that $\beta$-band coherence was significantly larger during verb processing in sentences requiring gap filling compared to those without syntactic gaps. Extending this line of evidence, \cite{weiss2012too} emphasized that \textbf{$\beta$-band activity plays a key role in cognitive and linguistic manipulations during language processing}, noting its involvement in higher-order linguistic functions such as word-category discrimination, semantic memory operations, and syntactic binding that supports meaning construction during sentence comprehension.

\begin{figure*}[t]
     \centering
    {\includegraphics[width=0.85\textwidth]{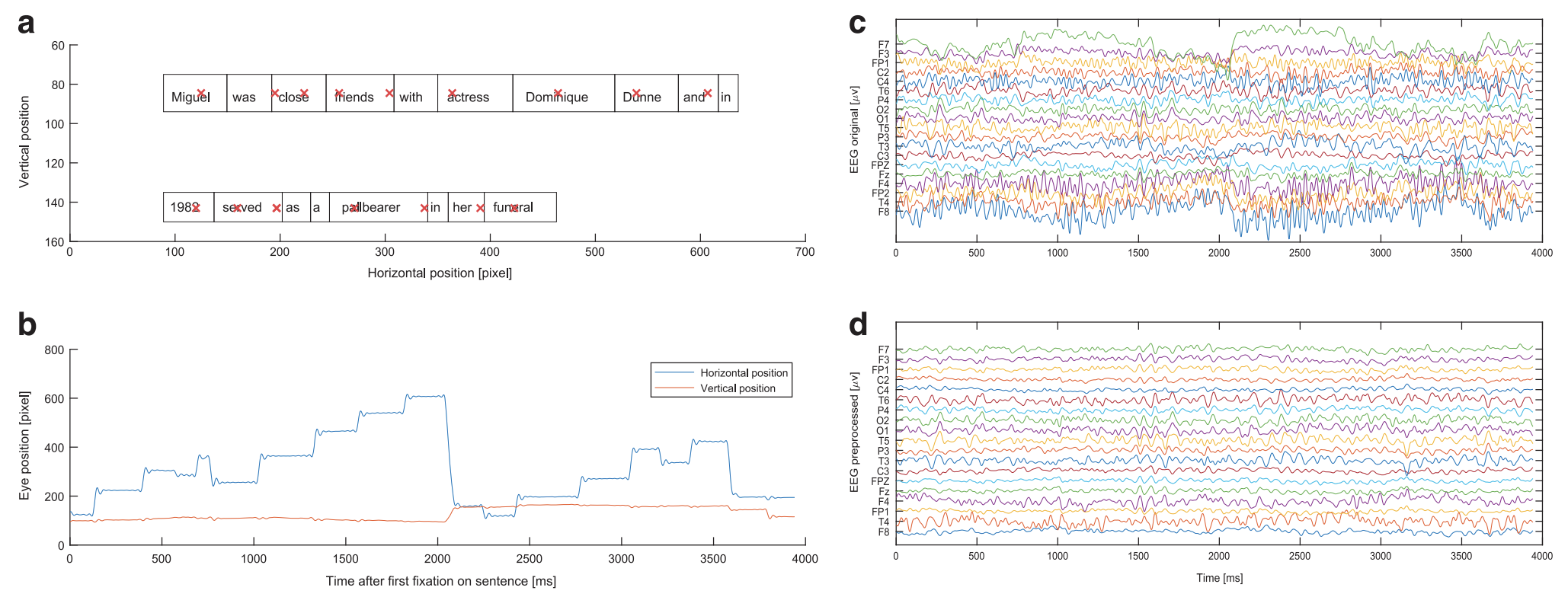}}
   \caption{Visualization of single trial EEG and eye-tracking data \citep{hollenstein2018zuco} (a) Fixations at the word level EEG marked with red cross and boxes. (b) Raw fixation gaze data. (c) Raw EEG signals. (d) EEG signals after preprocessing. }
	\label{fig:eye_fixation}
\end{figure*} 

Another important aspect of language comprehension is the \textbf{concurrent activation of multiple regions within a distributed cortical network}. \cite{marinkovic2004spatiotemporal} showed that processing written words begins in sensory-specific occipital areas and then progresses anteriorly along the ventral processing streams toward supramodal regions, including the temporal and inferior prefrontal cortices. The integration of a word into its contextual meaning peaks around \textbf{400 ms after word onset}, driven largely by coordinated activity between left temporal and inferior prefrontal regions during reading. EEG studies further support this spatiotemporal progression: \cite{fahimi2018semantic} reported that early word processing starts with visual processing in the occipital cortex, followed by activation along the ventral stream-first engaging left posterior occipital regions around \textbf{200 ms}, then reaching the posterior temporal cortex during the \textbf{semantic-processing window (around 400 ms)}. At later stages, anterior temporal, inferior frontal, and orbital regions become involved, reflecting the transition from perceptual processing to higher-order lexico-semantic integration.

Together, these findings demonstrate that EEG captures a rich hierarchy of lexical, syntactic, and semantic processes distributed across cortical networks, providing the neural foundation from which EEG-to-text generation models attempt to decode and reconstruct linguistic information.

\subsection{Data Acquisition}
\label{subsec:eeg2textdata}

We highlight major EEG datasets captured for text stimuli in Table~\ref{tab:datasets}. For such datasets, the most relevant factors are the number and demographics of subjects, the nature of tasks performed, the EEG recording setup, the method of aligning neural signals with text (often via eye-tracking), and the overall dataset size. 

\textbf{ZuCo Dataset experimental details:} To provide additional context, we describe the data acquisition process for \textbf{one key dataset}, the Zurich Cognitive Language Processing Corpus (ZuCo) \citep{hollenstein2018zuco}, which has become a benchmark for EEG-text research. The dataset was collected from \textbf{12 healthy native English speakers} engaged in naturalistic reading tasks, where \textbf{sentences were drawn from the Stanford Sentiment Treebank} (movie reviews)  \textbf{and the Wikipedia Relation Extraction corpus} (famous people labeled with relation types). Participants completed \textbf{three tasks}: sentiment analysis of movie reviews, relation recognition in Wikipedia sentences, and a task-specific relation classification, with a total reading duration of 4-6 hours per subject across two sessions. 

\textbf{Eye movements} were simultaneously recorded with an EyeLink 1000 Plus tracker (sampling rate of 500 Hz). Eye movement data is essential for providing word-level alignment between text and EEG signals, as it defines the precise onset and duration of fixations during reading. Figure~\ref{fig:eye_fixation} shows how eye position coordinates are used to capture word boundaries. High-density EEG was captured using a \textbf{128-channel} HydroCel system (500 Hz, 0.1-100 Hz bandpass). The final corpus includes over 21,000 words across 1,100 sentences and approximately 154,000 gaze fixations, providing a uniquely synchronized dataset of EEG and eye-tracking signals. 

\textbf{ZuCo Dataset frequency bands:} EEG signals captured across entire task period for \textbf{eight different frequency bands} resulting in a time-series for each frequency band - $\theta_1$ (4-6 Hz), $\theta$2 (6.5-8 Hz), $\alpha_1$ (8.5-10 Hz), $\alpha_2$ (10.5-13 Hz), $\beta_1$ (13.5-18 Hz), $\beta_2$ (18.5-30Hz), $\gamma_1$ (30.5-40 Hz), $\gamma_2$ (40-49.5 Hz). 

\textbf{Word level EEG Input for ZuCo:} Out of 128 channels, 9 EOG channels were used for artifact removal. Additionally, 14 channels primarily located on the neck and face were excluded from analysis.  \textbf{Each word-level EEG feature has a fixed dimension of 105.} To obtain word level EEG input, EEG data was aggregated on gaze duration. Features are concatenated from all 8 frequency bands, into one \textbf{feature vector with a dimension of 840.}

By enabling precise mapping between neural activity at word-level, ZuCo supports applications in NLP tasks such as sentiment analysis and relation extraction, as well as cognitive neuroscience research on natural reading. Beyond reading and comprehension analysis, the ZuCo dataset has also been used extensively in EEG-text generation studies.

\subsection{Overview of Generative Models and Techniques for Text Generation}
\label{subsec:overview_text_gen}

In the domain of text generation, technological advancements are largely driven by probabilistic language models. Early foundational work dates to the early 2000s, when \cite{bengio2003neural} proposed \textit{"learning a statistical model of the distribution of word sequences by neural networks"}, such that word representations capture the probability distribution of natural language sequences. Given a context c (a set of preceding words), the \textbf{model predicts the next word w in the sequence according to the conditional probability} $P(w \mid c)$, where $|c|$ denotes the size of the context window. This core \textbf{autoregressive principle} extends naturally to \textbf{Large Language Models (LLMs)}, which are trained on vast amount of data and can be trained or fine-tuned to perform tasks in diverse domains, from text-to-text generation (prompt-based) to \textbf{cross-modal generation} such as image-to-text, audio-to-text, or even EEG-to-text translation. 

The text generation models inherently perform temporal modeling i.e. predicting the next element in a sequence, which aligns closely with the temporal dynamics of EEG signals. Moreover, their ability to support many-to-many sequence generation (as in machine translation tasks) makes them particularly suitable for EEG-to-text generation applications. For EEG-to-text generation, we provide an overview of \textbf{language models} that are either used to \textbf{obtain robust EEG representations} for text generation (such as RNNs and GRUs) or that \textbf{utilize EEG representations to generate natural language text} (such as large language models (LLMs) like BART).

\begin{figure}[htbp]
    \centering
    \includegraphics[width=\linewidth]{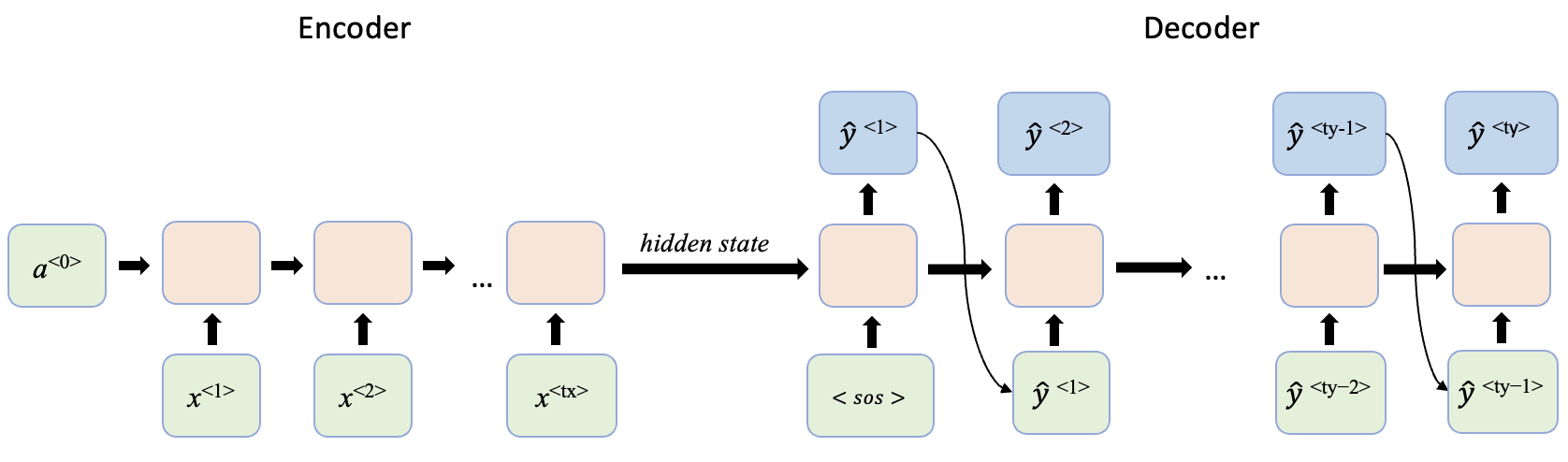}
    \caption{A traditional RNN architecture for many-to-many sequence generation in a machine translation task. It consists of an \textbf{encoder} that processes the input sequence $x$ and a \textbf{decoder} that takes the encoder’s hidden state $a$ as input and autoregressively generates the output sequence $\hat{y}$. The diagram shows an unrolled RNN illustrating sequence processing at each time steps $t$.}
    \label{fig:text-rnn}
\end{figure}

\textbf{Recurrent Neural Networks (RNNs)}, in contrast to simple feedforward neural networks, allow previous outputs to be reused as inputs through hidden states, enabling them to \textbf{model sequential dependencies in time-series data}. Given a sequence $x_1$, $x_2$, $x_4$, .., $x_{t-1}$, $x_t$, the activation corresponding to $x_1$ is used in processing $x_2$, and so on, as illustrated in Figure~\ref{fig:text-rnn}. This recurrent structure allows RNNs to \textbf{"remember" previous information and maintain context over time}, which is particularly useful for capturing temporal dependencies in data such as EEG signals and natural language. However, as information propagates over many time steps, RNNs struggle to capture long-term dependencies due to the vanishing and exploding gradient problems. To overcome these limitations, advanced variants such as \textbf{Long Short-Term Memory (LSTM)} and \textbf{Gated Recurrent Unit (GRU)} networks have been proposed. \textbf{LSTMs} introduce memory "cells" in the hidden state, regulated by three gates: the forget gate (determining what information to discard), the input gate (selecting what new information to store), and the output gate (controlling what information is passed to the next time step). \textbf{GRUs}, on the other hand, simplify this mechanism by using only two gates: reset and update, and by merging the cell and hidden states, thus \textbf{reducing computational complexity while maintaining comparable performance}. Additionally, Bidirectional RNNs (BiRNNs) extend these architectures by processing the sequence in both forward and backward directions, allowing the model to incorporate context from both past and future time steps. In the context of EEG-to-text generation, RNN-based architectures, including LSTMs, GRUs, and BiRNNs, are frequently employed for EEG feature extraction due to their ability to effectively capture the temporal dynamics inherent in EEG signals.

\begin{figure}[htbp]
    \centering
    \includegraphics[width=\linewidth, keepaspectratio, height=0.30\textheight]{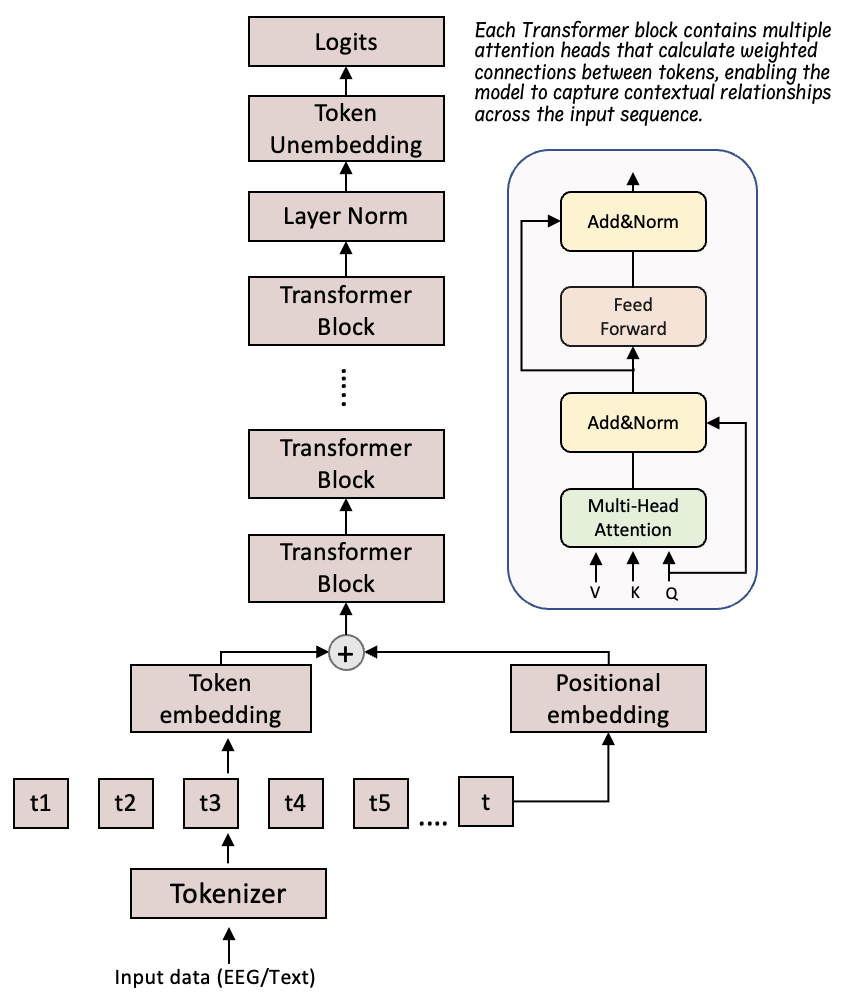}
    \caption{Multi-layer Transformer architecture and Transformer block illustration. The model takes a sequence as input and outputs logits, which are then used to generate text.}
    \label{fig:transformer}
\end{figure}

Over the past decade, \textbf{Transformer architectures} have dominated text generation, particularly in sequence-to-sequence (seq2seq) applications. While RNNs process data sequentially at each time step, they are limited in their ability to capture long-range dependencies within the data. \textbf{Transformers, with their self-attention mechanism, can model multiple dependencies simultaneously.} In a traditional Transformer based architecture, as shown in Figure~\ref{fig:transformer}, the input data is first divided into tokens, which are then passed to a series of transformer layers where the \textbf{attention mechanism} assigns weights that represent the relationships between these tokens, allowing the model to "attend" to the most relevant parts of the input sequence. This mechanism, combined with the Transformer’s ability to process data in parallel, greatly enhances its capacity to capture complex relationships and dependencies in large-scale datasets.

\begin{figure}[htbp]
    \centering
    \includegraphics[width=\linewidth, keepaspectratio, height=0.30\textheight]{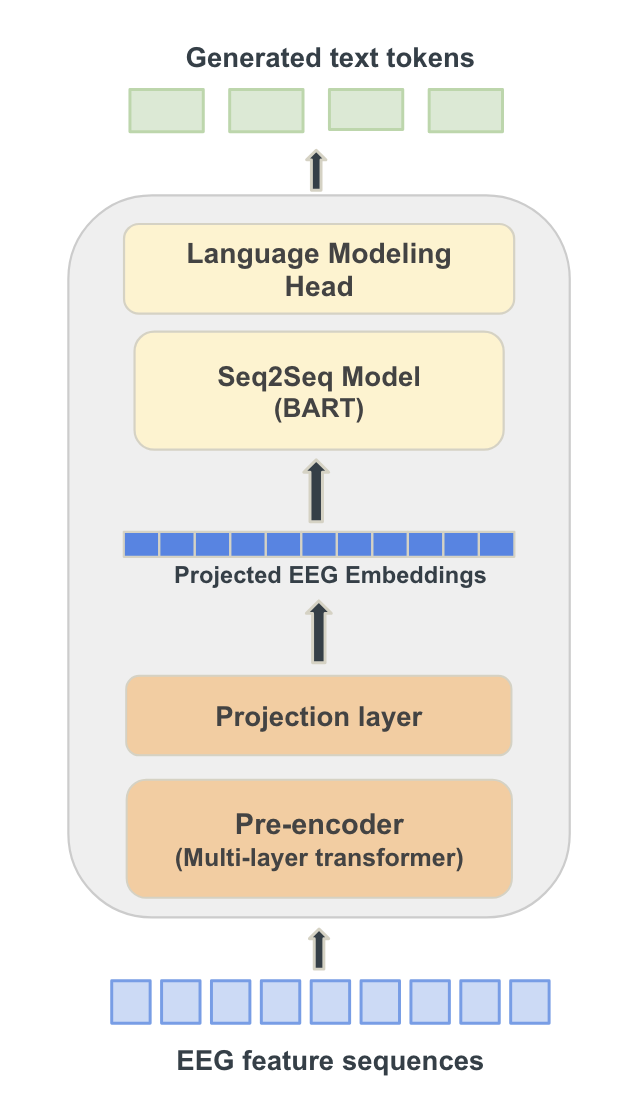}
    \caption{Illustration of the EEG-to-Text framework adapted from \cite{wang2022open}. The framework employs a multi-layer Transformer to extract EEG embeddings from input features, followed by an encoder-decoder based BART language model that generates text from the projected EEG embeddings.}
    \label{fig:eeg-text-transformer}
\end{figure}

\textbf{A typical EEG-to-text generation framework}, illustrated in Figure~\ref{fig:eeg-text-transformer}, consists of a \textbf{multi-layer Transformer}, used as a \textbf{pre-encoder} to extract embedding representations from EEG signals. These embeddings are then passed to a \textbf{pre-trained BART} (Bidirectional and Auto-Regressive Transformer) language model \textbf{to generate text from EEG-derived features}. \textbf{BART} employs a bidirectional encoder that captures contextual information in both directions to learn latent EEG representations, and an autoregressive decoder that generates coherent natural language text based on these learned representations.

In addition to this, for cross-modal learning, techniques such as \textbf{contrastive learning and masked signal modeling} have been employed. \textbf{Contrastive learning}, as described in Section~\ref{subsec:overview_image_gen}, helps the model learn more robust representations by distinguishing between EEG signals corresponding to the same text versus different text, or between signals from different subjects in EEG-to-text generation. In \textbf{masked signal modeling}, portions of the input sequence are masked, and the encoder is trained to reconstruct the original sequence, thereby learning \textbf{robust contextual representations}. Separate Transformer-based encoders, such as BART or other multi-layer Transformers, have been used independently for text and EEG masked signal modeling.

This overview provides a strong foundation for the upcoming sections, where we discuss how they have been integrated into various architectures for EEG-to-text generation tasks.

\subsection{Use Cases and Addressed Concerns}
\label{subsec:eeg_text_use_cases}

In the domain of generating text from EEG signals, early work has typically relied on \textbf{closed-vocabulary approaches}, where decoding is limited to a fixed set of predefined words \citep{biswal2019eegtotext, srivastava2020think2type, yang2023thoughts, rathod2024folded}. Among these, \citet{srivastava2020think2type} and \citet{yang2023thoughts} explore the use of Morse code representations, in which users’ active intent is mapped to Morse sequences that are subsequently translated into text.

More recent studies have moved toward \textbf{open-vocabulary generation}, enabling decoding beyond fixed word sets and aiming to support naturalistic conversation \citep{wang2022open, feng2023aligning, duan2023dewave, liu2024eeg2text, wang2024enhancing, amrani2024deep, tao2024see, mishra2024thought2text, ikegawa2024text, chen2025decoding, elgedawy2025bridging, masry2025ets, liu2025learning, lu2025eeg2text, jiang2025neural}. These works address several important concerns, including \textbf{subject-dependent variability} in EEG representations \citep{feng2023aligning, amrani2024deep, elgedawy2025bridging}, learning \textbf{cross-modal alignment} between EEG and text embeddings \citep{wang2024enhancing, tao2024see}, and \textbf{modeling long-term dependencies and contextual information} that may be missed by conventional architectures \citep{rathod2024folded, chen2025decoding, chen2025decodingtext}. 

A persistent challenge in EEG-to-text research is the \textbf{reliance on eye-tracking fixation data} to define word-level markers, which has been addressed by studies such as \citet{duan2023dewave} and \citet{liu2024eeg2text}. To mitigate the difficulty of defining word boundaries in EEG signals and to \textbf{broaden applicability across languages}, some studies have proposed \textbf{language-agnostic solutions} \citep{mishra2024thought2text, ikegawa2024text}, where signals are captured through visual stimuli and decoded via advances in image-text intermodality. \textbf{Data scarcity} also remains a limiting factor, and \citet{chen2025decoding} propose a VAE-based augmentation technique to expand EEG-text datasets and improve training efficiency.

Another major concern has been the semantic fidelity of generated text, since shallow lexical matching often fails to capture the intended meaning of the text. Approaches that explicitly encourage interpretable and semantically aligned representations, for example through contrastive learning \citep{feng2023aligning, tao2024see, liu2025learning}, have been introduced to \textbf{reduce semantic drift and improve the quality of open-vocabulary generation}. A pioneering study by \cite{zhou2024belt} trains a conformer- and transformer-based \textbf{multi-task model} with query prompts to enhance the encoding and decoding performance from EEG signals to \textbf{achieve coherent and readable sentence decoding results} from noninvasive EEG signals.

Relatedly, \textbf{cross-lingual generalization} has begun to receive attention, with studies like \cite{lu2025eeg2text} demonstrating that EEG-to-text decoding can extend beyond English by aligning Chinese EEG signals with text embeddings. Beyond reading-based tasks, EEG signals associated with handwriting have been used to decode individual alphabet letters to further generate fluent text, aiming toward a Spell-Based BCI System \citep{jiang2025neural}.

\begin{figure*}
     \centering
    {\includegraphics[width=\textwidth]{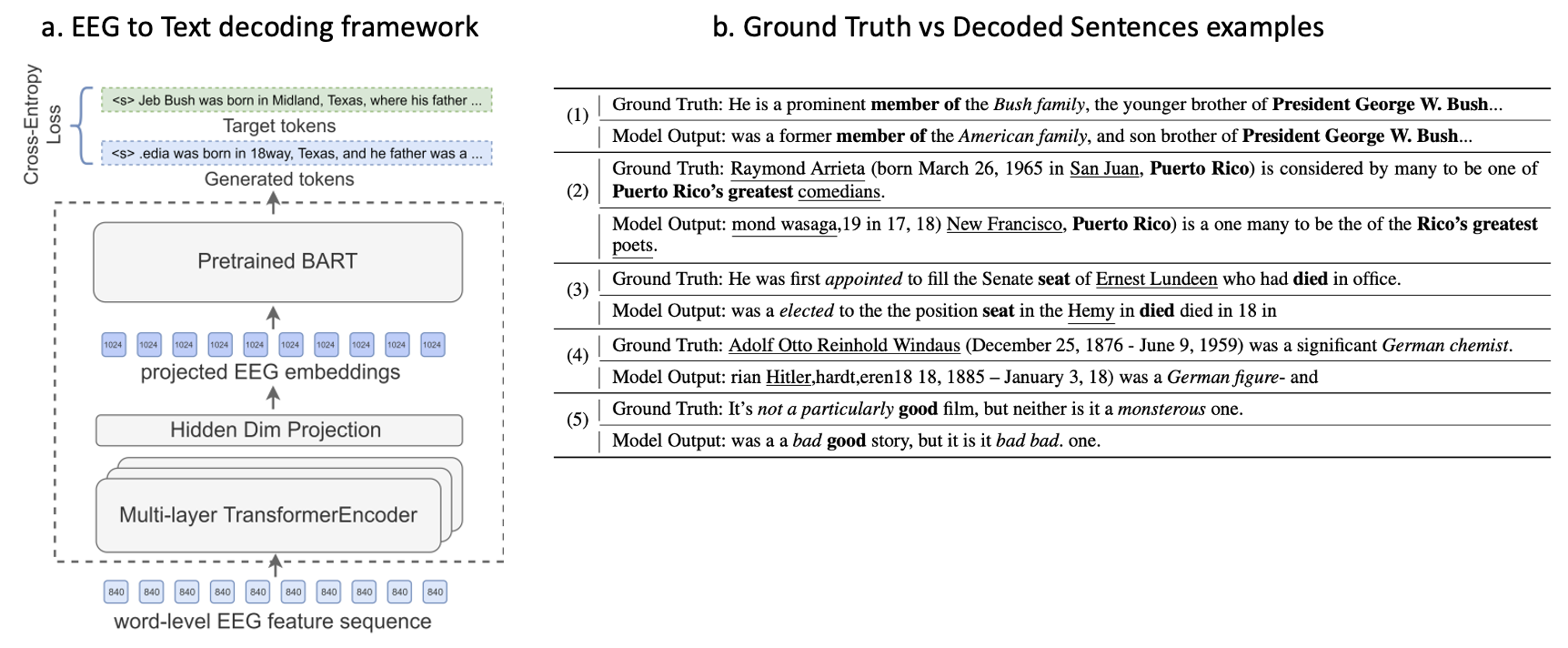}}
   \caption{(a) EEG decoding framework that takes word level EEG as input and follows a sequence-to-sequence neural machine translation task to predict output text. (b) Comparison of Ground Truth and Decoded sentences for exact word match and semantic resemblance \citep{wang2022open}.}
	\label{fig:eeg-text-decoding}
\end{figure*}

\subsection{Generative Architectures Used Across Studies}
\label{subsec:eeg_text_architectures}

EEG-to-text generation draws on a diverse set of modeling and representation learning approaches. We have found that \textbf{sequence-to-sequence (seq2seq) generation} serves as a major overarching framework. The idea is to formulate EEG to text generation as a \textbf{neural machine translation task} \citep{wang2022open} and maximizing the probability of decoded sentence which is the conditional probability of predicting the sequence $s_t$, given the EEG signal $\mathcal{E}$ and previously decoded sequences, where $T$ is the length of the target text sequence:

\begin{equation}
p(\mathcal{S}|\mathcal{E}) = \prod_{t=1}^{T} p(s_t \in \mathcal{V}|\mathcal{E}, s_{<t})
\end{equation}

After extracting word-level EEG inputs, as described in Section~\ref{subsec:eeg2textdata}, an EEG-to-text decoding framework similar to Figure~\ref{fig:eeg-text-decoding} is used to generate output text. Instead of requiring an exact sentence-level match, the generated text is evaluated against the original reference using semantic similarity and word-overlap metrics.

We now outline the major state-of-the-art techniques applied to EEG-to-text generation, explaining the underlying approach of each method and its relevance. 

\begin{itemize}
    \item \textbf{Large Language Models (LLMs}) are widely used due to their ability to capture \textbf{long-range dependencies in text} and generate coherent linguistic output. Several studies employ BART as the core text generation model \citep{wang2022open, liu2024eeg2text, wang2024enhancing, amrani2024deep, tao2024see, chen2025decodingtext}. In addition, \citet{mishra2024thought2text} fine-tuned LLMs on EEG embeddings alongside image-text data during training, enabling text generation directly from EEG signals during inference. \cite{liu2025learning} introduces a contrastive–generative framework that aligns EEG representations with a frozen Flan-T5 (LLM) latent space, \citep{jiang2025neural} explores letter-level decoding followed by generation with BART language model. 
    
    \item \textbf{Contrastive learning} is frequently applied to align EEG and text embeddings in a shared space, further \textbf{enhancing cross-modal alignment and improve feature discrimination} \citep{feng2023aligning, tao2024see, wang2024enhancing, liu2025learning}.

    \item \textbf{Masked signal modeling} is used as \textbf{self-supervised pretraining strategy} to learn robust EEG representations. For example, \citet{liu2024eeg2text} employ a transformer trained to reconstruct randomly masked segments of EEG data, thereby learning contextual and semantic dependencies at the sentence level. An integrated framework by \citet{tao2024see} combines contrastive learning with masked signal modeling, where word-level EEG feature sequences are randomly masked and sentence-level sequences deliberately masked, guided by an \textbf{intra-modality self-reconstruction objective.}
    
    \item \textbf{Recurrent Neural Networks (RNNs)} have been adopted to capture the sequential dynamics of EEG signals. \citet{amrani2024deep} use bidirectional GRUs to process word-level EEG signals of varying lengths, while \citet{chen2025decodingtext} extend this to a \textbf{hierarchical GRU} design that \textbf{captures both local contextual information and long-range dependencies} through the organization of hidden layers hierarchically. Similarly, \cite{elgedawy2025bridging} employs bidirectional GRUs to model temporal dynamics followed by a subject-specific adaptation layer to extract subject-dependent features.
    
    \item \textbf{Ensemble-based methods} have also been explored to enhance robustness and address class imbalance in the dataset. \citet{rathod2024folded} propose a folded ensemble deep CNN for text suggestion and a folded ensemble bidirectional LSTM for text generation, to improve accuracy of generated text.
\end{itemize}

\begin{table*}[t]
    \centering
    \footnotesize
    \renewcommand{\arraystretch}{1.5}
    \resizebox{\textwidth}{!}{%
    \begin{tabular}{>{\raggedright\arraybackslash}p{0.10\linewidth}>{\raggedright\arraybackslash}p{0.10\linewidth}p{0.4\textwidth}>{\raggedright\arraybackslash}p{0.4\linewidth}}
        \toprule
        \multicolumn{4}{c}{\textbf{EEG-to-Text Generation Evaluation Metrics}} \\
        \toprule
        \textbf{Category} & \textbf{Metric} & \textbf{Description / Usage in Studies} & \textbf{References} \\
        \midrule
        \multirow{3}{*}{\parbox{\linewidth}{Lexical overlap}} 
            & BLEU & $n$-gram precision overlap between generated and reference text. & \cite{papineni2002bleu, biswal2019eegtotext, wang2022open, feng2023aligning, mishra2024thought2text, zhou2024belt, elgedawy2025bridging, masry2025ets, liu2025learning, lu2025eeg2text, jiang2025neural}\\
        \cmidrule(lr){2-4}
            & ROUGE & Recall-oriented lexical overlap, emphasizes $n$-gram recall. & \cite{lin2004rouge, wang2022open, feng2023aligning, duan2023dewave, liu2024eeg2text, wang2024enhancing, mishra2024thought2text, zhou2024belt, elgedawy2025bridging, masry2025ets, liu2025learning, lu2025eeg2text}\\
        \cmidrule(lr){2-4}
            & METEOR & Includes stemming and synonymy for nuanced alignment. & \cite{banerjee2005meteor, biswal2019eegtotext, chen2025decodingtext, mishra2024thought2text, masry2025ets, elgedawy2025bridging}\\
        \midrule
        \multirow{2}{*}{\parbox{\linewidth}{Semantic similarity}} 
            & BERTScore & Embedding-based contextual similarity beyond lexical overlap. & \cite{zhang2019bertscore, amrani2024deep, mishra2024thought2text, elgedawy2025bridging, liu2025learning}\\
        \cmidrule(lr){2-4}
            & BLEURT & Learned evaluation metric based on pretrained language models. & \cite{sellam2020bleurt, chen2025decodingtext} \\
        \midrule
        \multirow{2}{*}{\parbox{\linewidth}{Error-based}} 
            & WER & Word-level transcription error rate. & \cite{feng2023aligning, jiang2025neural}\\
        \cmidrule(lr){2-4}
            & TER & Translation Error Rate; number of edits required to match reference. & \cite{chen2025decodingtext} \\
        \bottomrule
    \end{tabular}%
    }
    \caption{Evaluation metrics used in EEG-to-text generation studies.}
    \label{tab:eeg2text_metrics}
\end{table*}

\subsection{EEG Feature Encoding and Representation}
\label{subsec:eeg_text_feature_encoding}

Feature encoding is a critical step in EEG-to-text generation, as it transforms raw neural activity into structured representations aligned with linguistic output. Several distinct approaches have been proposed for encoding EEG signals, varying in how they model temporal dynamics, spatial dependencies, and higher-level semantic representations.

\textbf{Spatial-Temporal EEG Features} are critical for effective EEG-to-text decoding. \citet{biswal2019eegtotext} extract \textbf{shift-invariant features} using stacked CNNs and model \textbf{temporal dependencies} using RCNNs, followed by processing by hierarchical LSTMs for medical report generation. Similarly, \citet{srivastava2020think2type} combine spatial variation modeling using CNNs with extracting temporal dynamics with LSTMs , while \citet{chen2025decoding} enhance feature extraction under both classification and seq2seq objectives by incorporating \textbf{residual blocks to effectively capturing spatial-temporal EEG patterns}. This strategy continues in recent work, where to better represent \textbf{spatial feature maps and temporal dependencies across subjects}, bi-GRUs paired with subject-adaptive encoders and transformer layers are employed \cite{elgedawy2025bridging}. More recently, \citet{alharbi2024decoding} propose a hybrid deep learning model that combines 3D CNNs with various RNN architectures (LSTM, stacked LSTM, BiLSTM) to jointly capture spatial and temporal dynamics in EEG for speech decoding.

\textbf{Spectral and Statistical Features} are incorporated in some studies to enrich EEG representations beyond spatio-temporal patterns. Spectral features capture the distribution of signal energy across frequency bands, such as power spectral density, band power, or spectral entropy, providing information about oscillatory brain activity. \citet{yang2023thoughts} apply the \textbf{Short-Term Fourier Transform (STFT) to derive these spectral features} and concatenate them with \textbf{statistical measures} (e.g., minimum and maximum per channel), combining them with CNNs and RNNs to model spatial and temporal dynamics, respectively. Similarly, \citet{rathod2024folded} employ \textbf{Wavelet Transform (WT), Common Spatial Patterns (CSP), and statistical descriptors} to construct discriminative feature vectors for closed-vocabulary text classification. For open-vocabulary decoding, \cite{masry2025ets} use a combination of EEG spectral features and \textbf{synchronized eye-tracking features}: fixation duration (FFD), total reading time (TRT), and gaze duration (GD) to drive joint generation and sentiment classification.

\textbf{Contextual EEG Representations} have recently been advanced through transformer-based architectures that excel at modeling long-range dependencies. \citet{wang2022open} employ a multi-layer transformer encoder to map word-level EEG sequences into text representations, while \citet{feng2023aligning} use a transformer-based pre-encoder to \textbf{project word-level EEG features into the Seq2Seq embedding space}. Building on this, \citet{tao2024see} introduce a \textbf{cross-modal codebook that stores EEG embeddings alongside word embeddings} obtained from BART. To further enhance \textbf{contextual EEG representations} neural signals are aligned with pre-trained language model spaces. In this paradigm, \cite{liu2025learning} aligns neural features with a frozen Flan-T5 latent space, enabling semantically grounded and interpretable EEG-to-text decoding, while \cite{lu2025eeg2text} align transformer-based EEG decoding with MiniLM embeddings to improve cross-modal contextual decoding. To extract task-specific contextual EEG representations, \citet{zhou2024belt} proposed the \textbf{Q-Conformer}, comprising a discrete conformer, a Context Transformer (C-Former), and a learnable query prompt. The discrete conformer tokenizes EEG embeddings into discrete EEG tokens, while the C-Former extracts \textbf{task-specific contextual representations} through self-attention and cross-attention guided by the query prompt. The extracted representations are aligned with language using Byte Pair Encoding (BPE)-level contrastive learning and connected to a frozen LLM via prefix tuning.

\textbf{Marker-Free and Sentence-Level EEG Representations} address the limitations of relying on eye-tracking markers, which restrict the generalizability of word-level EEG features. \citet{duan2023dewave} extract both marker-aligned word-level features using multi-head transformers and raw EEG embeddings without markers using a multi-layer transformer encoder. For raw EEG waves, the encoder is trained to self-reconstruct waveforms while transforming signals into sequences of embeddings. \citet{liu2024eeg2text} pretrain a convolutional transformer on sentence-level EEG signals with a masking objective and employ a multi-view transformer to separately encode brain regions using dedicated convolutional transformers. Similarly, \citet{wang2024enhancing} integrate word-level and sentence-level EEG features, applying random masking at the word level and compulsory masking at the sentence level to enhance contextual representations.

\textbf{Hierarchical Temporal EEG Representations} continue to be effectively modeled using recurrent-hierarchical architectures. \citet{chen2025decodingtext} propose a hierarchical GRU decoder with a Masked Residual Attention Mechanism to capture both \textbf{local and global contextual information}. Similarly, \citet{amrani2024deep} employ bidirectional GRUs to process \textbf{variable-length word-level EEG sequences}, integrating them with a subject-specific 1D convolutional layer and a multi-layer transformer for richer feature encoding.

More recently, \textbf{Topographic maps} have been used by \citet{alharbi2024decoding} to capture the \textbf{brain’s dynamic responses}, where raw EEG for each 'imagined word' is represented as a single image composed of sixteen time-domain topographic brain maps, to capture both spatial distributions across electrodes and temporal dynamics of EEG into a unified representation.

\subsection{Evaluation Metrics}

Evaluation of EEG-to-text generation relies on established natural language processing metrics that compare generated text against reference text. As noted previously, the generated text is compared with the original reference using semantic similarity and word-overlap measures, rather than requiring an exact sentence level match. Figure~\ref{fig:eeg-text-decoding} (b) shows example of actual and decoded sentences. The metrics can broadly be grouped into \textbf{lexical overlap metrics, semantic similarity metrics, and error-based metrics.} Table \ref{tab:eeg2text_metrics} lists the key evaluation metrics with brief descriptions and exemplary studies that have used them. 

Several EEG-to-text studies adopt the same baseline and report a common set of evaluation metrics, enabling quantitative cross-study comparison. Accordingly, Table~\ref{tab:zuco_text_comparison} summarizes results obtained on the ZuCo dataset using the open-vocabulary EEG-to-text model proposed by \citet{wang2022open} as the baseline. ZuCo includes three reading tasks: sentiment reading (SR), normal reading of Wikipedia passages (NR), and task-specific reading of Wikipedia passages (TSR). ZuCo v1.0 contains recordings from 12 participants, whereas ZuCo v2.0 contains recordings from 18 participants. For each reading task, sentences were divided into training, development, and test sets using an 80\%/10\%/10\% split based on unique sentences, ensuring that all test sentences were unseen during training.

\begin{table*}[t]
    \centering

    \resizebox{\textwidth}{!}{%
        \begin{tabular}{>{\raggedright\arraybackslash}p{0.51\linewidth}>{\centering\arraybackslash}p{0.07\linewidth}>{\centering\arraybackslash}p{0.07\linewidth}>{\centering\arraybackslash}p{0.07\linewidth}>{\centering\arraybackslash}p{0.07\linewidth}>{\centering\arraybackslash}p{0.07\linewidth}>{\centering\arraybackslash}p{0.07\linewidth}>{\centering\arraybackslash}p{0.07\linewidth}}
            \toprule
            \multirow{2}{*}{\textbf{Method}}
            & \multicolumn{4}{c}{\textbf{BLEU-N (\%)}}
            & \multicolumn{3}{c}{\textbf{ROUGE-1 (\%)}} \\
            \cmidrule(lr){2-5}
            \cmidrule(lr){6-8}

            & \textbf{N=1}
            & \textbf{N=2}
            & \textbf{N=3}
            & \textbf{N=4}
            & \textbf{P}
            & \textbf{R}
            & \textbf{F1} \\
            \midrule

            EEG2Text / Wang and Ji (baseline) \citep{wang2022open}
            & 40.10
            & 23.10
            & 12.50
            & 6.80
            & 31.70
            & 28.80
            & 30.10 \\

            BrainBART-Large (w/ C-SCL) \citep{feng2023aligning}
            & 35.91
            & 25.96
            & 21.31
            & 18.89
            & --
            & --
            & -- \\

            DeWave \citep{duan2023dewave}
            & 41.35
            & 24.15
            & 13.92
            & 8.22
            & 33.71
            & 28.82
            & 30.69 \\

            EEG2TEXT (+ Multi-View Transformer) \citep{liu2024eeg2text}
            & 45.20
            & 29.10
            & 19.70
            & 14.10
            & 36.90
            & 32.00
            & 34.20 \\

            E2T-PTR \citep{wang2024enhancing}
            & 42.09
            & 25.13
            & 14.84
            & 8.99
            & 35.86
            & 30.01
            & 32.61 \\

            Deep Representation Learning for Open-Vocabulary
            EEG-to-Text Decoding \citep{amrani2024deep}
            & 42.75
            & 25.90
            & 15.66
            & 9.56
            & 36.71
            & 30.60
            & 33.28 \\

            ETS (BART) \citep{masry2025ets}
            & 43.39
            & 30.36
            & 23.67
            & \textbf{20.22}
            & 37.70
            & 35.77
            & 36.66 \\

            BELT-2 \citep{zhou2024belt}
            & 43.06
            & 25.57
            & 15.05
            & 9.09
            & 34.12
            & 30.28
            & 31.99 \\

            BELT-2 + LLM (T5) \citep{zhou2024belt}
            & \textbf{52.38}
            & \textbf{36.28}
            & \textbf{25.28}
            & 17.95
            & \textbf{39.47}
            & \textbf{36.08}
            & \textbf{37.59} \\

            \bottomrule
        \end{tabular}%
    }
    \caption{
        Cross-study comparison of EEG-to-text generation performance on the ZuCo dataset. All values are reported as percentages; dashes indicate metrics that were not reported, and the best results are highlighted in bold.
    }
    \label{tab:zuco_text_comparison}
\end{table*}

\subsection{EEG-to-Text Evaluation Methodology Concerns}
There have been growing efforts to develop more rigorous evaluation methodologies for open-vocabulary EEG-to-text models and to determine \textbf{whether conventional text-generation metrics accurately reflect the linguistic information encoded in EEG signals.} \citet{jo2024eeg} argue that existing open-vocabulary EEG-to-text studies often lack robust evaluation protocols and identify two key concerns. First, several studies use teacher-forced evaluation, in which the decoder receives the ground-truth previous token, together with the EEG representation, when predicting the next token. Although teacher forcing is commonly used during training, its use during evaluation makes it difficult to determine whether the generated text is driven primarily by the EEG signal or by the language model. Second, many studies do not compare their models against appropriate control baselines, such as random noise. Evaluating the same model using EEG and matched random or control inputs can help determine whether the decoded linguistic information is genuinely attributable to EEG rather than to the language model or the evaluation procedure. Consequently, commonly reported BLEU and ROUGE scores should be interpreted with caution, as they may overestimate the extent to which current models decode linguistic information from EEG.
\newline
\newline
In a subsequent study, the same authors \citep{jo2025evaluating} proposed the following best practices for more rigorous evaluation of EEG-to-text models:
\begin{itemize}

    \item \textbf{Noise-baseline testing:} Compare performance on EEG data with performance on random noise of matching dimensionality to distinguish genuine EEG-based learning from language-model priors or memorized textual patterns. Where applicable, report the signal-to-noise ratio (SNR) used to construct the baseline.
    
    \item \textbf{Statistical significance testing:} Report statistical comparisons between EEG and noise baselines using bootstrapped significance testing, along with confidence intervals and effect sizes.
    
    \item \textbf{Non-teacher-forced evaluation:} Prioritize evaluation under non-teacher-forced conditions, where models generate text autoregressively without access to ground-truth tokens during inference. 
    
    \item \textbf{Standardized data handling:} Use sentence-level train, validation, and test splits with no overlap between sets to prevent memorization of textual patterns.
    
    \item \textbf{Transparent preprocessing:} Clearly document preprocessing procedures, including filtering parameters, artifact removal methods, feature extraction settings (e.g., frequency bands and temporal windows), and the resulting feature dimensionality.
    
    \item \textbf{Performance comparison reporting:} Demonstrate that EEG-based models significantly outperform noise baselines before claiming genuine EEG signal learning, and report confidence intervals and effect sizes for all statistical comparisons.

\end{itemize}
\section{EEG-to-Audio/Speech Generation}
\label{sec:eeg_speech}

\subsection{Neural Basis and Signal Considerations}
\label{subsec:neural_basis_audio}

Different types of sound stimuli, such as \textbf{pure tones, speech, and music}, elicit distinct patterns of brain activity across cortical regions. When captured through EEG, this activity is reflected in the amplitudes or spectral power of different EEG frequency bands which vary across brain regions and reveal how auditory information is processed in the brain. Primarily, \textbf{sound is identified and processed in the auditory cortex located within the temporal lobe}, however, auditory processing is not confined to this region alone. Several studies have examined how the brain responds to different types of auditory stimuli.

As observed by \citet{krause1997relative}, neural signals evoked by sound stimuli that do not involve higher-level processing such as semantic interpretation or memorization are likely processed primarily within the auditory cortex, which is part of the temporal lobe. In contrast, \textbf{sound stimuli requiring speech perception engage additional brain regions beyond the auditory cortex}. To evaluate this distinction, the authors compared EEG responses while subjects listened to text played backward (lacking high-level processing) and text played forward (involving semantic comprehension).

For both types of stimuli, synchronization and desynchronization of EEG $ \alpha $ band activity (8-10 Hz and 10-12 Hz) were observed. Generally, increased cortical activity is associated with a decrease in $ \alpha $ amplitude, meaning that $ \alpha $ desynchronization signifies cortical activation. In the 10-12 Hz band, listening to forward (meaningful) text elicited desynchronization, whereas listening to backward (non-semantic) text elicited synchronization in the parieto-occipital area. The increased activity (synchronization) observed in the parieto-occipital area while listenting to backward text indicates that \textbf{auditory stimuli not involving higher level processing (semantics or memorization) are processed directly on the auditory cortex}. This dissociation was not observed in the 8-10 Hz band, where both stimulus types produced event-related desynchronization (ERD), which the authors attributed to non-specific cognitive processes such as sustained attention. 

However, some studies examining the characteristics of different brain waves across brain regions have reported frequency-specific activity in multiple cortical areas, even for sound stimuli that do not require higher-level processing. \citet{di2018experimental} investigated EEG responses to intermittent pure tone stimuli, analyzing $ \theta $ (4-7.5 Hz), $ \alpha $ (8-13 Hz), and $ \beta $ (14-30 Hz) waves across different brain regions, specifically, the four lobes (frontal, temporal, parietal, occipital) and both hemispheres (eight regions in total). The study found that \textbf{mean theta-wave amplitudes} differed significantly among the frontal, temporal, and parietal regions (but not the occipital region) across four time periods, one with sound stimulation and three without, highlighting the \textbf{significance of temporal factors on auditory EEG responses}. Significant interaction effects were also observed for \textbf{mean alpha wave amplitude} in the frontal region, and a significant main effect of hemisphere was found in the temporal region. Additionally, differences in the interaction among hemisphere, frequency, and loudness were detected in the mean beta-wave amplitude within the parietal region, suggesting that \textbf{auditory processing of even simple tone stimuli involves complex, region and frequency dependent neural dynamics}.

Other studies, such as \citet{farahani2021brain}, have examined both cortical and subcortical contributions to auditory processing in the brain. \textbf{Relatively high sound modulation frequencies (e.g., 40 and 80 Hz) are believed to engage subcortical neural generators more strongly than cortical ones} \citep{herdman2002intracerebral}. To investigate these mechanisms, researchers often analyze Auditory Steady-State Responses (ASSRs), which are brain responses evoked by modulated or repetitive acoustic stimuli and serve as reliable indicators of auditory temporal processing. Because of its excellent temporal resolution, EEG is particularly well-suited for capturing the auditory system’s synchronized responses to such stimuli, providing rich information about the dynamics of neural activity and the interaction between cortical and subcortical brain networks involved in auditory perception.

\textbf{These findings form the neurophysiological basis for EEG-to-audio generation}. The observed synchronization and desynchronization patterns, frequency-specific activations, and cortical-subcortical interactions reflect how the brain encodes different aspects of auditory information. \textbf{EEG signals thus capture both low-level acoustic features and higher-level speech perception.} EEG-to-audio techniques aim to leverage these patterns to reconstruct or generate audio by learning the mapping between neural activity and auditory representations.

\subsection{Data Acquisition}
\label{subsec:eeg2speechdata}

We summarize major EEG datasets collected for audio-related stimuli in Table~\ref{tab:datasets}. For speech-synthesis tasks, datasets may be recorded using either audio or text prompts, with participants engaged in reading, vocalized speech, or imagined speech activities. We highlight a representative dataset to illustrate how EEG data are collected, including details on the recording equipment and experimental stimuli and tasks.

\textbf{Dataset Experimental Details:} A notable resource for EEG-to-speech synthesis is the \textbf{KARA ONE corpus} \citep{zhao2015classifying}, which includes recordings of both \textbf{imagined and vocalized speech}. This multimodal dataset integrates \textbf{EEG, facial motion}, and \textbf{audio} data. Recordings were collected from \textbf{12 participants} using a \textbf{64-channel} Neuroscan Quick-Cap for EEG acquisition and a \textbf{Microsoft Kinect (v1.8)} sensor to capture synchronized facial video and speech audio.

Each recording session lasted approximately \textbf{30--40 minutes} and consisted of repeated sequences of four successive states: (i) a 5-second rest state, (ii) a stimulus state in which a text prompt appeared on screen and its auditory utterance was played through speakers, followed by a 2-second preparation phase (iii) a 5-second \textbf{imagined speech} state, and (iv) an overt \textbf{speaking state}, in which participants vocalized the prompt while facial and audio data were captured by the Kinect.  

In total, each prompt was presented \textbf{12 times} across \textbf{132 trials}. The stimuli included \textbf{seven phonemic/syllabic prompts} (/iy/, /uw/, /piy/, /tiy/, /diy/, /m/, /n/) and \textbf{four phonetically similar words} (\textit{pat}, \textit{pot}, \textit{knew}, \textit{gnaw}), selected to balance nasals, plosives, and vowels, as well as voiced and unvoiced phonemes. The data was filtered between 1 and 50 Hz, and mean values were subtracted from each channel. This multimodal dataset provides a valuable benchmark for studying both imagined and spoken speech decoding from EEG. 

\subsection{Overview of Generative Models and Techniques for Audio Generation}
\label{subsec:overview_audio_gen}

In this section, we provide an overview of the machine learning models and techniques used for EEG-to-audio generation. Some of the underlying architectures, such as RNNs, GRUs, Transformers, and Latent Diffusion Models, have already been discussed in Section~\ref{subsec:overview_image_gen} and Section~\ref{subsec:overview_text_gen}. Readers are referred to these sections for an overview of their architectures and functionality. 

\begin{figure}[htbp]
    \centering
    \includegraphics[width=\linewidth]{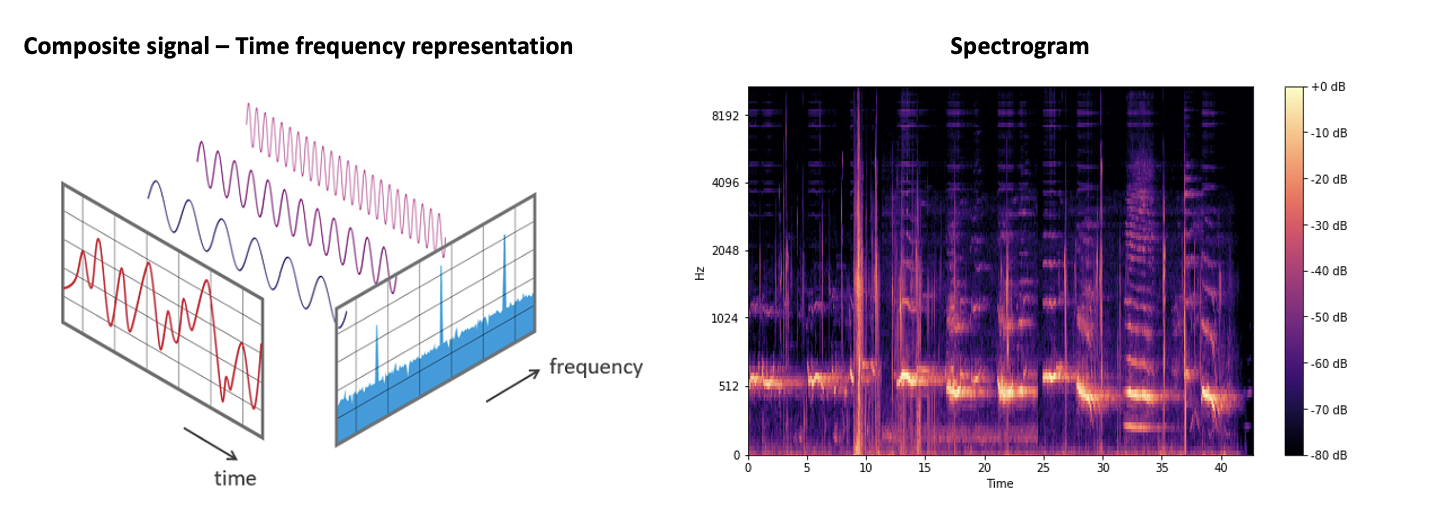}
    \caption{Time-frequency representation of a composite signal and a spectrogram representation showing time on the x-axis, frequency on the y-axis, and color intensity indicating signal amplitude.}
    \label{fig:wave-spectrogram}
\end{figure}

\begin{figure*}[t]
     \centering
    {\includegraphics[width=0.75\textwidth]{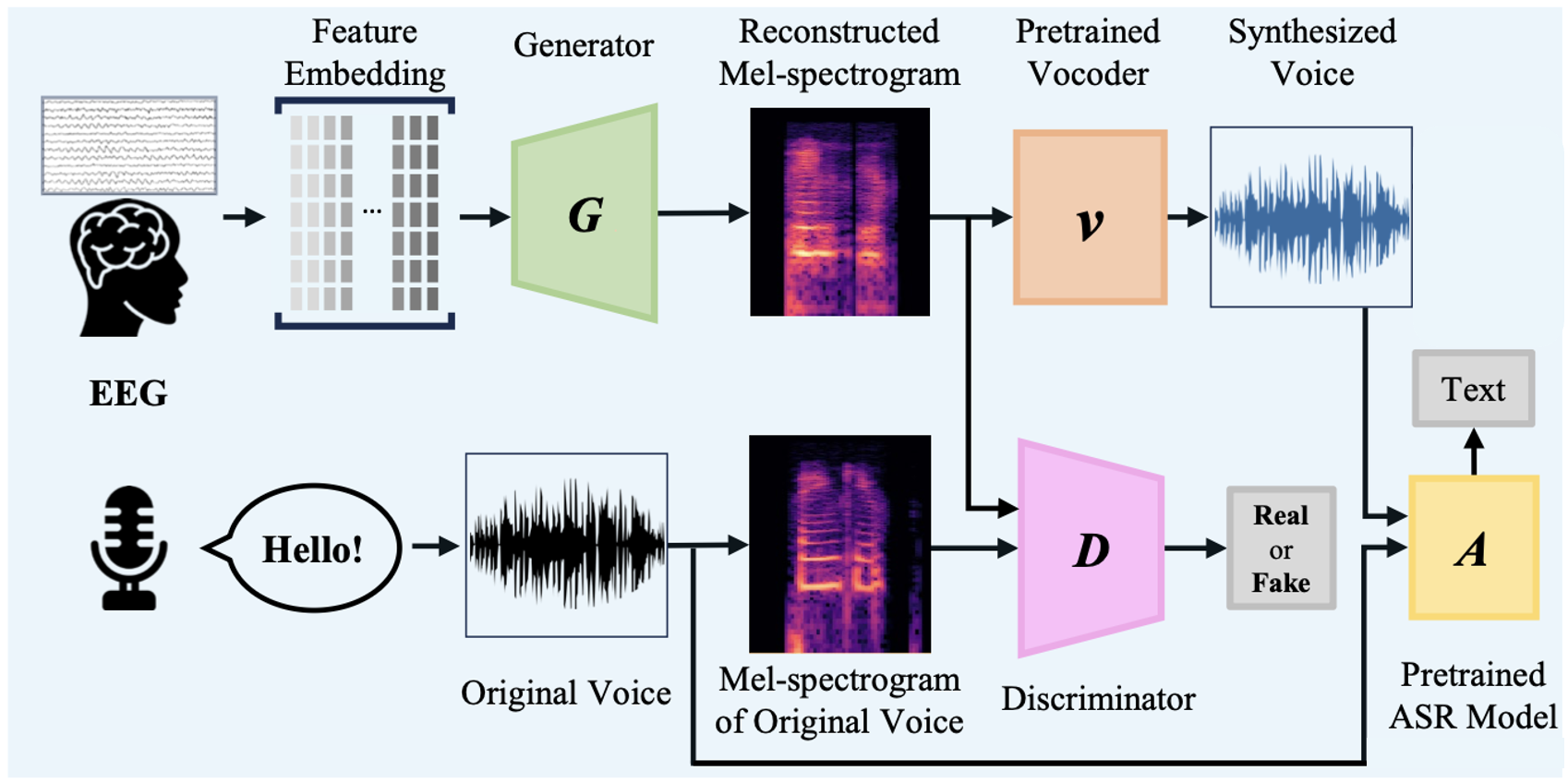}}
   \caption{Overall framework for EEG to speech synthesis \citep{park2024towards}: $G$ denotes the generator, which maps EEG embeddings to mel-spectrograms; $D$ is the discriminator that compares real and generated mel-spectrograms; $V$ is a pre-trained vocoder that converts mel-spectrograms into audio waveforms; and $A$ is a pre-trained ASR model for speech-to-text conversion.}
	\label{fig:eeg2speech}
\end{figure*}

\textbf{Signal-based data, contains both frequency and time-domain components}, is illustrated in Figure~\ref{fig:wave-spectrogram}. Instead of a single frequency, the actual signal is a composite waveform formed by the superposition of multiple frequencies. In the context of EEG, \textbf{band-pass filtering} is applied to separate specific frequency bands from the raw composite EEG data. When applying deep learning techniques, instead of using raw signal data directly, it is often transformed into an image representation. These image representations are typically mel-spectrograms. Of the surveyed studies in the domain of EEG-to-Audio/Speech generation, \textbf{most studies use mel-spectrogram as the intermediate input.}

A spectrogram, shown in Figure~\ref{fig:wave-spectrogram}, is a \textbf{visual representation of a signal with time on the x-axis and frequency on the y-axis}, providing a snapshot of how the frequency content of a composite waveform evolves over time. It uses color intensity to indicate the amplitude or power of each frequency component. Spectrograms are created using Fourier Transform, which decompose the signal into its constituent frequencies and displays the amplitude of each frequency present in the signal. In the case of audio signals, a mel-spectrogram is used, which applies the mel scale (a nonlinear, logarithmic frequency scale based on human auditory perception) on the y-axis and represents amplitude in decibels through color. \textbf{Mel-spectrograms more closely reflect human perception of audio and music.} For EEG-to-audio generation, to maintain representation consistency between modalities, several studies used mel-spectrograms for both EEG and audio data.

Deep learning models such as \textbf{Convolutional Neural Networks (CNNs)}, \textbf{Recurrent Neural Networks (RNNs)}, and hybrid architectures like \textbf{CNN-GRU} can be employed to generate spectrograms. In the EEG-to-audio generation domain, these models are trained to \textbf{construct spectrograms from EEG signals by minimizing the loss between EEG-generated and audio-generated spectrograms}, as shown in Figure~\ref{fig:eeg2speech}. \textbf{Neural vocoders} such as the High-Fidelity Generative Adversarial Network (HiFi-GAN) \citep{kong2020hifi} are then used to reconstruct audio waveforms using the learned spectrogram representations. \textbf{HiFi-GAN}, in addition to being a GAN-based architecture, employs two types of discriminators: \textbf{multi-scale discriminators}, which analyze the generated audio at different scales or resolutions, and \textbf{multi-period discriminators}, which learn diverse implicit structures by examining different parts of the input. This combination allows HiFi-GAN to \textbf{capture both the overall speech structure and fine-grained details}, thereby producing longer, high-fidelity audio outputs.

An overview of composite signals, mel-spectrogram representations, and the model architectures discussed in this and previous sections for image and text generation provides a strong foundation for the upcoming sections, where we examine how these models have been integrated into various architectures for EEG-to-audio/speech generation tasks.

\subsection{Use Cases and Addressed Concerns}
\label{subsec:eeg_audio_use_cases}

For EEG-based audio/speech generation, reported use cases include \textbf{speech synthesis} \citep{krishna2021advancing, lee2023speech},\textbf{music decoding and reconstruction} \citep{ramirez2022eeg2mel, postolache2024naturalistic}, \textbf{emotive music generation} \citep{jiang2024eeg}, \textbf{voice reconstruction} \citep{lee2023towards}, \textbf{talking-face generation} \cite{park2024towards}, and \textbf{speech recovery} \citep{mizuno2024investigation}. While many of these studies focus on decoding auditory information during listening tasks in speech or music perception \citep{krishna2021advancing, ramirez2022eeg2mel, park2024towards, mizuno2024investigation, postolache2024naturalistic, jiang2024eeg, lee2024toward, lee2025enhancing}, others explore speaking tasks and imagined speech \citep{krishna2021advancing, lee2023towards, lee2023speech, xiong2025synthesizing, park2025eeg}. 

For more naturalistic communication, \citet{lee2023towards} convert EEG signals recorded during imagined speech into the user’s own voice, enabling \textbf{personalized speech synthesis}. Similarly, \citet{park2024towards} generate speech from EEG while simultaneously producing a synchronized talking face with accurate lip movements. 

For \textbf{speech decoding from listened speech}, \citet{lee2024toward} propose an end-to-end framework that directly reconstructs natural speech waveforms from non-invasive EEG recorded while subjects listen to closed-vocabulary sentences. In a related work, the same authors \cite{lee2025enhancing} extend this framework with a BiLSTM based phoneme predictor that outputs decoded phoneme sequences in the text modality, enabling listened speech to be reconstructed in both modalities: speech waveforms and textual phoneme sequences.

These studies address \textbf{key challenges}, including the generation of fragmented or abstract outputs \citep{park2024towards}, the difficulty of synthesizing complete and continuous speech from EEG \citep{mizuno2024investigation}, the limitation of music reconstruction to relatively simple compositions with limited timbres \citep{postolache2024naturalistic}, the absence of temporally aligned acoustic targets in imagined speech \cite{park2025eeg}, the lack of standardized vocabularies for aligning EEG with audio data \citep{jiang2024eeg}, the reliance on intermediate acoustic feature mappings such as mel-spectrograms \citep{lee2024toward} and maintaining the integrity of signals in both the EEG and audio domains by preventing information loss introduced through transitional signals \citep{ma2025fully}.

\begin{table*}[t]
    \centering
    \footnotesize
    \renewcommand{\arraystretch}{1.5}
    \resizebox{\textwidth}{!}{%
    \begin{tabular}{>{\raggedright\arraybackslash}p{0.10\linewidth}>{\raggedright\arraybackslash}p{0.10\linewidth}p{0.50\textwidth}>{\raggedright\arraybackslash}p{0.30\linewidth}}
        \toprule
        \multicolumn{4}{c}{\textbf{EEG-to-Audio/Speech Generation Evaluation Metrics}} \\
        \toprule
        \textbf{Category} & \textbf{Metric} & \textbf{Description / Usage in Studies} & \textbf{References} \\
        \midrule
        \multirow{2}{*}{\parbox{\linewidth}{Signal-based}}
            & RMSE & Root Mean Square Error; measures the average reconstruction error between generated and reference signals. & \cite{krishna2021advancing, park2024towards, park2025eeg} \\
        \cmidrule(lr){2-4}
            & PCC & Pearson Correlation Coefficient; measures the linear correlation between reconstructed and reference signals. & \cite{park2025eeg, ma2025fully} \\
        \midrule
        \multirow{3}{*}{\parbox{\linewidth}{Spectral / acoustic}}
            & MCD & Mel Cepstral Distortion; quantifies spectral distortion between generated and reference speech. & \cite{krishna2021advancing, park2024towards, lee2024toward, lee2025enhancing, park2025eeg, ma2025fully} \\
        \cmidrule(lr){2-4}
            & SSIM & Structural Similarity Index; assesses perceptual similarity between generated and reference spectrograms. & \cite{ramirez2022eeg2mel, ma2025fully} \\
        \cmidrule(lr){2-4}
            & PSNR & Peak Signal-to-Noise Ratio; measures reconstruction fidelity of generated spectrograms. & \cite{ramirez2022eeg2mel} \\
        \midrule
        \multirow{3}{*}{\parbox{\linewidth}{Linguistic accuracy}}
            & WER & Word Error Rate; percentage of word-level transcription errors, indicating intelligibility of reconstructed speech. & \cite{mizuno2024investigation, park2025eeg} \\
        \cmidrule(lr){2-4}
            & CER & Character Error Rate; measures character-level transcription accuracy. & \cite{mizuno2024investigation, park2025eeg} \\
        \cmidrule(lr){2-4}
            & BERTScore & Embedding-based semantic similarity between reconstructed and reference transcripts. & \cite{mizuno2024investigation} \\
        \midrule
        \multirow{2}{*}{\parbox{\linewidth}{Perceptual quality}}
            & FAD & Fréchet Audio Distance; measures perceptual similarity by comparing the feature distributions of generated and reference audio. & \cite{postolache2024naturalistic} \\
        \cmidrule(lr){2-4}
            & MOS & Mean Opinion Score; listener-based subjective evaluation of speech naturalness and quality. & \cite{lee2023towards} \\
        \midrule
        \multirow{1}{*}{\parbox{\linewidth}{Task-specific}}
            & Hits@$k$ & Retrieval-based evaluation metric measuring whether the correct target appears within the top-$k$ retrieved results. & \cite{jiang2024eeg} \\
        \bottomrule
    \end{tabular}%
    }
    \caption{Evaluation metrics commonly used in EEG-to-audio and EEG-to-speech generation studies.}
    \label{tab:eeg2audio_metrics}
\end{table*}

\subsection{Generative Architectures Used Across Studies}
\label{subsec:eeg_audio_architectures}

EEG-to-speech and EEG-to-audio generation builds on a range of modeling and representation learning approaches. As we highlighted before, a prominent strategy is to leverage the shared signal modality between EEG and audio by mapping both into a \textbf{common intermediate representation, such as a mel-spectrogram}.

We now outline the major state-of-the-art techniques for EEG-to-speech generation, highlighting the underlying approaches of each method and their relevance.

\begin{itemize}
    \item \textbf{Convolutional Neural Network (CNN)-based models} have been widely employed to generate audio waveforms from EEG input \citep{krishna2021advancing, ramirez2022eeg2mel}. \citet{krishna2021advancing} propose a CNN architecture comprising temporal convolution layers, 1D convolutional layers, and a time-distributed layer to synthesize audio waveforms directly from EEG recorded during both speaking and listening tasks. Similarly, \citet{ramirez2022eeg2mel} use sequential CNN regressors to reconstruct music stimuli by mapping EEG features to mel-spectrogram representations.

    \item \textbf{Recurrent Neural Networks (RNNs)}-based models, particularly GRUs and LSTMs, have been widely used for sequential EEG encoding in speech reconstruction tasks. In the NeuroTalk framework, \citet{lee2023towards} employ a \textbf{GRU-based generator to reconstruct mel-spectrograms} from imagined-speech EEG, which are subsequently converted into speech waveforms using the \textbf{HiFi-GAN vocoder} \citep{kong2020hifi} and transcribed into text using HuBERT \citep{hsu2021hubert} for evaluation. \citet{park2024towards} further extend NeuroTalk by integrating synchronized talking-face generation using Wave2Lip \citep{prajwal2020lip} and Apple's Avatar API to achieve realistic lip synchronization with the synthesized speech.

    \item \textbf{GAN-based Architectures} have also been employed for EEG-to-speech generation. \citet{ma2025fully} proposed a DualGAN architecture with multi-scale optimization and cycle-consistency loss for end-to-end EEG-to-speech translation. The model incorporates a multi-scale feature extraction module based on InceptionV4 \citep{szegedy2017inception} to improve feature learning during signal downsampling. Furthermore, the cycle-consistency loss facilitates cross-modal translation between the EEG and audio domains while maintaining the integrity of signals in both domains. In addition, pretrained HiFi-GAN vocoders  have been widely adopted in EEG-to-speech pipelines to convert generated mel-spectrograms into speech waveforms, serving as the final speech synthesis stage in models such as NeuroTalk and EEG-to-Voice \citep{lee2023towards, park2024towards, park2025eeg}.
    
    \item \textbf{Encoder-decoder based architectures.} \citet{park2025eeg} employ a subject-specific encoder-decoder architecture for direct EEG-to-voice reconstruction. By leveraging local temporal context, the model generates mel-spectrograms from EEG without requiring explicit temporal alignment or dynamic time warping (DTW). The generated mel-spectrograms are subsequently converted into speech waveforms using a pretrained HiFi-GAN vocoder, transcribed using HuBERT, and further refined using an instruction-tuned language model to correct ASR errors while preserving semantic structure.

    \item \textbf{Transformers and latent diffusion models} have recently been adopted \textbf{to capture long-range dependencies and improve audio reconstruction quality}. \citet{mizuno2024investigation} employ transformer-based architectures for EEG-to-speech reconstruction, while \citet{jiang2024eeg} use a transformer encoder \textbf{to generate emotive music from EEG signals}. In the domain of naturalistic music decoding, \citet{postolache2024naturalistic} integrate EEG features with AudioLDM2 \citep{liu2024audioldm}, a pre-trained latent diffusion model, guided by a ControlNet adapter \citep{zhang2023adding} to achieve controllable and high-quality music generation.

\end{itemize}

\subsection{EEG Feature Encoding and Representation}
\label{subsec:eeg_audio_feature_encoding}

Feature encoding determines how neural signals are transformed into intermediate acoustic representations or directly mapped into audio features. Existing studies employ a variety of strategies, including the extraction of articulatory and acoustic features, temporal feature encoding, and the use of mel-spectrograms as shared representations.

\textbf{Articulatory and acoustic features} provide an interpretable bridge between EEG signals and speech outputs. \citet{krishna2021advancing} incorporate an attention model to predict articulatory features from EEG activity and an attention-regression model to convert these into acoustic features for waveform synthesis.

\textbf{Temporal feature encoding} has also been explored as a pathway to represent EEG dynamics for speech and music generation. \citet{jiang2024eeg} derive EEG tokens through a multi-step process that includes DBSCAN clustering to extract temporal features, which are then augmented with positional encoding to yield structured EEG tokens that can be used in downstream decoding tasks.

\textbf{Mel-spectrograms} provide a shared latent space between EEG and audio signals and thereby facilitating cross-modal translation. \citet{ramirez2022eeg2mel} employ a sequential CNN regressor to map EEG signals directly to time-aligned music spectrograms and \cite{xiong2025synthesizing} use a CNN-GRU based decoder to convert learned EEG latent representation to mel-spectrograms. Similarly, \citet{park2025eeg} propose a subject-specific generator comprising an EEG embedding layer, an encoder, a local-context adapter, and a mel-spectrogram decoder that reconstructs mel-spectrograms from preprocessed EEG signals.

In a related approach, \citet{lee2023speech} adapt spoken EEG into the subspace of imagined EEG by applying \textbf{Common Spatial Pattern (CSP) filters} trained on imagined speech, capturing temporal oscillatory patterns and reducing distributional differences between spoken and imagined EEG to enable user-specific voice synthesis. Similarly, \citet{postolache2024naturalistic} apply CSP filtering while temporally aligning EEG with speech, using triggers to mark onset intervals and segment continuous brain signals into utterance-specific intervals.

\subsection{Evaluation Metrics}

Evaluation of EEG-to-speech generation relies on established speech and audio metrics that \textbf{assess similarity to reference signals, intelligibility, perceptual quality, and subjective human judgments}. These can be broadly grouped into signal-based metrics, spectrogram similarity metrics, linguistic accuracy metrics, perceptual quality metrics, and subjective assessments. Table \ref{tab:eeg2audio_metrics} lists the key evaluation metrics with brief descriptions and exemplary studies that have used them.

\section{Cross-Modal Analysis of EEG-Based Generation}
\label{sec:cross_modal_analysis}

\begin{table*}[!t]
    \centering
    \footnotesize
    \renewcommand{\arraystretch}{1.3}
    \resizebox{\textwidth}{!}{%
        \begin{tabular}{p{0.21\linewidth}p{0.19\linewidth}p{0.19\linewidth}p{0.17\linewidth}p{0.24\linewidth}}
            \toprule
            \textbf{Encoding Methodology} & \textbf{EEG-to-Image} & \textbf{EEG-to-Text} & \textbf{EEG-to-Audio} & \textbf{Cross-Modal Interpretation} \\
            \midrule

            CNN-based spatial/local encoding &
            Common &
            Common &
            Common &
            Broadly transferable \\
            \midrule

            LSTM/GRU temporal encoding &
            Common &
            Common &
            Common &
            Broadly transferable \\
            \midrule

            Hybrid spatial-temporal encoding &
            Common &
            Common &
            Common &
            Shared architectural foundation \\
            \midrule

            Transformer-based encoding &
            Emerging &
            Prominent &
            Emerging &
            Transferable, but unevenly adopted \\
            \midrule

            Attention mechanisms &
            Channel, frequency, and semantic attention &
            Contextual and cross-modal attention &
            Temporal and contextual attention &
            Role varies by modality \\
            \midrule

            Contrastive alignment &
            Common &
            Common &
            Limited &
            Strong image-text, audio-text transfer potential \\
            \midrule

            Masked signal modeling &
            Reported &
            Reported &
            Not commonly reported &
            Modality-independent potential \\
            \midrule

            Graph-based encoding &
            Prominent &
            Rare &
            Rare &
            Mostly image-oriented \\
            \midrule

            Word- or marker-aligned features &
            -- &
            Prominent &
            -- &
            Text-specific \\
            \midrule

            Mel-spectrogram representation &
            -- &
            -- &
            Prominent &
            Audio-specific \\

            \bottomrule
        \end{tabular}%
    }
    \caption{Cross-modal comparison of EEG feature-encoding strategies used in EEG-based media generation.}
    \label{tab:cross_modal_encoding}
\end{table*}

\subsection{Cross-Modal EEG Feature Encoding: General Principles and Modality-Specific Adaptation}

Across EEG-to-image, EEG-to-text, and EEG-to-audio generation, feature encoding serves a common objective: transforming noisy, high-dimensional EEG signals into compact representations that preserve information relevant to the target output. Although the specific architectures vary, the surveyed studies reveal \textbf{several recurring principles}, including temporal modeling, spatial modeling, joint spatial-temporal encoding, representation alignment, self-supervised learning, and subject-specific adaptation.

\textbf{Temporal and spatial modeling} represents the most widely shared EEG encoding principles across modalities. Since EEG signals contain both temporal dependencies and spatial relationships across electrodes, RNN-based, CNN-based, and hybrid architectures are commonly used across EEG-to-image, EEG-to-text, and EEG-to-audio studies. LSTMs and GRUs have been employed to capture temporal dynamics in image reconstruction \citep{kavasidis2017brain2image, singh2023eeg2image, singh2024learning}, text generation \citep{srivastava2020think2type, amrani2024deep, chen2025decodingtext, elgedawy2025bridging}, and mel-spectrogram generation for speech or audio reconstruction \citep{lee2023towards, xiong2025synthesizing}. Similarly, CNN-based encoders have been used for spatial and local feature extraction in image pipelines \citep{wakita2021photorealistic, mishra2023neurogan, zeng2023dm}, text pipelines \citep{biswal2019eegtotext, srivastava2020think2type, alharbi2024decoding}, and audio pipelines \citep{krishna2021advancing, ramirez2022eeg2mel}. These studies indicate that hybrid spatial-temporal encoders provide one of the most generalizable architectural foundations for cross-modal EEG generation.

\textbf{Transformer- and attention-based encoding} represents another transferable strategy, although its role varies across modalities. In EEG-to-text generation, transformers are frequently used as contextual EEG encoders because they can model long-range dependencies across word-level and sentence-level EEG sequences \citep{wang2022open, feng2023aligning, duan2023dewave, liu2024eeg2text}. In EEG-to-image pipelines, attention mechanisms are commonly used to emphasize informative channels, frequency bands, or semantic features \citep{mishra2023neurogan, song2023decoding, li2024visual}. Transformer- and attention-based components are also increasingly used to refine EEG representations and project them into pretrained visual-semantic or generative spaces. In EEG-to-audio generation, transformer encoders have been applied to capture long-range temporal dependencies and support speech, audio, and music reconstruction \citep{mizuno2024investigation, jiang2024eeg}. Transformer-based modeling is therefore applicable across modalities, although its use as a direct contextual EEG encoder is most established in EEG-to-text generation.

\textbf{Cross-modal representation alignment} is used to reduce the semantic gap between EEG signals and the target modality. Contrastive objectives align EEG embeddings with representations obtained from images, captions, words, or pretrained multimodal models. In EEG-to-image generation, contrastive learning has been used to align EEG with visual or text-image embeddings and improve class discrimination \citep{singh2023eeg2image, lan2023seeing, song2023decoding, mehmood2025catvis, rezvani2025interpretable}. In EEG-to-text generation, related objectives align neural representations with word embeddings or pretrained language-model spaces \citep{feng2023aligning, tao2024see, wang2024enhancing, liu2025learning}. Comparable contrastive EEG--audio alignment is less common in the surveyed audio literature, indicating that this strategy is transferable but remains unevenly explored across modalities.

\textbf{Masked signal modeling} provides a potentially modality-independent self-supervised strategy for EEG feature learning. Rather than relying entirely on labeled EEG-stimulus pairs, masked modeling trains the encoder to reconstruct missing signal segments or latent tokens. This approach has been applied in EEG-to-image reconstruction through masked autoencoding \citep{bai2306dreamdiffusion} and in EEG-to-text generation through word-level and sentence-level masking objectives \citep{liu2024eeg2text, tao2024see, wang2024enhancing}. Its also use  represents an opportunity for future cross-modal adaptation in EEG-to-audio generation.

\textbf{Subject-specific and subject-adaptive encoding} is also relevant across modalities because EEG patterns vary substantially between individuals. Subject-wise latent alignment has been introduced in image-generation systems \citep{kneeland2026enigma}, subject-specific convolutional or adaptation layers have been used in text decoding \citep{amrani2024deep, elgedawy2025bridging}, and subject-specific generators or encoders have been adopted for speech reconstruction \citep{park2025eeg}. These approaches suggest that subject adaptation should be treated as a general component of EEG representation learning rather than as a modality-specific design choice.

Despite these shared principles, modality-specific preferences emerge from the structure of the target representation. \textbf{EEG-to-image generation} places greater emphasis on spatial organization, visual-category discrimination, and alignment with visual-semantic spaces. Consequently, graph-based models, attention mechanisms for identifying informative channels, class-discriminative objectives, and projection into CLIP or diffusion-conditioning spaces are particularly relevant \citep{khaleghi2022visual, song2023decoding, lan2023seeing, mehmood2025catvis, cheng2025fine}. \textbf{EEG-to-text generation} prioritizes sequential organization, variable-length contextual information, and alignment with linguistic representations. Word-level segmentation, hierarchical recurrent models, transformer encoders, and projection into pretrained language-model embedding spaces are therefore commonly employed \citep{wang2022open, duan2023dewave, tao2024see, zhou2024belt, liu2025learning}. Marker-free and sentence-level encoders have also emerged as alternatives to eye-tracking-dependent word segmentation \citep{duan2023dewave, liu2024eeg2text, wang2024enhancing}. \textbf{EEG-to-audio generation} places greater emphasis on preserving fine-grained temporal continuity and acoustic structure. Temporal convolutions, recurrent encoders, CSP-based spatial filtering, articulatory features, and mel-spectrogram reconstruction are therefore especially relevant \citep{krishna2021advancing, ramirez2022eeg2mel, lee2023speech, postolache2024naturalistic, park2025eeg}.

Overall, the surveyed literature suggests a general cross-modal framework in which EEG signals are first transformed by a spatial–temporal encoder and then adapted to the requirements of the target modality. The resulting representations may be aligned with visual, linguistic, or acoustic latent spaces before being passed to a modality-specific generation module. Table~\ref{tab:cross_modal_encoding} summarizes the shared encoding strategies and modality-specific preferences identified across the three modalities.

\subsection{Cross-Modal Evaluation Methodologies}
\label{appendix:evaluation_methodologies}

Table~\ref{tab:cross_modal_eval} presents a cross-modal comparison of the evaluation methodologies used in EEG-based media generation studies. Although the overall evaluation objectives are similar across modalities, the specific metrics and experimental protocols differ substantially, revealing opportunities for both unified benchmarking and methodological transfer. The metrics are organized into common evaluation dimensions to identify gaps within individual modalities and to examine whether successful evaluation strategies can be adapted across EEG-to-image, EEG-to-text, and EEG-to-audio generation.

A major gap across all three modalities is the lack of standardized evaluation protocols. Issues such as subject leakage, inconsistent data-splitting strategies, and inflated performance estimates are not modality-specific and therefore require common evaluation guidelines. Establishing modality-independent protocols for subject-wise splitting, cross-subject validation, reporting standards, and baseline comparison would improve the reliability and comparability of results across the field.

As modern generative models increasingly rely on learned embedding spaces, there is an opportunity to develop evaluation measures that assess reconstruction and semantic fidelity in a modality-agnostic manner. Evaluation approaches developed within one modality, such as retrieval-based assessment, embedding-based similarity, or human-centered perceptual evaluation, may also be adapted to address corresponding gaps in other EEG-based generative tasks.

\begin{table*}[!t]
    \centering
    \footnotesize
    \renewcommand{\arraystretch}{1.3}
    \resizebox{\textwidth}{!}{%
        \begin{tabular}{p{0.17\linewidth}p{0.18\linewidth}p{0.18\linewidth}p{0.18\linewidth}p{0.12\linewidth}p{0.27\linewidth}}
            \toprule
            \textbf{Evaluation Dimension} & \textbf{Image} & \textbf{Text} & \textbf{Audio/Speech} & \textbf{Shared?} & \textbf{Cross-pollination Opportunity} \\
            \midrule

            Semantic fidelity &
            CLIP Score&
            BERTScore, BLEURT&
            BERTScore (transcripts) &
            Partial &
            Develop unified multimodal embedding-based semantic evaluation (e.g., CLIP/CLAP/BERTScore). \\
            \midrule
            Perceptual quality &
            FID, IS, LPIPS, Inception Score, Diversity Score, Feature Distance (AlexNet, SwAV based)&
            Human fluency and relevance judgments &
            FAD, MOS &
            No &
            Extend perceptual and human-centered evaluation protocols across all modalities. \\
            \midrule
            Signal / Reconstruction fidelity &
            SSIM, PixCorr&
            BLEU, ROUGE, METEOR, WER, TER&
            RMSE, PCC, MCD, SSIM (spectrograms), WER, CER&
            Partial &
            Investigate modality-independent reconstruction fidelity measures for continuous outputs. \\
            \midrule
            Task-specific performance &
            Top-$k$ Accuracy, Recall@$k$ &
            -&
            Hits@$k$&
            Partial &
            Standardize downstream and retrieval-based evaluation across modalities. \\
            \midrule
            Evaluation protocol &
            Limited standardized protocols &
            Teacher-forced vs.\ autoregressive decoding, noise baselines, statistical significance testing (recommended best practices) &
            Limited standardized protocols &
            No&
            Adopt standardized evaluation protocols, including subject-independent testing, statistical significance analysis, and appropriate baseline comparisons across all modalities. \\

            \bottomrule
        \end{tabular}%
     }
    \caption{Cross-modal comparison of evaluation methodologies used in EEG-based media generation.}
    \label{tab:cross_modal_eval}
\end{table*}

\section{Limitations and Future Work}
\label{sec:limitations_and_future_work}

This section outlines the major limitations observed across all modalities and explores potential directions for advancing the field by addressing these challenges and leveraging the gaps identified in existing studies.

\begin{enumerate}
    \item \textbf{ML features vs. EEG-domain features.} In existing studies, temporal and spatial EEG features are typically extracted using machine-learning architectures such as CNNs and autoregressive sequence models. However, these learned representations are rarely compared with \textbf{traditional EEG-domain features}, including time-domain waveform oscillations, Event-Related Potentials (ERPs), or phase-synchronization indices. However, direct quantitative comparisons between modern embedding-based representations and traditional EEG-domain features remain scarce, particularly in EEG-based generative decoding, as few studies evaluate these feature representations under identical experimental settings. Establishing a \textbf{comparison between ML-derived features and neuroscience-informed EEG features}, or integrating EEG-domain features to guide or augment the generative process, remains an open and promising direction for future research.
        
    \item \textbf{Embedding based features lack cognitive intuitiveness.} Most current EEG-to-media models rely on embedding-based representations learned by deep networks, but these features are difficult to interpret and offer limited insight into the underlying cognitive processes. This \textbf{lack of interpretability reduces our ability to connect model representations with established neuroscientific knowledge} and hinders trust in the decoding pipeline. There is a need to integrate principles from cognitive psychology and neuroscience to design features that more directly reflect brain activity. Additionally, emerging ML interpretability techniques could be leveraged to map embedding-based representations onto cognitive processes and improve the transparency of generative models.

    \item \textbf{Algorithm Optimization.} Although some studies explore multimodal fusion, this area remains largely underdeveloped. \textbf{Advancing multimodal fusion strategies and designing EEG-specific deep learning frameworks} could lead to more effective modeling of how the brain processes information across different modalities. Such optimized architectures may better capture the complementary relationships between EEG signals and visual, auditory, or linguistic stimuli, ultimately improving the quality and robustness of EEG-to-media generation.

    {\item \textbf{Laboratory-grade versus affordable and portable EEG acquisition.} Most EEG-based media-generation studies rely on carefully controlled acquisition conditions and research-grade EEG systems. Practical deployment, however, may require affordable, portable, and lower-density devices. This transition introduces two important challenges:
    \begin{itemize}
        \item \textit{Reduced spatial information:} Moving from high-density laboratory systems with (128\text{--}256) channels to portable systems with substantially fewer electrodes, typically (8\text{--}32) channels, reduces the spatial coverage of scalp activity. Sparse electrode sampling limits the ability to distinguish localized and distributed neural responses, potentially reducing the information available for decoding.
        \item \textit{Reduced and more variable signal quality:} Recordings collected outside controlled laboratory environments are more susceptible to muscle activity, eye movements, electrode-contact variability, and environmental interference. These artifacts can lower the signal-to-noise ratio (SNR) and make it more difficult to learn reliable mappings between EEG signals and visual, linguistic, or acoustic representations.
    \end{itemize}
    
    Recent studies have explored several approaches to mitigate these hardware limitations. One direction is the use of generative AI for EEG spatial super-resolution. For example, \citet{wang2025generative} proposed a spatio-temporal adaptive diffusion model that reconstructs high-resolution, 256-channel EEG from recordings containing 64 or fewer channels. The reconstructed signals improved downstream classification and source-localization performance, demonstrating the potential of generative super-resolution for EEG analysis. However, its effectiveness as a preprocessing stage for EEG-to-image, EEG-to-text, or EEG-to-audio generation has not yet been directly evaluated.

    A second direction is the development of streamlined generation pipelines for portable or low-density EEG. \citet{guenther2024image} demonstrated image classification and reconstruction from visually evoked activity recorded using a portable eight-channel EEG system. Similarly, \citet{lopez2025guess} introduced a streamlined EEG-to-image framework that uses a ControlNet adapter to condition a latent diffusion model, reducing the need for extensive preprocessing and multiple auxiliary components.

    Large-scale data collection may also improve decoding from affordable hardware. The Alljoined-1.6M dataset contains more than 1.6 million visual-stimulus trials from 20 participants, recorded using a 32-channel consumer-grade system costing approximately \$2,200 \citep{xu2025alljoined}. The study demonstrates semantic category decoding, image retrieval, and EEG-to-image reconstruction despite the lower signal fidelity of the acquisition system. It also reports approximately log-linear improvements in decoding performance as the amount of training data increases, indicating that data scaling remains beneficial for consumer-grade EEG. However, these findings do not establish that dataset scale fully compensates for reduced channel coverage or lower signal quality. Further research is needed to evaluate these approaches under uncontrolled recording conditions, across devices and subjects, and in real-time EEG-based media-generation systems.}

    \item \textbf{Cross-subject generalization and evaluation.} 
    Cross-subject generalization in EEG-based media generation remains largely understudied. Nevertheless, many of the challenges identified in EEG classification literature are equally applicable to generative models, making this body of work a valuable reference. \citet{li2026cross} comprehensively review cross-subject EEG decoding methods for classification tasks, categorizing existing approaches into feature alignment, adversarial learning, feature disentanglement, contrastive learning, transfer learning, and related techniques. These methodologies provide promising directions for learning subject-invariant representations in EEG-based generative AI. 
    
    Reliable assessment of such methods also requires appropriate evaluation protocols. \citet{li2026cross} emphasize the importance of \textbf{subject-independent evaluation}, in which the participants included in the training and test sets are strictly disjoint. By contrast, in a \textbf{subject-dependent evaluation} setup, recordings from the same participant may appear in both sets, particularly when data are randomly split without accounting for subject identity. This overlap can introduce data leakage and produce overly optimistic performance estimates \citep{brookshire2024data}. Supporting this concern, \citet{wang2026causesperformancedegradationcrosssubject} show that performance can decline substantially under subject-independent evaluation, with some benchmark datasets exhibiting absolute \textbf{F1-score reductions exceeding 30 percentage points} relative to subject-dependent evaluation. These findings highlight the need for participant-disjoint evaluation protocols to accurately assess whether EEG-based generative models can generalize to unseen subjects.

    \item \textbf{Inherent limitations of EEG.} EEG suffers from a \textbf{relatively low signal-to-noise ratio}, which poses challenges for reliably capturing fine-grained neural information. Existing studies typically mitigate this limitation through preprocessing techniques such as filtering, artifact removal, and spatial filtering. However, the lack of standardized preprocessing pipelines makes it difficult to fairly compare models and assess the robustness of EEG-based generative systems. Recent work has shown promising progress in EEG media generation, and the combination of EEG with complementary modalities, such as functional near-infrared spectroscopy (fNIRS), could further improve signal quality and robustness. Moreover, \textbf{EEG primarily captures cortical surface activity, with limited sensitivity to subcortical structures} that play crucial roles in auditory, emotional, and temporal processing. Consequently, EEG-to-media generation models rely predominantly on cortical correlates and may fail to capture important subcortical neural activity, particularly for tasks such as speech reconstruction and auditory decoding. Integrating EEG with complementary modalities that better capture subcortical dynamics (e.g., fNIRS, MEG, or auditory brainstem responses) could provide richer neural representations and improve generation quality.

    \item \textbf{Lack of standardized benchmarks and cross-study comparison challenge.} A major challenge is the absence of standardized paradigms for EEG-to-media research, including consistent protocols for data collection, data splitting, feature extraction, model development, and evaluation. Studies also differ substantially in the datasets, generative architectures, and evaluation metrics they use, making direct cross-study comparison difficult and obscuring how generation performance has evolved over time. Without standardized experimental settings, it is difficult to determine whether reported improvements are genuinely attributable to architectural advances or instead arise from differences in datasets, splits, preprocessing, or evaluation procedures. Similar concerns have been raised in recent benchmarking studies of EEG foundation models, where variations in pre-training objectives, fine-tuning strategies, downstream tasks, and evaluation protocols hinder \textbf{fair comparison} across models \citep{xiongeeg}.
    Establishing standardized baselines, such as shared datasets, reference frameworks, and evaluation metrics, would enable more reliable comparisons, improve reproducibility, and support systematic progress in future research.

\end{enumerate}
\section{Ethical Considerations and Emerging Regulation}

EEG and other neural signals are widely studied for clinical applications and brain–computer interfaces (BCIs), which enable users to interact with external devices through brain activity. As neural decoding becomes more advanced, concerns arise about what information can be inferred from these signals and the resulting implications for privacy, autonomy, and “mind reading.” Although most EEG studies remain limited to controlled settings, they raise ethical issues that are also relevant to other forms of neural data.

To address these ethical considerations, several frameworks have been proposed to guide the responsible conduct of human research. One such framework is the \textbf{seven ethical requirements} proposed by \citet{emanuel2000makes}, which provide a strong foundation that can also be applied to EEG-based studies. These requirements include social or scientific value, scientific validity, fair subject selection, a favorable risk-benefit ratio, independent review, informed consent, and respect for the rights, interests, and well-being of participants. These principles provide a useful foundation for EEG data-collection studies. The seven ethical requirements proposed by \citet{emanuel2000makes}, which provide a general foundation for EEG data-collection studies, are summarized in Appendix Table~\ref{tab:ethical_requirements}.

Beyond these general requirements, EEG and other neural-signal research raises additional ethical concerns. The following sections discuss the key issues identified in the literature.

\subsection{Subject Re-Identification Risks}
Researchers have investigated raw EEG signals and derived features for subject identification, authentication, and biometric recognition, with several studies reporting high classification performance. Early studies focused on specific tasks: \citet{poulos1999person} obtained classification scores of 72\% to 84\% for EEG signals collected in a resting state with eyes closed, while \citet{paranjape2001electroencephalogram} observed classification accuracy ranging from 49\% to 85\% for EEG collected from 40 subjects with eyes open. \citet{brigham2010subject} achieved a higher accuracy of 98.96\% across 102 subjects using EEG collected during an imagined speech task.

More recent work has focused on methods that generalize subject identification across multiple cognitive tasks. \citet{kong2019eeg} proposed an EEG-based biometric identification method that extracts phase synchronization (PS) features for subject identification across a variety of tasks. The method achieved 97\% classification accuracy on two datasets in the beta and gamma bands, collected from 20 subjects watching a video and 12 subjects driving, respectively. It also achieved 97\% accuracy on a motor imagery BCI dataset across all frequency bands, showing that phase synchronization of EEG signals has task-free biometric properties. Other studies use multi-task methods that integrate EEG features from multiple tasks \citep{ruiz2016cerebre, ruiz2017permanence} to improve the reliability of biometric systems.

However, subject identification from EEG signals is not without limitations, as discussed by \citet{chan2018challenges}. Classification and identification systems often require extensive or complicated setups, and performance can deteriorate over time, making identification harder even a few days after initial data collection. Physiological changes such as aging and psychological changes such as stress or shifting mental states also affect brainwaves, which can reduce identification accuracy in real-world or long-term settings. Together, these findings suggest that although EEG poses a genuine re-identification risk, particularly for datasets collected under controlled conditions, maintaining reliable subject identification over extended periods and in practical deployments still remains a significant challenge.

\subsection{Mental Privacy and Cognitive Liberty}

\textbf{A major ethical concern is the compromise of mental privacy.} EEG signals are among the most sensitive forms of human data, as they encode neural correlates of cognition, perception, emotion, and potentially intent. The trajectory of EEG-based generative AI raises longer-term concerns about non-consensual or coercive applications, including workplace monitoring, lie detection, and surveillance. If neural data become commoditized, they could be used by third parties to influence users' behavior and, in more extreme cases, enable neurosurveillance with broader societal and political implications.

When mental privacy is at risk, \textbf{cognitive liberty} is also threatened. As noted by \citet{rainey2020brain}, brain activity collected to infer mental contents can provide insight into a person's thoughts, potentially undermining an individual's unique and privileged access to their own thoughts and, consequently, their cognitive liberty. The authors also express concern that technologies capable of decoding neural activity may influence an individual's thought processes and self-conception. If mental data were to become accessible to third parties or collected and analyzed beyond an individual's control, individuals may no longer feel free to think certain thoughts, even for the purpose of analysis, for fear that those thoughts could be inferred, scrutinized, or exposed.

For EEG-based studies, these concerns emphasize the importance of ensuring that participants are fully informed about how their neural data will be collected, analyzed, stored, shared, and used. They also highlight the need for regulatory safeguards governing the responsible collection, management, and use of neural data to protect mental privacy and cognitive liberty.

\subsection{Informed Consent}

Informed consent plays a central role in protecting research participants. Its three pillars, disclosure, capacity, and voluntariness, provide a useful framework for guiding the ethical considerations that should be addressed before, during, and after EEG experiments. Relevant information about the study should be disclosed to a capable participant who voluntarily chooses to enroll. 

As discussed by \citet{hendriks2019ethical}, several factors should be considered when obtaining informed consent from research participants -

\textbf{Disclosure:} Participants should be informed about the risks and benefits of the study, the procedures involved, and any additional procedures that are not part of standard EEG recording.

\textbf{Capacity:} Researchers should assess a participant's capacity to provide informed consent by evaluating their understanding, appreciation, reasoning, and ability to communicate a choice regarding participation. In some cases, participants may have the capacity to consent but be unable to communicate their decision. In such situations, alternative communication methods should be used to enable participants to provide consent or assent.

\textbf{Voluntariness:} Researchers should ensure that participants understand that participation is voluntary and should be sensitive to any potential pressure that may influence their decision to participate.

These pillars provide a strong foundation for informed consent, but a single framework may not be sufficient across different cultural and social contexts. \citet{botes2025decoding} emphasize that the informed consent process should be culturally attuned in addition to being ethically sound to foster genuine understanding, trust, and voluntariness. The authors recommend using consent materials that are accessible across languages and adapted to local literacy levels. They also advocate extending the consent process beyond the individual to include family or kin when health-related decisions are made collectively within a cultural context. These considerations can help ensure that informed consent is both ethical and culturally appropriate across diverse populations.

Other significant challenges to informed consent in neurological research include cognitive and communication impairments, mistrust of medical research, time constraints, literacy barriers, lack of available social support, and practical or resource-related constraints \citep{sankary2023overcoming}. To address these challenges, the authors recommend involving family members or close others in the consent process when appropriate, training researchers to obtain informed consent effectively, encouraging participants to review consent materials before discussions, using written and visual aids to improve understanding, and actively addressing misconceptions, mistrust, and external pressures that may influence participants' decision-making.

\subsection{Emerging Regulations}
As neurotechnology continues to advance across academia and industry, establishing appropriate ethical and regulatory frameworks has become increasingly important. Such frameworks can help ensure the responsible development and deployment of neurotechnology while protecting individuals from the misuse of neural data. In response, governments and international organizations have introduced a range of legal, policy, and governance initiatives to regulate the collection and use of neural data. Some of the key developments are summarized below.

\begin{itemize}
    \item \textbf{Chile (2021):} Chile became the first country to introduce constitutional protections related to neurorights by amending its Constitution to protect the physical and mental integrity of its citizens, including special protection for brain activity and the information derived from it \citep{chile2021law21383}.
    
    \item \textbf{OECD Recommendation (2019):} The OECD Recommendation provides the first international standard for the responsible innovation of neurotechnology. It outlines principles for promoting responsible innovation, safety, transparency, privacy, human rights, and the anticipation of unintended uses and misuse of neurotechnology \citep{oecd2019neurotechnology}.
    
    \item \textbf{UNESCO Recommendation (2025):}  UNESCO adopted the Recommendation on the Ethics of Neurotechnology \citep{unesco2025neurotechnology}. The Recommendation highlights the potential misuse of neural data for surveillance, targeted marketing, and political influence, as well as the risk of widening social inequalities through unequal access to neurotechnology. It calls on Member States to establish governance and regulatory frameworks that protect human rights, mental privacy, autonomy, freedom of thought, and mental integrity while ensuring the responsible development and deployment of neurotechnology.
    
    \item \textbf{Colorado House Bill 24-1058 (2024):} Colorado became the first U.S. state to explicitly protect neural data by amending the Colorado Privacy Act. The law classifies biological data, including neural data, as sensitive data, requires opt-in consent for its processing, and extends consumer rights, including access, correction, and deletion, to such data \citep{colorado2024hb1058}.
    
    \item \textbf{California SB 1223 (2025):} California enacted SB 1223, amending the CCPA/CPRA to explicitly classify neural data as sensitive personal information. The law defines neural data as information generated by measuring activity of the central or peripheral nervous system and extends existing CCPA protections, including consumers' right to limit the use and disclosure of such data \citep{colorado2024hb1058}.
    
    \item \textbf{Management of Individuals' Neural Data (MIND) Act (United States, proposed, 2025):} The proposed MIND Act directs the Federal Trade Commission  to study the governance of neural and related data and develop recommendations for their collection, processing, use, transfer, and protection. It emphasizes informed consent, transparency, and safeguards against the misuse of neural data for behavioral inference, influence, and manipulation \citep{mindact2025}.
\end{itemize}

\section{Conclusion}
\label{sec:conclusion}

With rapid progress in machine learning, generative AI in particular, EEG is emerging as a viable foundation for cross-modal generation, including text, image, and speech synthesis directly from brain signals. Despite persistent challenges such as low signal-to-noise ratio and limited spatial resolution, this growing research area demonstrates a meaningful shift toward more ambitious, non-invasive brain–computer interface (BCI) capabilities.

In this survey, we reviewed the landscape of EEG-based generative AI research across text, image, and audio domains, highlighting representative use cases, datasets, generative frameworks, and open challenges. We provided a comprehensive synthesis of methods employed in EEG-driven generative tasks, summarizing major datasets and the core model architectures adopted across the surveyed studies. To ground these developments, we also included foundational background on generative modeling and an overview of the neural basis through which EEG reflects perceptual and cognitive activity elicited by different types of stimuli. In addition, we identified key limitations of existing work, including small and heterogeneous datasets, low signal fidelity, limited cross-subject generalization, and the need for more interpretable and neurally aligned generative frameworks.

Progress in this field will require the development of standardized benchmarks with consistent data splits, preprocessing pipelines, and evaluation metrics. Such standardization would improve reproducibility, enable fairer comparisons, and help clarify which methodological innovations genuinely advance performance. Ethical considerations, including privacy, informed consent, and responsible use of neural data, should remain central as research continues to evolve.

Overall, while EEG-driven generative AI remains an emerging and highly experimental area, continued research, supported by better datasets, unified evaluation practices, and deeper integration with neuroscientific principles, is essential for advancing understanding and expanding the scope of what non-invasive neural decoding can achieve.

\bibliographystyle{elsarticle-harv} 
\bibliography{references}

@article{cao2020review,
	title        = {A review of artificial intelligence for EEG-based brain- computer interfaces and applications},
	author       = {Cao, Zehong},
	year         = 2020,
	journal      = {Brain Science Advances},
	publisher    = {SAGE Publications Sage UK: London, England},
	volume       = 6,
	number       = 3,
	pages        = {162--170}
}

@article{goodfellow2020generative,
  title={Generative adversarial networks},
  author={Goodfellow, Ian and Pouget-Abadie, Jean and Mirza, Mehdi and Xu, Bing and Warde-Farley, David and Ozair, Sherjil and Courville, Aaron and Bengio, Yoshua},
  journal={Communications of the ACM},
  volume={63},
  number={11},
  pages={139--144},
  year={2020},
  publisher={ACM New York, NY, USA}
}

@article{vaswani2017attention,
  title={Attention is all you need},
  author={Vaswani, Ashish and Shazeer, Noam and Parmar, Niki and Uszkoreit, Jakob and Jones, Llion and Gomez, Aidan N and Kaiser, {\L}ukasz and Polosukhin, Illia},
  journal={Advances in neural information processing systems},
  volume={30},
  year={2017}
}

@inproceedings{papineni2002bleu,
	title        = {Bleu: a method for automatic evaluation of machine translation},
	author       = {Papineni, Kishore and Roukos, Salim and Ward, Todd and Zhu, Wei-Jing},
	year         = 2002,
	booktitle    = {Proceedings of the 40th annual meeting of the Association for Computational Linguistics},
	pages        = {311--318}
}

@inproceedings{banerjee2005meteor,
	title        = {METEOR: An automatic metric for MT evaluation with improved correlation with human judgments},
	author       = {Banerjee, Satanjeev and Lavie, Alon},
	year         = 2005,
	booktitle    = {Proceedings of the acl workshop on intrinsic and extrinsic evaluation measures for machine translation and/or summarization},
	pages        = {65--72}
}

@article{zhang2019bertscore,
	title        = {Bertscore: Evaluating text generation with bert},
	author       = {Zhang, Tianyi and Kishore, Varsha and Wu, Felix and Weinberger, Kilian Q and Artzi, Yoav},
	year         = 2019,
	journal      = {arXiv preprint arXiv:1904.09675}
}

@article{liu2024eeg2text,
	title        = {EEG2TEXT: Open Vocabulary EEG-to-Text Decoding with EEG Pre-Training and Multi-View Transformer},
	author       = {Liu, Hanwen and Hajialigol, Daniel and Antony, Benny and Han, Aiguo and Wang, Xuan},
	year         = 2024,
	journal      = {arXiv preprint arXiv:2405.02165}
}

@inproceedings{tirupattur2018thoughtviz,
	title        = {Thoughtviz: Visualizing human thoughts using generative adversarial network},
	author       = {Tirupattur, Praveen and Rawat, Yogesh Singh and Spampinato, Concetto and Shah, Mubarak},
	year         = 2018,
	booktitle    = {Proceedings of the 26th ACM international conference on Multimedia},
	pages        = {950--958}
}

@inproceedings{spampinato2017deep,
	title        = {Deep learning human mind for automated visual classification},
	author       = {Spampinato, Concetto and Palazzo, Simone and Kavasidis, Isaak and Giordano, Daniela and Souly, Nasim and Shah, Mubarak},
	year         = 2017,
	booktitle    = {Proceedings of the IEEE conference on computer vision and pattern recognition},
	pages        = {6809--6817}
}

@article{lan2023seeing,
  title={Seeing through the brain: image reconstruction of visual perception from human brain signals},
  author={Lan, Yu-Ting and Ren, Kan and Wang, Yansen and Zheng, Wei-Long and Li, Dongsheng and Lu, Bao-Liang and Qiu, Lili},
  journal={arXiv preprint arXiv:2308.02510},
  year={2023}
}

@article{palazzo2020decoding,
	title        = {Decoding brain representations by multimodal learning of neural activity and visual features},
	author       = {Palazzo, Simone and Spampinato, Concetto and Kavasidis, Isaak and Giordano, Daniela and Schmidt, Joseph and Shah, Mubarak},
	year         = 2020,
	journal      = {IEEE Transactions on Pattern Analysis and Machine Intelligence},
	publisher    = {IEEE},
	volume       = 43,
	number       = 11,
	pages        = {3833--3849}
}

@inproceedings{kavasidis2017brain2image,
	title        = {Brain2image: Converting brain signals into images},
	author       = {Kavasidis, Isaak and Palazzo, Simone and Spampinato, Concetto and Giordano, Daniela and Shah, Mubarak},
	year         = 2017,
	booktitle    = {Proceedings of the 25th ACM international conference on Multimedia},
	pages        = {1809--1817}
}

@article{feng2023aligning,
	title        = {Aligning semantic in brain and language: A curriculum contrastive method for electroencephalography-to-text generation},
	author       = {Feng, Xiachong and Feng, Xiaocheng and Qin, Bing and Liu, Ting},
	year         = 2023,
	journal      = {IEEE Transactions on Neural Systems and Rehabilitation Engineering},
	publisher    = {IEEE}
}

@inproceedings{singh2023eeg2image,
	title        = {EEG2IMAGE: image reconstruction from EEG brain signals},
	author       = {Singh, Prajwal and Pandey, Pankaj and Miyapuram, Krishna and Raman, Shanmuganathan},
	year         = 2023,
	booktitle    = {ICASSP 2023-2023 IEEE International Conference on Acoustics, Speech and Signal Processing (ICASSP)},
	pages        = {1--5},
	organization = {IEEE}
}

@article{hollenstein2019zuco,
	title        = {ZuCo 2.0: A dataset of physiological recordings during natural reading and annotation},
	author       = {Hollenstein, Nora and Troendle, Marius and Zhang, Ce and Langer, Nicolas},
	year         = 2019,
	journal      = {arXiv preprint arXiv:1912.00903}
}

@inproceedings{bhattasali2020alice,
	title        = {The Alice Datasets: fMRI \& EEG observations of natural language comprehension},
	author       = {Bhattasali, Shohini and Brennan, Jonathan and Luh, Wen-Ming and Franzluebbers, Berta and Hale, John},
	year         = 2020,
	booktitle    = {Proceedings of the Twelfth Language Resources and Evaluation Conference},
	pages        = {120--125}
}

@article{hollenstein2018zuco,
	title        = {ZuCo, a simultaneous EEG and eye-tracking resource for natural sentence reading},
	author       = {Hollenstein, Nora and Rotsztejn, Jonathan and Troendle, Marius and Pedroni, Andreas and Zhang, Ce and Langer, Nicolas},
	year         = 2018,
	journal      = {Scientific data},
	publisher    = {Nature Publishing Group},
	volume       = 5,
	number       = 1,
	pages        = {1--13}
}

@inproceedings{biswal2019eegtotext,
	title        = {Eegtotext: learning to write medical reports from eeg recordings},
	author       = {Biswal, Siddharth and Xiao, Cao and Westover, M Brandon and Sun, Jimeng},
	year         = 2019,
	booktitle    = {Machine Learning for Healthcare Conference},
	pages        = {513--531},
	organization = {PMLR}
}

@article{srivastava2020think2type,
	title        = {Think2Type: Thoughts to Text using EEG Waves},
	author       = {Srivastava, Aditya and Shinde, Tanvi},
	year         = 2020,
	journal      = {International Journal of Engineering Research \& Technology (IJERT)},
	volume       = 9,
	number       = {06},
	pages        = {2278--018}
}

@inproceedings{wang2022open,
	title        = {Open vocabulary electroencephalography-to-text decoding and zero-shot sentiment classification},
	author       = {Wang, Zhenhailong and Ji, Heng},
	year         = 2022,
	booktitle    = {Proceedings of the AAAI Conference on Artificial Intelligence},
	volume       = 36,
	pages        = {5350--5358}
}

@article{duan2023dewave,
	title        = {Dewave: Discrete eeg waves encoding for brain dynamics to text translation},
	author       = {Duan, Yiqun and Zhou, Jinzhao and Wang, Zhen and Wang, Yu-Kai and Lin, Chin-Teng},
	year         = 2023,
	journal      = {arXiv preprint arXiv:2309.14030}
}

@article{yang2023thoughts,
	title        = {Thoughts of brain EEG signal-to-text conversion using weighted feature fusion-based multiscale dilated adaptive DenseNet with attention mechanism},
	author       = {Yang, Jing and Awais, Muhammad and Hossain, Md Amzad and Yee, Lip and Haowei, Ma and Mehedi, Ibrahim M and Iskanderani, AIM},
	year         = 2023,
	journal      = {Biomedical Signal Processing and Control},
	publisher    = {Elsevier},
	volume       = 86,
	pages        = 105120
}

@article{wang2024enhancing,
	title        = {Enhancing EEG-to-Text Decoding through Transferable Representations from Pre-trained Contrastive EEG-Text Masked Autoencoder},
	author       = {Wang, Jiaqi and Song, Zhenxi and Ma, Zhengyu and Qiu, Xipeng and Zhang, Min and Zhang, Zhiguo},
	year         = 2024,
	journal      = {arXiv preprint arXiv:2402.17433}
}

@article{amrani2024deep,
	title        = {Deep Representation Learning for Open Vocabulary Electroencephalography-to-Text Decoding},
	author       = {Amrani, Hamza and Micucci, Daniela and Napoletano, Paolo},
	year         = 2024,
	journal      = {IEEE Journal of Biomedical and Health Informatics},
	publisher    = {IEEE}
}

@article{tao2024see,
	title        = {SEE: Semantically Aligned EEG-to-Text Translation},
	author       = {Tao, Yitian and Liang, Yan and Wang, Luoyu and Li, Yongqing and Yang, Qing and Zhang, Han},
	year         = 2024,
	journal      = {arXiv preprint arXiv:2409.16312}
}

@article{mishra2024thought2text,
	title        = {Thought2Text: Text Generation from EEG Signal using Large Language Models (LLMs)},
	author       = {Mishra, Abhijit and Shukla, Shreya and Torres, Jose and Gwizdka, Jacek and Roychowdhury, Shounak},
	year         = 2024,
	journal      = {arXiv preprint arXiv:2410.07507}
}

@article{rathod2024folded,
	title        = {Folded ensemble deep learning based text generation on the brain signal},
	author       = {Rathod, Vasundhara S and Tiwari, Ashish and Kakde, Omprakash G},
	year         = 2024,
	journal      = {Multimedia Tools and Applications},
	publisher    = {Springer},
	pages        = {1--29}
}

@article{chen2025decodingtext,
	title        = {Decoding text from electroencephalography signals: A novel Hierarchical Gated Recurrent Unit with Masked Residual Attention Mechanism},
	author       = {Chen, Qiupu and Wang, Yimou and Wang, Fenmei and Sun, Duolin and Li, Qiankun},
	year         = 2025,
	journal      = {Engineering Applications of Artificial Intelligence},
	publisher    = {Elsevier},
	volume       = 139,
	pages        = 109615
}

@misc{nemrodovneural,
	title        = {The neural dynamics of facial identity processing: insights from EEG-based pattern analysis and image reconstruction},
	author       = {Nemrodov, D and Niemeier, M and Patel, A and Nestor, A},
	year         = 2018,
	journal      = {eNeuro. 2018; 5: ENEURO. 0358-17.2018}
}

@article{li2020semi,
	title        = {Semi-supervised cross-modal image generation with generative adversarial networks},
	author       = {Li, Dan and Du, Changde and He, Huiguang},
	year         = 2020,
	journal      = {Pattern Recognition},
	publisher    = {Elsevier},
	volume       = 100,
	pages        = 107085
}

@article{wakita2021photorealistic,
	title        = {Photorealistic reconstruction of visual texture from EEG signals},
	author       = {Wakita, Suguru and Orima, Taiki and Motoyoshi, Isamu},
	year         = 2021,
	journal      = {Frontiers in Computational Neuroscience},
	publisher    = {Frontiers Media SA},
	volume       = 15,
	pages        = 754587
}

@article{bai2306dreamdiffusion,
	title        = {Dreamdiffusion: Generating high-quality images from brain eeg signals},
	author       = {Bai, Yunpeng and Wang, Xintao and Cao, Yan-pei and Ge, Yixiao and Yuan, Chun and Shan, Ying},
	year         = 2023,
	journal      = {arXiv preprint arXiv:2306.16934}
}

@article{khaleghi2022visual,
	title        = {Visual saliency and image reconstruction from EEG signals via an effective geometric deep network-based generative adversarial network},
	author       = {Khaleghi, Nastaran and Rezaii, Tohid Yousefi and Beheshti, Soosan and Meshgini, Saeed and Sheykhivand, Sobhan and Danishvar, Sebelan},
	year         = 2022,
	journal      = {Electronics},
	publisher    = {MDPI},
	volume       = 11,
	number       = 21,
	pages        = 3637
}

@misc{shimizu2022improving,
	title        = {Improving classification and reconstruction of imagined images from EEG signals. bioRxiv. Retrieved July 5, 2022},
	author       = {Shimizu, H and Srinivasan, R},
	year         = 2022
}

@article{zeng2023dm,
	title        = {DM-RE2I: A framework based on diffusion model for the reconstruction from EEG to image},
	author       = {Zeng, Hong and Xia, Nianzhang and Qian, Dongguan and Hattori, Motonobu and Wang, Chu and Kong, Wanzeng},
	year         = 2023,
	journal      = {Biomedical Signal Processing and Control},
	publisher    = {Elsevier},
	volume       = 86,
	pages        = 105125
}

@article{mishra2023neurogan,
	title        = {NeuroGAN: image reconstruction from EEG signals via an attention-based GAN},
	author       = {Mishra, Rahul and Sharma, Krishan and Jha, Ranjeet Ranjan and Bhavsar, Arnav},
	year         = 2023,
	journal      = {Neural Computing and Applications},
	publisher    = {Springer},
	volume       = 35,
	number       = 12,
	pages        = {9181--9192}
}

@inproceedings{singh2024learning,
	title        = {Learning Robust Deep Visual Representations from EEG Brain Recordings},
	author       = {Singh, Prajwal and Dalal, Dwip and Vashishtha, Gautam and Miyapuram, Krishna and Raman, Shanmuganathan},
	year         = 2024,
	booktitle    = {Proceedings of the IEEE/CVF Winter Conference on Applications of Computer Vision},
	pages        = {7553--7562}
}

@article{ahmadieh2024visual,
	title        = {Visual image reconstruction based on EEG signals using a generative adversarial and deep fuzzy neural network},
	author       = {Ahmadieh, Hajar and Gassemi, Farnaz and Moradi, Mohammad Hasan},
	year         = 2024,
	journal      = {Biomedical Signal Processing and Control},
	publisher    = {Elsevier},
	volume       = 87,
	pages        = 105497
}

@article{song2023decoding,
	title        = {Decoding Natural Images from EEG for Object Recognition},
	author       = {Song, Yonghao and Liu, Bingchuan and Li, Xiang and Shi, Nanlin and Wang, Yijun and Gao, Xiaorong},
	year         = 2023,
	journal      = {arXiv preprint arXiv:2308.13234}
}

@inproceedings{sugimoto2024image,
	title        = {Image Generation using EEG data: A Contrastive Learning based Approach},
	author       = {Sugimoto, Yuma and Pongthanisorn, Goragod and Capi, Genci},
	year         = 2024,
	booktitle    = {2024 IEEE Canadian Conference on Electrical and Computer Engineering (CCECE)},
	pages        = {794--798},
	organization = {IEEE}
}

@article{li2024visual,
	title        = {Visual decoding and reconstruction via eeg embeddings with guided diffusion},
	author       = {Li, Dongyang and Wei, Chen and Li, Shiying and Zou, Jiachen and Qin, Haoyang and Liu, Quanying},
	year         = 2024,
	journal      = {arXiv preprint arXiv:2403.07721}
}

@inproceedings{krishna2021advancing,
	title        = {Advancing speech synthesis using EEG},
	author       = {Krishna, Gautam and Tran, Co and Carnahan, Mason and Tewfik, Ahmed H},
	year         = 2021,
	booktitle    = {2021 10th International IEEE/EMBS Conference on Neural Engineering (NER)},
	pages        = {199--204},
	organization = {IEEE}
}

@article{ramirez2022eeg2mel,
	title        = {EEG2Mel: Reconstructing sound from brain responses to music},
	author       = {Ramirez-Aristizabal, Adolfo G and Kello, Chris},
	year         = 2022,
	journal      = {arXiv preprint arXiv:2207.13845}
}

@inproceedings{lee2023towards,
	title        = {Towards voice reconstruction from EEG during imagined speech},
	author       = {Lee, Young-Eun and Lee, Seo-Hyun and Kim, Sang-Ho and Lee, Seong-Whan},
	year         = 2023,
	booktitle    = {Proceedings of the AAAI Conference on Artificial Intelligence},
	volume       = 37,
	pages        = {6030--6038}
}

@inproceedings{lee2023speech,
	title        = {Speech Synthesis from Brain Signals Based on Generative Model},
	author       = {Lee, Young-Eun and Kim, Sang-Ho and Lee, Seo-Hyun and Lee, Jung-Sun and Kim, Soowon and Lee, Seong-Whan},
	year         = 2023,
	booktitle    = {2023 11th International Winter Conference on Brain-Computer Interface (BCI)},
	pages        = {1--4},
	organization = {IEEE}
}

@inproceedings{park2024towards,
	title        = {Towards EEG-based Talking-face Generation for Brain Signal-driven Dynamic Communication},
	author       = {Park, Ji-Ha and Lee, Seo-Hyun and Lee, Seong-Whan},
	year         = 2024,
	booktitle    = {2024 46th Annual International Conference of the IEEE Engineering in Medicine and Biology Society (EMBC)},
	pages        = {1--5},
	organization = {IEEE}
}

@article{postolache2024naturalistic,
	title        = {Naturalistic Music Decoding from EEG Data via Latent Diffusion Models},
	author       = {Postolache, Emilian and Polouliakh, Natalia and Kitano, Hiroaki and Connelly, Akima and Rodol{\`a}, Emanuele and Cosmo, Luca and Akama, Taketo},
	year         = 2024,
	journal      = {arXiv preprint arXiv:2405.09062}
}

@article{jiang2024eeg,
	title        = {EEG-driven automatic generation of emotive music based on transformer},
	author       = {Jiang, Hui and Chen, Yu and Wu, Di and Yan, Jinlin},
	year         = 2024,
	journal      = {Frontiers in Neurorobotics},
	publisher    = {Frontiers Media SA},
	volume       = 18,
	pages        = 1437737
}

@inproceedings{mizuno2024investigation,
	title        = {An Investigation on the Speech Recovery from EEG Signals Using Transformer},
	author       = {Mizuno, Tomoaki and Kishida, Takuya and Yoshimura, Natsue and Nakashika, Toru},
	year         = 2024,
	booktitle    = {2024 Asia Pacific Signal and Information Processing Association Annual Summit and Conference (APSIPA ASC)},
	pages        = {1--6},
	organization = {IEEE}
}

@article{lawhern2018eegnet,
	title        = {EEGNet: a compact convolutional neural network for EEG-based brain--computer interfaces},
	author       = {Lawhern, Vernon J and Solon, Amelia J and Waytowich, Nicholas R and Gordon, Stephen M and Hung, Chou P and Lance, Brent J},
	year         = 2018,
	journal      = {Journal of neural engineering},
	publisher    = {iOP Publishing},
	volume       = 15,
	number       = 5,
	pages        = {056013}
}

@inproceedings{bria2021sinc,
	title        = {Sinc-based convolutional neural networks for EEG-BCI-based motor imagery classification},
	author       = {Bria, Alessandro and Marrocco, Claudio and Tortorella, Francesco},
	year         = 2021,
	booktitle    = {International Conference on Pattern Recognition},
	pages        = {526--535},
	organization = {Springer}
}

@article{kong2020hifi,
	title        = {Hifi-gan: Generative adversarial networks for efficient and high fidelity speech synthesis},
	author       = {Kong, Jungil and Kim, Jaehyeon and Bae, Jaekyoung},
	year         = 2020,
	journal      = {Advances in neural information processing systems},
	volume       = 33,
	pages        = {17022--17033}
}

@article{hsu2021hubert,
	title        = {Hubert: Self-supervised speech representation learning by masked prediction of hidden units},
	author       = {Hsu, Wei-Ning and Bolte, Benjamin and Tsai, Yao-Hung Hubert and Lakhotia, Kushal and Salakhutdinov, Ruslan and Mohamed, Abdelrahman},
	year         = 2021,
	journal      = {IEEE/ACM transactions on audio, speech, and language processing},
	publisher    = {IEEE},
	volume       = 29,
	pages        = {3451--3460}
}

@inproceedings{prajwal2020lip,
	title        = {A lip sync expert is all you need for speech to lip generation in the wild},
	author       = {Prajwal, KR and Mukhopadhyay, Rudrabha and Namboodiri, Vinay P and Jawahar, CV},
	year         = 2020,
	booktitle    = {Proceedings of the 28th ACM international conference on multimedia},
	pages        = {484--492}
}

@inproceedings{zhang2023adding,
	title        = {Adding conditional control to text-to-image diffusion models},
	author       = {Zhang, Lvmin and Rao, Anyi and Agrawala, Maneesh},
	year         = 2023,
	booktitle    = {Proceedings of the IEEE/CVF International Conference on Computer Vision},
	pages        = {3836--3847}
}

@article{liu2024audioldm,
	title        = {Audioldm 2: Learning holistic audio generation with self-supervised pretraining},
	author       = {Liu, Haohe and Yuan, Yi and Liu, Xubo and Mei, Xinhao and Kong, Qiuqiang and Tian, Qiao and Wang, Yuping and Wang, Wenwu and Wang, Yuxuan and Plumbley, Mark D},
	year         = 2024,
	journal      = {IEEE/ACM Transactions on Audio, Speech, and Language Processing},
	publisher    = {IEEE}
}

@article{chen2025decoding,
  title={Decoding EEG speech perception with transformers and VAE-based data augmentation},
  author={Chen, Terrance Yu-Hao and Chen, Yulin and Soederhaell, Pontus and Agrawal, Sadrishya and Shapovalenko, Kateryna},
  journal={arXiv preprint arXiv:2501.04359},
  year={2025}
}

@inproceedings{lin2004rouge,
	title        = {Rouge: A package for automatic evaluation of summaries},
	author       = {Lin, Chin-Yew},
	year         = 2004,
	booktitle    = {Text summarization branches out},
	pages        = {74--81}
}

@article{sellam2020bleurt,
	title        = {BLEURT: Learning robust metrics for text generation},
	author       = {Sellam, Thibault and Das, Dipanjan and Parikh, Ankur P},
	year         = 2020,
	journal      = {arXiv preprint arXiv:2004.04696}
}

@article{salimans2016improved,
	title        = {Improved techniques for training gans},
	author       = {Salimans, Tim and Goodfellow, Ian and Zaremba, Wojciech and Cheung, Vicki and Radford, Alec and Chen, Xi},
	year         = 2016,
	journal      = {Advances in neural information processing systems},
	volume       = 29
}

@article{zeng2023dcae,
	title        = {DCAE: A dual conditional autoencoder framework for the reconstruction from EEG into image},
	author       = {Zeng, Hong and Xia, Nianzhang and Tao, Ming and Pan, Deng and Zheng, Haohao and Wang, Chu and Xu, Feifan and Zakaria, Wael and Dai, Guojun},
	year         = 2023,
	journal      = {Biomedical Signal Processing and Control},
	publisher    = {Elsevier},
	volume       = 81,
	pages        = 104440
}

@article{xu2024alljoined,
	title        = {Alljoined--A dataset for EEG-to-Image decoding},
	author       = {Xu, Jonathan and Aristimunha, Bruno and Feucht, Max Emanuel and Qian, Emma and Liu, Charles and Shahjahan, Tazik and Spyra, Martyna and Zhang, Steven Zifan and Short, Nicholas and Kim, Jioh and others},
	year         = 2024,
	journal      = {arXiv preprint arXiv:2404.05553}
}

@article{kumar2018envisioned,
	title        = {Envisioned speech recognition using EEG sensors},
	author       = {Kumar, Pradeep and Saini, Rajkumar and Roy, Partha Pratim and Sahu, Pawan Kumar and Dogra, Debi Prosad},
	year         = 2018,
	journal      = {Personal and Ubiquitous Computing},
	publisher    = {Springer},
	volume       = 22,
	pages        = {185--199}
}

@article{orima2021analysis,
	title        = {Analysis and synthesis of natural texture perception from visual evoked potentials},
	author       = {Orima, Taiki and Motoyoshi, Isamu},
	year         = 2021,
	journal      = {Frontiers in Neuroscience},
	publisher    = {Frontiers Media SA},
	volume       = 15,
	pages        = 698940
}

@article{hebart2019things,
	title        = {THINGS: A database of 1,854 object concepts and more than 26,000 naturalistic object images},
	author       = {Hebart, Martin N and Dickter, Adam H and Kidder, Alexis and Kwok, Wan Y and Corriveau, Anna and Van Wicklin, Caitlin and Baker, Chris I},
	year         = 2019,
	journal      = {PloS one},
	publisher    = {Public Library of Science San Francisco, CA USA},
	volume       = 14,
	number       = 10,
	pages        = {e0223792}
}

@article{grootswagers2022human,
	title        = {Human EEG recordings for 1,854 concepts presented in rapid serial visual presentation streams},
	author       = {Grootswagers, Tijl and Zhou, Ivy and Robinson, Amanda K and Hebart, Martin N and Carlson, Thomas A},
	year         = 2022,
	journal      = {Scientific Data},
	publisher    = {Nature Publishing Group UK London},
	volume       = 9,
	number       = 1,
	pages        = 3
}

@article{kaneshiro2015representational,
	title        = {A representational similarity analysis of the dynamics of object processing using single-trial EEG classification},
	author       = {Kaneshiro, Blair and Perreau Guimaraes, Marcos and Kim, Hyung-Suk and Norcia, Anthony M and Suppes, Patrick},
	year         = 2015,
	journal      = {Plos one},
	publisher    = {Public Library of Science San Francisco, CA USA},
	volume       = 10,
	number       = 8,
	pages        = {e0135697}
}

@inproceedings{losorelli2017nmed,
	title        = {NMED-T: A Tempo-Focused Dataset of Cortical and Behavioral Responses to Naturalistic Music.},
	author       = {Losorelli, Steven and Nguyen, Duc T and Dmochowski, Jacek P and Kaneshiro, Blair},
	year         = 2017,
	booktitle    = {ISMIR},
	volume       = 3,
	pages        = 5
}

@misc{kaneshiro2016naturalistic,
  title={Naturalistic music EEG dataset—hindi (nmed-h)},
  author={Kaneshiro, Blair and Nguyen, Duc T and Dmochowski, Jacek P and Norcia, Anthony M and Berger, Jonathan},
  howpublished={Stanford Digital Repository},
  year={2016},
  publisher={Stanford Digit. Repository}
}

@inproceedings{zhao2015classifying,
	title        = {Classifying phonological categories in imagined and articulated speech},
	author       = {Zhao, Shunan and Rudzicz, Frank},
	year         = 2015,
	booktitle    = {2015 IEEE international conference on acoustics, speech and signal processing (ICASSP)},
	pages        = {992--996},
	organization = {IEEE}
}

@article{ikegawa2024text,
	title        = {Text and image generation from intracranial electroencephalography using an embedding space for text and images},
	author       = {Ikegawa, Yuya and Fukuma, Ryohei and Sugano, Hidenori and Oshino, Satoru and Tani, Naoki and Tamura, Kentaro and Iimura, Yasushi and Suzuki, Hiroharu and Yamamoto, Shota and Fujita, Yuya and others},
	year         = 2024,
	journal      = {Journal of Neural Engineering},
	publisher    = {IOP Publishing},
	volume       = 21,
	number       = 3,
	pages        = {036019}
}

@article{zander2010enhancing,
  title={Enhancing human-computer interaction with input from active and passive brain-computer interfaces},
  author={Zander, Thorsten O and Kothe, Christian and Jatzev, Sabine and Gaertner, Matti},
  journal={Brain-computer interfaces: Applying our minds to human-computer interaction},
  pages={181--199},
  year={2010},
  publisher={Springer}
}

@incollection{wolpaw2010brain,
  title={Brain signals for brain--computer interfaces},
  author={Wolpaw, Jonathan R and Boulay, Chadwick B},
  booktitle={Brain-Computer Interfaces: Revolutionizing Human-Computer Interaction},
  pages={29--46},
  year={2010},
  publisher={Springer}
}

@article{broderick2019semantic,
  title={Semantic context enhances the early auditory encoding of natural speech},
  author={Broderick, Michael P and Anderson, Andrew J and Lalor, Edmund C},
  journal={Journal of Neuroscience},
  volume={39},
  number={38},
  pages={7564--7575},
  year={2019},
  publisher={Society for Neuroscience}
}

@article{elgedawy2025bridging,
  title={Bridging Brain Signals and Language: A Deep Learning Approach to EEG-to-Text Decoding},
  author={Gedawy, Mostafa El and Nabil, Omnia and Mamdouh, Omar and Nady, Mahmoud and Adel, Nour Alhuda and Fares, Ahmed},
  journal={arXiv preprint arXiv:2502.17465},
  year={2025}
}

@article{masry2025ets,
  title={ETS: Open Vocabulary Electroencephalography-To-Text Decoding and Sentiment Classification},
  author={Masry, Mohamed and Amen, Mohamed and Elzyat, Mohamed and Hamed, Mohamed and Magdy, Norhan and Khaled, Maram},
  journal={arXiv preprint arXiv:2506.14783},
  year={2025}
}

@article{liu2025learning,
  title={Learning Interpretable Representations Leads to Semantically Faithful EEG-to-Text Generation},
  author={Liu, Xiaozhao and Shen, Dinggang and Liu, Xihui},
  journal={arXiv preprint arXiv:2505.17099},
  year={2025}
}

@article{lu2025eeg2text,
  title={EEG2TEXT-CN: An Exploratory Study of Open-Vocabulary Chinese Text-EEG Alignment via Large Language Model and Contrastive Learning on ChineseEEG},
  author={Lu, Jacky Tai-Yu and Chiang, Jung and Chen, Chi-Sheng and Tung, Anna Nai-Yun and Hu, Hsiang Wei and Cheng, Yuan Chiao},
  journal={arXiv preprint arXiv:2506.00854},
  year={2025}
}

@article{jiang2025neural,
  title={Neural Spelling: A Spell-Based BCI System for Language Neural Decoding},
  author={Jiang, Xiaowei and Zhou, Charles and Duan, Yiqun and Zhao, Ziyi and Do, Thomas and Lin, Chin-Teng},
  journal={arXiv preprint arXiv:2501.17489},
  year={2025}
}

@article{weiss2012too,
  title={“Too many betas do not spoil the broth”: the role of beta brain oscillations in language processing},
  author={Weiss, Sabine and Mueller, Horst M},
  journal={Frontiers in psychology},
  volume={3},
  pages={201},
  year={2012},
  publisher={Frontiers Research Foundation}
}

@article{bastiaansen2006oscillatory,
  title={Oscillatory neuronal dynamics during language comprehension},
  author={Bastiaansen, Marcel and Hagoort, Peter},
  journal={Progress in brain research},
  volume={159},
  pages={179--196},
  year={2006},
  publisher={Elsevier}
}

@article{guenther2024image,
  title={Image classification and reconstruction from low-density EEG},
  author={Guenther, Sven and Kosmyna, Nataliya and Maes, Pattie},
  journal={Scientific Reports},
  volume={14},
  number={1},
  pages={16436},
  year={2024},
  publisher={Nature Publishing Group UK London}
}

@inproceedings{lopez2025guess,
  title={Guess what i think: Streamlined EEG-to-image generation with latent diffusion models},
  author={Lopez, Eleonora and Sigillo, Luigi and Colonnese, Federica and Panella, Massimo and Comminiello, Danilo},
  booktitle={ICASSP 2025-2025 IEEE International Conference on Acoustics, Speech and Signal Processing (ICASSP)},
  pages={1--5},
  year={2025},
  organization={IEEE}
}

@inproceedings{mehmood2025catvis,
  title={CATVis: Context-Aware Thought Visualization},
  author={Mehmood, Tariq and Ahmad, Hamza and Shakeel, Muhammad Haroon and Taj, Murtaza},
  booktitle={International Conference on Medical Image Computing and Computer-Assisted Intervention},
  pages={98--108},
  year={2025},
  organization={Springer}
}

@article{rezvani2025interpretable,
  title={Interpretable EEG-to-Image Generation with Semantic Prompts},
  author={Rezvani, Arshak and Akbari, Ali and Arani, Kosar Sanjar and Mirian, Maryam and Arasteh, Emad and McKeown, Martin J},
  journal={arXiv preprint arXiv:2507.07157},
  year={2025}
}

@article{cheng2025fine,
  title={Fine-grained image generation with EEG multi-level semantics},
  author={Cheng, Wenjie and Tan, Jun and Wang, Lizhi and Herrero, Mar{\'\i}a Trinidad and Zeng, Hong},
  journal={Computer Methods and Programs in Biomedicine},
  pages={108909},
  year={2025},
  publisher={Elsevier}
}

@article{xiong2025synthesizing,
  title={Synthesizing intelligible utterances from EEG of imagined speech},
  author={Xiong, Wenjing and Ma, Lin and Li, Haifeng},
  journal={Frontiers in Neuroscience},
  volume={19},
  pages={1565848},
  year={2025},
  publisher={Frontiers Media SA}
}

@article{toffolo2022evoking,
  title={Evoking the N400 event-related potential (ERP) component using a publicly available novel set of sentences with semantically incongruent or congruent eggplants (endings)},
  author={Toffolo, Kathryn K and Freedman, Edward G and Foxe, John J},
  journal={Neuroscience},
  volume={501},
  pages={143--158},
  year={2022},
  publisher={Elsevier}
}

@article{lee2024toward,
  title={Toward fully-end-to-end listened speech decoding from eeg signals},
  author={Lee, Jihwan and Kommineni, Aditya and Feng, Tiantian and Avramidis, Kleanthis and Shi, Xuan and Kadiri, Sudarsana and Narayanan, Shrikanth},
  journal={arXiv preprint arXiv:2406.08644},
  year={2024}
}

@inproceedings{lee2025enhancing,
  title={Enhancing listened speech decoding from eeg via parallel phoneme sequence prediction},
  author={Lee, Jihwan and Feng, Tiantian and Kommineni, Aditya and Kadiri, Sudarsana Reddy and Narayanan, Shrikanth},
  booktitle={ICASSP 2025-2025 IEEE International Conference on Acoustics, Speech and Signal Processing (ICASSP)},
  pages={1--5},
  year={2025},
  organization={IEEE}
}

@article{zhang2024chisco,
  title={Chisco: An EEG-based BCI dataset for decoding of imagined speech},
  author={Zhang, Zihan and Ding, Xiao and Bao, Yu and Zhao, Yi and Liang, Xia and Qin, Bing and Liu, Ting},
  journal={Scientific Data},
  volume={11},
  number={1},
  pages={1265},
  year={2024},
  publisher={Nature Publishing Group UK London}
}

@article{alharbi2024decoding,
  title={Decoding imagined speech from EEG data: A hybrid deep learning approach to capturing spatial and temporal features},
  author={Alharbi, Yasser F and Alotaibi, Yousef A},
  journal={Life},
  volume={14},
  number={11},
  pages={1501},
  year={2024},
  publisher={MDPI}
}

@article{krause1997relative,
  title={Relative alpha desynchronization and synchronization during speech perception},
  author={Krause, Christina M and P{\"o}rn, Bodil and Lang, A Heikki and Laine, Matti},
  journal={Cognitive brain research},
  volume={5},
  number={4},
  pages={295--299},
  year={1997},
  publisher={Elsevier}
}

@article{pfurtscheller1994event,
  title={Event-related desynchronization (ERD) during visual processing},
  author={Pfurtscheller, Gert and Neuper, Chrita and Mohl, W},
  journal={International journal of psychophysiology},
  volume={16},
  number={2-3},
  pages={147--153},
  year={1994},
  publisher={Elsevier}
}

@article{krause1996event,
  title={Event-related. EEG desynchronization and synchronization during an auditory memory task},
  author={Krause, Christina M and Lang, A Heikki and Laine, Matti and Kuusisto, Mika and P{\"o}rn, Bodil},
  journal={Electroencephalography and clinical neurophysiology},
  volume={98},
  number={4},
  pages={319--326},
  year={1996},
  publisher={Elsevier}
}

@article{di2018experimental,
  title={An experimental study on EEG characteristics induced by intermittent pure tone stimuli at different frequencies},
  author={Di, Guo-Qing and Fan, Meng-Chi and Lin, Qin-Hao},
  journal={Applied Acoustics},
  volume={141},
  pages={46--53},
  year={2018},
  publisher={Elsevier}
}

@article{farahani2021brain,
  title={Brain mapping of auditory steady-state responses: A broad view of cortical and subcortical sources},
  author={Farahani, Ehsan Darestani and Wouters, Jan and van Wieringen, Astrid},
  journal={Human brain mapping},
  volume={42},
  number={3},
  pages={780--796},
  year={2021},
  publisher={Wiley Online Library}
}

@article{herdman2002intracerebral,
  title={Intracerebral sources of human auditory steady-state responses},
  author={Herdman, Anthony T and Lins, Otavio and Van Roon, Patricia and Stapells, David R and Scherg, Michael and Picton, Terence W},
  journal={Brain topography},
  volume={15},
  number={2},
  pages={69--86},
  year={2002},
  publisher={Springer}
}

@inproceedings{odena2018generator,
  title={Is generator conditioning causally related to GAN performance?},
  author={Odena, Augustus and Buckman, Jacob and Olsson, Catherine and Brown, Tom and Olah, Christopher and Raffel, Colin and Goodfellow, Ian},
  booktitle={International conference on machine learning},
  pages={3849--3858},
  year={2018},
  organization={PMLR}
}

@inproceedings{zhang2019self,
  title={Self-attention generative adversarial networks},
  author={Zhang, Han and Goodfellow, Ian and Metaxas, Dimitris and Odena, Augustus},
  booktitle={International conference on machine learning},
  pages={7354--7363},
  year={2019},
  organization={PMLR}
}

@article{yang2023diffusion,
  title={Diffusion models: A comprehensive survey of methods and applications},
  author={Yang, Ling and Zhang, Zhilong and Song, Yang and Hong, Shenda and Xu, Runsheng and Zhao, Yue and Zhang, Wentao and Cui, Bin and Yang, Ming-Hsuan},
  journal={ACM computing surveys},
  volume={56},
  number={4},
  pages={1--39},
  year={2023},
  publisher={ACM New York, NY, USA}
}

@article{kingma2019introduction,
  title={An introduction to variational autoencoders},
  author={Kingma, Diederik P and Welling, Max and others},
  journal={Foundations and Trends{\textregistered} in Machine Learning},
  volume={12},
  number={4},
  pages={307--392},
  year={2019},
  publisher={Now Publishers, Inc.}
}

@article{ho2020denoising,
  title={Denoising diffusion probabilistic models},
  author={Ho, Jonathan and Jain, Ajay and Abbeel, Pieter},
  journal={Advances in neural information processing systems},
  volume={33},
  pages={6840--6851},
  year={2020}
}

@article{bengio2003neural,
  title={A neural probabilistic language model},
  author={Bengio, Yoshua and Ducharme, R{\'e}jean and Vincent, Pascal and Jauvin, Christian},
  journal={Journal of machine learning research},
  volume={3},
  number={Feb},
  pages={1137--1155},
  year={2003}
}

@article{bastiaansen2002event,
  title={Event-related theta power increases in the human EEG during online sentence processing},
  author={Bastiaansen, Marcel CM and Van Berkum, Jos JA and Hagoort, Peter},
  journal={Neuroscience letters},
  volume={323},
  number={1},
  pages={13--16},
  year={2002},
  publisher={Elsevier}
}

@article{bastiaansen2008see,
  title={I see what you mean: theta power increases are involved in the retrieval of lexical semantic information},
  author={Bastiaansen, Marcel CM and Oostenveld, Robert and Jensen, Ole and Hagoort, Peter},
  journal={Brain and language},
  volume={106},
  number={1},
  pages={15--28},
  year={2008},
  publisher={Elsevier}
}

@article{weiss2005increased,
  title={Increased neuronal communication accompanying sentence comprehension},
  author={Weiss, Sabine and Mueller, Horst M and Schack, Baerbel and King, Jonathan W and Kutas, Martha and Rappelsberger, Peter},
  journal={International journal of psychophysiology},
  volume={57},
  number={2},
  pages={129--141},
  year={2005},
  publisher={Elsevier}
}

@article{hagoort2004integration,
  title={Integration of word meaning and world knowledge in language comprehension},
  author={Hagoort, Peter and Hald, Lea and Bastiaansen, Marcel and Petersson, Karl Magnus},
  journal={science},
  volume={304},
  number={5669},
  pages={438--441},
  year={2004},
  publisher={American Association for the Advancement of Science}
}

@article{weiss2003contribution,
  title={The contribution of EEG coherence to the investigation of language},
  author={Weiss, Sabine and Mueller, Horst M},
  journal={Brain and language},
  volume={85},
  number={2},
  pages={325--343},
  year={2003},
  publisher={Elsevier}
}

@article{haarmann2002neural,
  title={Neural synchronization mediates on-line sentence processing: EEG coherence evidence from filler-gap constructions},
  author={Haarmann, Henk J and Cameron, Katherine A and Ruchkin, Daniel S},
  journal={Psychophysiology},
  volume={39},
  number={6},
  pages={820--825},
  year={2002},
  publisher={Cambridge University Press}
}

@article{fahimi2018semantic,
  title={Semantic features reveal different networks during word processing: An EEG source localization study},
  author={Fahimi Hnazaee, Mansoureh and Khachatryan, Elvira and Van Hulle, Marc M},
  journal={Frontiers in human neuroscience},
  volume={12},
  pages={503},
  year={2018},
  publisher={Frontiers Media SA}
}

@article{marinkovic2004spatiotemporal,
  title={Spatiotemporal dynamics of word processing in the human cortex},
  author={Marinkovi{\'c}, Ksenija},
  journal={The Neuroscientist},
  volume={10},
  number={2},
  pages={142--152},
  year={2004},
  publisher={SAGE Publications}
}

@article{khadir2023brain,
  title={Brain activity characteristics of RGB stimulus: an EEG study},
  author={Khadir, Alireza and Maghareh, Mohammad and Sasani Ghamsari, Shamim and Beigzadeh, Borhan},
  journal={Scientific Reports},
  volume={13},
  number={1},
  pages={18988},
  year={2023},
  publisher={Nature Publishing Group UK London}
}

@article{lutzenberger1995visual,
  title={Visual stimulation alters local 40-Hz responses in humans: an EEG-study},
  author={Lutzenberger, Werner and Pulverm{\"u}ller, Friedemann and Elbert, Thomas and Birbaumer, Niels},
  journal={Neuroscience letters},
  volume={183},
  number={1-2},
  pages={39--42},
  year={1995},
  publisher={Elsevier}
}

@article{demiralp2007gamma,
  title={Gamma amplitudes are coupled to theta phase in human EEG during visual perception},
  author={Demiralp, Tamer and Bayraktaroglu, Zubeyir and Lenz, Daniel and Junge, Stefanie and Busch, Niko A and Maess, Burkhard and Ergen, Mehmet and Herrmann, Christoph S},
  journal={International journal of psychophysiology},
  volume={64},
  number={1},
  pages={24--30},
  year={2007},
  publisher={Elsevier}
}

@article{emanuel2000makes,
  title={What makes clinical research ethical?},
  author={Emanuel, Ezekiel J and Wendler, David and Grady, Christine},
  journal={Jama},
  volume={283},
  number={20},
  pages={2701--2711},
  year={2000},
  publisher={American Medical Association}
}

@inproceedings{paranjape2001electroencephalogram,
  title={The electroencephalogram as a biometric},
  author={Paranjape, RB and Mahovsky, J and Benedicenti, L and Koles, Z},
  booktitle={Canadian Conference on Electrical and Computer Engineering 2001. Conference Proceedings (Cat. No. 01TH8555)},
  volume={2},
  pages={1363--1366},
  year={2001},
  organization={IEEE}
}

@inproceedings{poulos1999person,
  title={Person identification based on parametric processing of the EEG},
  author={Poulos, Marios and Rangoussi, Maria and Chrissikopoulos, Vassilios and Evangelou, Angelos},
  booktitle={ICECS'99. Proceedings of ICECS'99. 6th IEEE international conference on electronics, circuits and systems (cat. No. 99EX357)},
  volume={1},
  pages={283--286},
  year={1999},
  organization={IEEE}
}

@article{botes2025decoding,
  title={Decoding the Brain, Respecting the Person: A Neuroethical Inquiry into Consent and Cognitive Liberty in South Africa},
  author={Botes, Marietjie and Labuschaigne, Melodie and Casteleyn, Camille and Inkster, Becky and Sheppard, Mark},
  journal={Neuroethics},
  volume={18},
  number={3},
  pages={43},
  year={2025},
  publisher={Springer}
}

@inproceedings{brigham2010subject,
  title={Subject identification from electroencephalogram (EEG) signals during imagined speech},
  author={Brigham, Katharine and Kumar, BVK Vijaya},
  booktitle={2010 Fourth IEEE International Conference on Biometrics: Theory, Applications and Systems (BTAS)},
  pages={1--8},
  year={2010},
  organization={IEEE}
}

@article{sankary2023overcoming,
  title={Overcoming barriers to informed consent in neurological research: Perspectives from a national survey},
  author={Sankary, Lauren R and Zelinsky, Megan E and Ford, Paul J and Blackstone, Eric C and Fox, Robert J},
  journal={Research ethics},
  volume={19},
  number={1},
  pages={42--61},
  year={2023},
  publisher={SAGE Publications Sage UK: London, England}
}

@article{ruiz2016cerebre,
  title={CEREBRE: A novel method for very high accuracy event-related potential biometric identification},
  author={Ruiz-Blondet, Maria V and Jin, Zhanpeng and Laszlo, Sarah},
  journal={IEEE Transactions on Information Forensics and Security},
  volume={11},
  number={7},
  pages={1618--1629},
  year={2016},
  publisher={IEEE}
}

@article{ruiz2017permanence,
  title={Permanence of the CEREBRE brain biometric protocol},
  author={Ruiz-Blondet, Maria V and Jin, Zhanpeng and Laszlo, Sarah},
  journal={Pattern Recognition Letters},
  volume={95},
  pages={37--43},
  year={2017},
  publisher={Elsevier}
}

@article{chan2018challenges,
  title={Challenges and future perspectives on electroencephalogram-based biometrics in person recognition},
  author={Chan, Hui-Ling and Kuo, Po-Chih and Cheng, Chia-Yi and Chen, Yong-Sheng},
  journal={Frontiers in neuroinformatics},
  volume={12},
  pages={66},
  year={2018},
  publisher={Frontiers Media SA}
}

@article{kong2019eeg,
  title={EEG fingerprints: phase synchronization of EEG signals as biomarker for subject identification},
  author={Kong, Wanzeng and Wang, Luyun and Xu, Sijia and Babiloni, Fabio and Chen, Hang},
  journal={IEEE Access},
  volume={7},
  pages={121165--121173},
  year={2019},
  publisher={IEEE}
}

@article{hendriks2019ethical,
  title={Ethical challenges of risk, informed consent, and posttrial responsibilities in human research with neural devices: a review},
  author={Hendriks, Saskia and Grady, Christine and Ramos, Khara M and Chiong, Winston and Fins, Joseph J and Ford, Paul and Goering, Sara and Greely, Henry T and Hutchison, Katrina and Kelly, Michael L and others},
  journal={JAMA neurology},
  volume={76},
  number={12},
  pages={1506--1514},
  year={2019}
}

@article{rainey2020brain,
  title={Brain recording, mind-reading, and neurotechnology: ethical issues from consumer devices to brain-based speech decoding},
  author={Rainey, Stephen and Martin, St{\'e}phanie and Christen, Andy and M{\'e}gevand, Pierre and Fourneret, Eric},
  journal={Science and engineering ethics},
  volume={26},
  number={4},
  pages={2295--2311},
  year={2020},
  publisher={Springer}
}

@misc{chile2021law21383,
title = {Ley N.° 21.383: Modifica la Carta Fundamental, para establecer el desarrollo científico y tecnológico al servicio de las personas},
author = {{República de Chile}},
year = {2021},
institution = {Biblioteca del Congreso Nacional de Chile},
url = {https://www.leychile.cl/navegar?idNorma=1166983},
note = {Published October 25, 2021. Accessed July 6, 2026}
}

@misc{unesco2025neurotechnology,
author = {{UNESCO}},
title = {Recommendation on the Ethics of Neurotechnology},
year = {2025},
institution = {United Nations Educational, Scientific and Cultural Organization},
url = {https://www.unesco.org/en/legal-affairs/recommendation-ethics-neurotechnology},
note = {Adopted 11 November 2025}
}

@misc{colorado2024hb1058,
author = {{Colorado General Assembly}},
title = {{HB24-1058: Protect Privacy of Biological Data}},
year = {2024},
url = {https://leg.colorado.gov/bills/hb24-1058},
note = {Signed April 17, 2024}
}

@misc{mindact2025,
author = {{United States Congress}},
title = {{S.2925: Management of Individuals' Neural Data Act of 2025}},
year = {2025},
url = {https://www.congress.gov/bill/119th-congress/senate-bill/2925},
note = {Introduced September 29, 2025}
}

@misc{oecd2019neurotechnology,
author = {{Organisation for Economic Co-operation and Development (OECD)}},
title = {Recommendation of the Council on Responsible Innovation in Neurotechnology},
year = {2019},
institution = {OECD},
number = {OECD/LEGAL/0457},
url = {https://legalinstruments.oecd.org/en/instruments/OECD-LEGAL-0457},
note = {Adopted 11 December 2019}
}

@article{jo2024eeg,
  title={Are eeg-to-text models working?},
  author={Jo, Hyejeong and Yang, Yiqian and Han, Juhyeok and Duan, Yiqun and Xiong, Hui and Lee, Won Hee},
  journal={arXiv preprint arXiv:2405.06459},
  year={2024}
}

@article{jo2025evaluating,
  title={Evaluating EEG-to-text models through noise-based performance analysis},
  author={Jo, Hyejeong and Yang, Yiqian and Han, Juhyeok and Duan, Yiqun and Xiong, Hui and Lee, Won Hee},
  journal={Scientific Reports},
  year={2025},
  publisher={Nature Publishing Group UK London}
}

@article{li2026cross,
  title={Cross-subject generalization for EEG decoding: a survey of deep learning methods},
  author={Li, Taida and Yan, Yujun and Dou, Fei and Song, Wenzhan and Zhang, Xiang},
  journal={Progress in Biomedical Engineering},
  volume={8},
  number={2},
  pages={022013},
  year={2026},
  publisher={IOP Publishing}
}

@article{brookshire2024data,
  title={Data leakage in deep learning studies of translational EEG},
  author={Brookshire, Geoffrey and Kasper, Jake and Blauch, Nicholas M and Wu, Yunan Charles and Glatt, Ryan and Merrill, David A and Gerrol, Spencer and Yoder, Keith J and Quirk, Colin and Lucero, Ch{\'e}},
  journal={Frontiers in neuroscience},
  volume={18},
  pages={1373515},
  year={2024},
  publisher={Frontiers Media SA}
}

@misc{wang2026causesperformancedegradationcrosssubject,
      title={What Causes Performance Degradation in Cross-Subject EEG Classification?}, 
      author={Yihe Wang and Taida Li and Yujun Yan and Wenzhan Song and Xiang Zhang},
      year={2026},
      eprint={2410.03057},
      archivePrefix={arXiv},
      primaryClass={cs.CE},
      url={https://arxiv.org/abs/2410.03057}, 
}

@article{wang2025generative,
  title={Generative AI enables EEG super-resolution via spatio-temporal adaptive diffusion learning},
  author={Wang, Shuqiang and Zhou, Tong and Shen, Yanyan and Li, Ye and Huang, Guoheng and Hu, Yong},
  journal={IEEE Transactions on Consumer Electronics},
  volume={71},
  number={1},
  pages={1034--1045},
  year={2025},
  publisher={IEEE}
}

@article{xu2025alljoined,
  title={Alljoined-1.6 M: A Million-Trial EEG-Image Dataset for Evaluating Affordable Brain-Computer Interfaces},
  author={Xu, Jonathan and Nunes, Ugo Bruzadin and Jiang, Wangshu and Ryther, Samuel and Pringle, Jordan and Scotti, Paul S and Delorme, Arnaud and Kneeland, Reese},
  journal={arXiv preprint arXiv:2508.18571},
  year={2025}
}

@article{zhou2024belt,
  title={Belt-2: Bootstrapping eeg-to-language representation alignment for multi-task brain decoding},
  author={Zhou, Jinzhao and Duan, Yiqun and Chang, Fred and Do, Thomas and Wang, Yu-Kai and Lin, Chin-Teng},
  journal={arXiv preprint arXiv:2409.00121},
  year={2024}
}

@article{kneeland2026enigma,
  title={ENIGMA: EEG-to-Image in 15 Minutes Using Less Than 1\% of the Parameters},
  author={Kneeland, Reese and Jiang, Wangshu and Nunes, Ugo Bruzadin and Scotti, Paul Steven and Delorme, Arnaud and Xu, Jonathan},
  journal={arXiv preprint arXiv:2602.10361},
  year={2026}
}

@inproceedings{xiongeeg,
  title={EEG-FM-Bench: A Comprehensive Benchmark for the Systematic Evaluation and Diagnostic Analyses of EEG Foundation Models},
  author={Xiong, Wei and Li, Jiangtong and Li, Jie and Zhu, Kun and Jiang, Changjun},
  booktitle={Forty-third International Conference on Machine Learning},
  year={2026}
}

@article{park2025eeg,
  title={EEG-to-Voice Decoding of Spoken and Imagined speech Using Non-Invasive EEG},
  author={Park, Hanbeot and Cho, Yunjeong and Kim, Hunhee},
  journal={arXiv preprint arXiv:2512.22146},
  year={2025}
}

@article{ma2025fully,
  title={Fully end-to-end EEG to speech translation using multi-scale optimized dual generative adversarial network with cycle-consistency loss},
  author={Ma, Chen and Zhang, Yue and Guo, Yina and Liu, Xin and Shangguan, Hong and Wang, Juan and Zhao, Luqing},
  journal={Neurocomputing},
  volume={616},
  pages={128916},
  year={2025},
  publisher={Elsevier}
}

@inproceedings{szegedy2017inception,
  title={Inception-v4, inception-resnet and the impact of residual connections on learning},
  author={Szegedy, Christian and Ioffe, Sergey and Vanhoucke, Vincent and Alemi, Alexander},
  booktitle={Proceedings of the AAAI conference on artificial intelligence},
  volume={31},
  number={1},
  pages = {4278–4284},
  year={2017}
}

@article{sabharwal2024comprehensive,
  title={Comprehensive Review of EEG-to-Output Research: Decoding Neural Signals into Images, Videos, and Audio},
  author={Sabharwal, Yashvir and Rama, Balaji},
  journal={arXiv preprint arXiv:2412.19999},
  year={2024}
}

@article{barbera2026using,
  title={On using ai for eeg-based bci applications: problems, current challenges and future trends},
  author={Barbera, Thomas and Burger, Jacopo and D’Amelio, Alessandro and Zini, Simone and Bianco, Simone and Lanzarotti, Raffaella and Napoletano, Paolo and Boccignone, Giuseppe and Contreras-Vidal, Jose Luis},
  journal={International Journal of Human--Computer Interaction},
  volume={42},
  number={11},
  pages={7791--7810},
  year={2026},
  publisher={Taylor \& Francis}
}

@article{wang2026generative,
  title={Generative AI Empowers Brain-Computer Interfaces: A Review-Perspective on Technical Realities and Future Visions},
  author={Wang, Shuqiang and Guo, Yi and Dong, Yihang and Shen, Yanyan and Zhang, Zhiguo and Cheung, Albert C and Pu, Jiangbo and Zhong, Sheng-hua and Tong, Raymond Kai-Yu and Li, Ye and others},
  journal={IEEE Transactions on Consumer Electronics},
  year={2026},
  publisher={IEEE}
}

@article{habashi2023generative,
  title={Generative adversarial networks in EEG analysis: an overview},
  author={Habashi, Ahmed G and Azab, Ahmed M and Eldawlatly, Seif and Aly, Gamal M},
  journal={Journal of neuroengineering and rehabilitation},
  volume={20},
  number={1},
  pages={40},
  year={2023},
  publisher={Springer}
}

@article{eldawlatly2024role,
  title={On the role of generative artificial intelligence in the development of brain-computer interfaces},
  author={Eldawlatly, Seif},
  journal={BMC Biomedical Engineering},
  volume={6},
  number={1},
  pages={4},
  year={2024},
  publisher={Springer}
}


\clearpage
\appendix

\section{Baseline Datasets and Implementations}
\label{appendix:baseline}

In this section, we summarize key datasets, one for each modality, previously described in the modality-specific subsections, together with reference resources for recommended baseline architectures used to generate images, text, and audio from EEG signals. Our aim is to provide researchers with a clear starting point by presenting both widely used open-source datasets and representative baseline models for each modality. 

\subsection{EEG-to-Image}

\noindent \textbf{Key resources:} EEG-Based Visual Classification dataset \cite{spampinato2017deep, palazzo2020decoding} and EEG encoder framework for Brain2Image \cite{kavasidis2017brain2image}.

\begin{description}
    \item[Dataset:] The EEG-Based Visual Classification dataset contains 128-channel EEG recordings from 6 subjects while viewing images from the ImageNet dataset. Stimuli comprised 40 object classes with 50 images per class, for a total of 2,000 images. Dataset is publicly available at 
\href{https://tinyurl.com/eeg-visual-classification}{this link}.

    \item[EEG encoder framework:] An open-source repository of LSTM- and deep neural network (DNN)-based EEG encoders, primarily developed for visual classification tasks from EEG signals. Code implementation is available at \href{https://github.com/perceivelab/eeg_visual_classification}{this link}.

\end{description}

\subsection{EEG-to-Text}

\noindent \textbf{Key resources:} Zurich Cognitive Language Processing Corpus 2.0 (ZuCo 2.0) dataset \cite{hollenstein2019zuco}; EEG-to-Text sequence-to-sequence decoding framework leveraging pre-trained language models (e.g., BART) \cite{wang2022open}.

\begin{description}
    \item[Dataset:] The ZuCo 2.0 dataset provides multimodal recordings of eye gaze and 128-channel EEG brain activity from 18 participants. Subjects read a stimulus set of 739 English sentences, including 349 in a normal reading paradigm and 390 in a task-specific paradigm. Dataset is publicly available at \href{https://osf.io/2urht/wiki/home/}{this link}.
    
    \item[EEG-to-Text decoding framework:] An open-source sequence-to-sequence framework for decoding EEG signals into text, leveraging pre-trained language models such as BART. Code implementation is available at \href{https://github.com/MikeWangWZHL/EEG-To-Text}{this link}.
\end{description}

\subsection{EEG-to-Audio}

\noindent \textbf{Key resources:} KARA One dataset \cite{zhao2015classifying}; Voice reconstruction framework from EEG during imagined speech using mel-spectrograms as intermediate representations \cite{lee2023towards}.

\begin{description}
    \item[Dataset:] The KARA One dataset provides multimodal recordings from 8 participants across three modalities: 64-channel EEG, facial motion, and audio. Data were collected during both imagined and vocalized phonemic and single-word prompts. Dataset is publicly available at \href{https://www.cs.toronto.edu/~complingweb/data/karaOne/karaOne.html}{this link}.
    
    \item[EEG-to-Speech synthesis framework:] The EEG-to-speech synthesis framework leverages mel-spectrograms as an intermediate representation. It comprises (i) a generator that maps EEG embeddings to mel-spectrograms, (ii) a discriminator that evaluates real versus generated spectrograms, and (iii) a vocoder that converts mel-spectrograms into audio waveforms. Code implementation is available at \href{https://github.com/youngeun1209/NeuroTalk}{this link}.
\end{description}

\textit{We further encourage future work to adopt these  suggested baselines to facilitate meaningful performance comparisons across architectures, thereby enabling more systematic and reproducible progress in generative modeling from EEG signals.}

\newpage
\onecolumn
\section{Ethical requirements of research trial}
\label{appendix:ethical_requirements}
\setcounter{table}{0}

\begin{table}[H]
    \centering
    \footnotesize
    \renewcommand{\arraystretch}{1.3}
    \resizebox{\textwidth}{!}{%
        \begin{tabular}{p{0.18\linewidth}p{0.34\linewidth}p{0.22\linewidth}p{0.26\linewidth}}
            \toprule
            \textbf{Requirement} &
            \textbf{Explanation} &
            \textbf{Justifying Ethical Values} &
            \textbf{Expertise for Evaluation} \\
            \midrule

            Social or scientific value &
            Evaluation of a treatment, intervention, or theory that will improve health and well-being or increase knowledge &
            Scarce resources and nonexploitation &
            Scientific knowledge; citizens' understanding of social priorities \\
            \midrule

            Scientific validity &
            Use of accepted scientific principles and methods, including statistical techniques, to produce reliable and valid data &
            Scarce resources and nonexploitation &
            Scientific and statistical knowledge; knowledge of the condition and population to assess feasibility \\
            \midrule

            Fair subject selection &
            Selection of subjects so that stigmatized and vulnerable individuals are not targeted for risky research and the rich and socially powerful are not favored for potentially beneficial research &
            Justice &
            Scientific knowledge; ethical and legal knowledge \\
            \midrule

            Favorable risk--benefit ratio &
            Minimization of risks; enhancement of potential benefits; risks to the subject are proportionate to the benefits to the subject and society &
            Nonmaleficence, beneficence, and nonexploitation &
            Scientific knowledge; citizens' understanding of social values \\
            \midrule

            Independent review &
            Review of the design of the research trial, its proposed subject population, and risk-benefit ratio by individuals unaffiliated with the research &
            Public accountability; minimizing influence of potential conflicts of interest &
            Intellectual, financial, and otherwise independent researchers; scientific and ethical knowledge \\
            \midrule

            Informed consent &
            Provision of information to subjects about the purpose of the research, its procedures, potential risks, benefits, and alternatives, so that the individual understands this information and can make a voluntary decision whether to enroll and continue to participate &
            Respect for subject autonomy &
            Scientific knowledge; ethical and legal knowledge \\
            \midrule

            Respect for potential and enrolled subjects &
            Respect for subjects by: (1) permitting withdrawal from the research; (2) protecting privacy through confidentiality; (3) informing subjects of newly discovered risks or benefits; (4) informing subjects of results of clinical research; and (5) maintaining the welfare of subjects &
            Respect for subject autonomy and welfare &
            Scientific knowledge; ethical and legal knowledge; knowledge of the particular subject population \\
            \bottomrule
        \end{tabular}%
    }
    \caption{Seven requirements for determining whether a research trial is ethical, adapted from \citet{emanuel2000makes}. The requirements are listed chronologically from the conception of research through its formulation and implementation.}
    \label{tab:ethical_requirements}
\end{table}


\end{document}